\documentclass{article}

\usepackage[nonatbib,main,final]{neurips_2026}

\usepackage[utf8]{inputenc} 
\usepackage[T1]{fontenc}    
\usepackage{hyperref}       
\usepackage{url}            
\usepackage{booktabs}       
\usepackage{amsfonts}       
\usepackage{nicefrac}       
\usepackage{microtype}      
\usepackage{xcolor}         

\usepackage[numbers,sort&compress]{natbib}
\usepackage{amssymb}
\usepackage{amsmath}
\usepackage{algorithm}
\usepackage{algpseudocode}
\usepackage{caption}
\usepackage{soul}
\usepackage{wrapfig}
\usepackage{makecell}
\usepackage{subcaption}
\usepackage{graphicx}
\usepackage{enumitem}
\usepackage{multirow}
\usepackage{placeins}
\usepackage{wrapfig}

\usepackage{amsthm}

\newtheorem{proposition}{Proposition}
\newtheorem{corollary}{Corollary}

\newtheorem{remark}{Remark}

\title{Neural Field Thermal Tomography: A Differentiable Physics Framework for Non-Destructive Evaluation}

\author{%
  Tao Zhong$^{1}$,
  Yixun Hu$^{1}$,
  Dongzhe Zheng$^{1}$,
  Aditya Sood$^{1}$,
  Christine Allen-Blanchette$^{1}$ \\
  $^1$Princeton University\\
  \texttt{\{tzhong, ca15\}@princeton.edu}
}

\begin{document}

\maketitle

\begin{abstract}
Inverse problems for stiff parabolic partial differential equations (PDEs), such as the inverse heat conduction problem (IHCP), are severely ill-posed: the forward map rapidly damps high-frequency interior structure before it reaches the boundary. Soft-constrained physics-informed neural networks (PINNs), which embed the PDE as a residual penalty, decouple the diffusivity from the surface data in their standard two-network inverse formulation, and the evaluated variants fit boundary measurements while leaving the interior field essentially untouched. We propose Neural Field Thermal Tomography (NeFTY), a hard-constrained neural field framework for label-free three-dimensional inverse heat conduction. NeFTY represents the unknown diffusivity as a continuous coordinate-based neural network, and at every optimization step passes the candidate field through a differentiable implicit-Euler heat solver with harmonic-mean interface flux, so that the simulated temperature satisfies the discretized heat equation up to the solver tolerance. Adjoint gradients propagate the surface reconstruction error back to the network weights at solver-level memory cost, making test-time inversion tractable on a single GPU. Across synthetic 3D benchmarks, NeFTY outperforms soft-constrained PINN variants, a voxel-grid baseline, and classical L-BFGS voxel inversion through the same solver on label-free volumetric recovery, and it transfers to real thermography data, surpassing classical signal-processing baselines in both defect segmentation and depth estimation. Additional details at \href{https://cab-lab-princeton.github.io/nefty/}{\small \texttt{cab-lab-princeton.github.io/nefty}}.
\end{abstract}

\section{Introduction}
Heat diffusion blurs interior structure before it reaches the surface: high spatial frequencies inside a specimen are damped on their way to the boundary, so the map from internal material properties to surface measurements discards much of the information needed to invert it~\cite{hadamard1888rayon,gahleitner2024photothermal}. Inverse problems for such stiff parabolic partial differential equations (PDEs) are therefore severely ill-posed. Soft physics-informed losses can fit boundary data without recovering the underlying field, because their gradients become poorly conditioned and dominated by the surface signal in this stiff regime~\cite{raissi2019physics,wang2022and}. Classical adjoint-based inverse heat conduction~\cite{jarny1991general,alifanov1994inverse} and differentiable physics~\cite{onken2020discretize,holl2020learning,holl2024bf,de2018end} build the discretized solver into the optimization loop, so the governing equations hold as a hard constraint at every step. We combine this hard constraint with a neural-field prior for the three-dimensional inverse heat conduction problem (IHCP).

\begin{wrapfigure}{r}{0.42\linewidth}
  \centering
  \includegraphics[width=\linewidth]{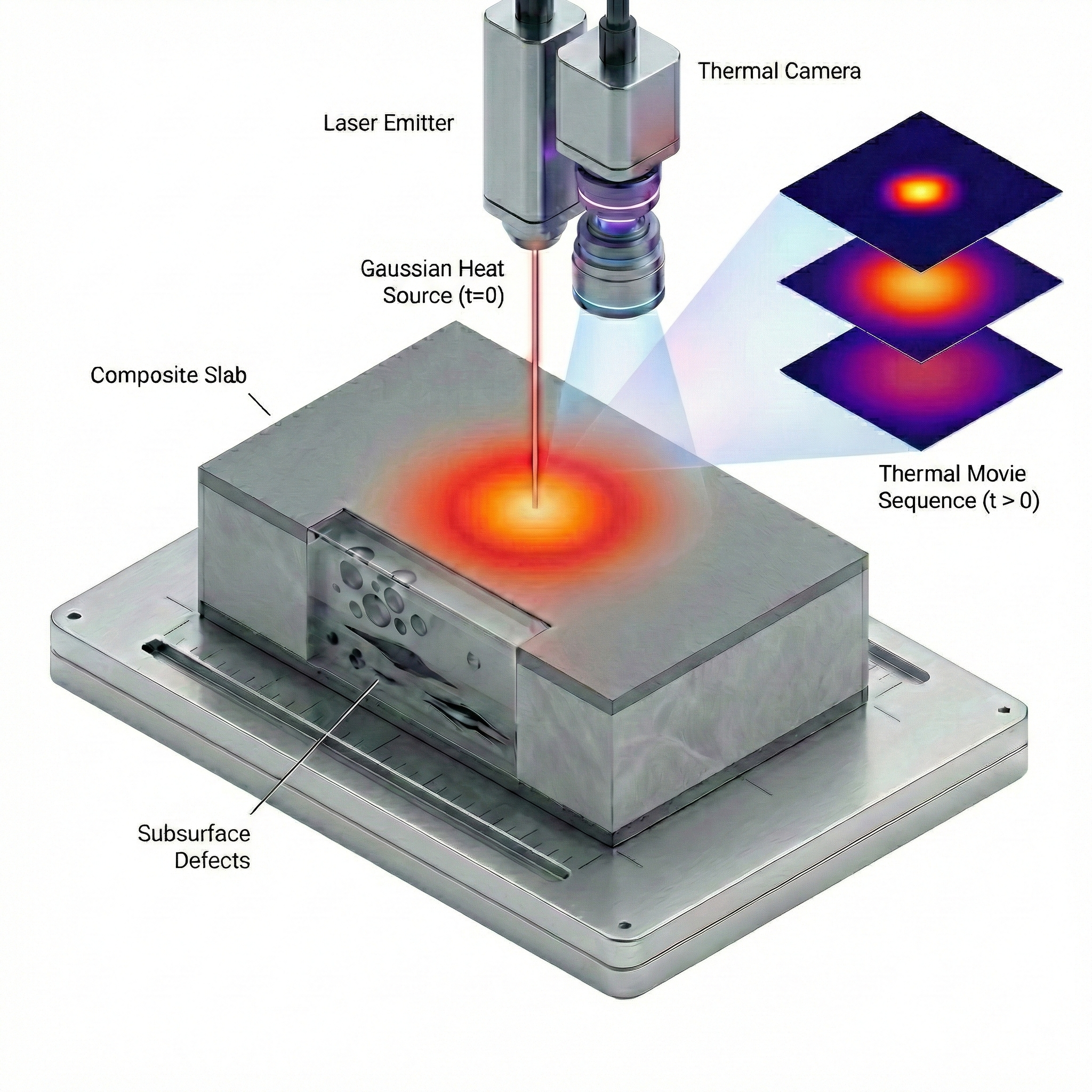}
  \captionsetup{belowskip=-6pt}
  \caption{\textbf{IHCP setup.} A high-speed camera measures the transient surface temperature following pulsed laser heating. NeFTY reconstructs the 3D subsurface diffusivity field from these measurements.}
  \label{fig:teaser}
  \vspace{-10pt}
\end{wrapfigure}

Non-destructive testing (NDT) by pulsed thermography is the working example throughout (Figure~\ref{fig:teaser}). A pulsed laser deposits energy on the specimen surface, and a high-speed camera records the subsequent surface temperature decay. Subsurface delaminations, voids, and inclusions perturb the diffusive flux and leave faint thermal contrast on the boundary~\cite{kovacs2020deep,rosa2025advanced,peng2025machine}. Reconstructing the volumetric thermal diffusivity field $\alpha(x,y,z)$ from these surface signals is the inverse heat conduction problem. Unlike hyperbolic ultrasonic or radar imaging~\cite{burgholzer2018acoustic,vavilov1992dynamic}, the parabolic forward map damps
internal features rapidly with depth, so small surface perturbations can correspond to large interior variations and the inverse map becomes severely ill-posed~\cite{leontiou2024three,qian2023physics}.

Existing thermography and learning approaches each address part of this volumetric inversion. Classical thermography pipelines enhance defect contrast in transformed signal domains: Thermographic Signal Reconstruction (TSR)~\cite{shepard2002reconstruction,shepard2015advances} and Pulsed Phase Thermography (PPT)~\cite{maldague2002advances} reduce to 1D pixel-wise inversions that ignore lateral diffusion and bias estimates for shallow or low-aspect-ratio defects~\cite{perez2025integrating}, and the Virtual Wave Concept (VWC)~\cite{burgholzer2017three} images defect geometry. None recovers a volumetric diffusivity field. Soft-constrained physics-informed neural networks (PINNs) target the volumetric problem in principle, yet in their standard two-network form they meet a gradient pathology in stiff transient diffusion: the PDE residual is high and rugged, so the temperature surrogate fits the boundary measurements while the diffusivity network saturates at a near-trivial field~\cite{wang2022and}. Supervised volumetric learning sidesteps these issues but requires 3D ground-truth labels that are unavailable in real NDT.

To this end, we introduce Neural Field Thermal Tomography (NeFTY), a hard-constrained neural field framework for label-free 3D inverse heat conduction. NeFTY represents the unknown diffusivity as a continuous coordinate-based neural network~\cite{sitzmann2020implicit,mildenhall2021nerf} with positional encoding, frequency annealing, and a bounded output. At every optimization step, the candidate field is passed through a differentiable implicit-Euler heat solver with harmonic-mean interface flux, so the temperature satisfies the discretized heat equation up to the iterative solver tolerance. Adjoint gradients propagate the surface reconstruction error back to the network weights at solver-level memory cost, making test-time inversion tractable on a single GPU. On synthetic homogeneous and layered settings, NeFTY reaches volumetric IoU of $0.45$ and $0.37$, while soft-constrained PINN variants and a voxel-grid baseline remain at or near zero, and it outperforms classical thermography baselines on real PVC thermal data in both 2D defect segmentation and 2.5D depth recovery.

We summarize our contributions as follows:
\begin{itemize}[leftmargin=*,topsep=2pt,itemsep=1pt]
    \item We propose a hard-constrained neural field formulation for label-free 3D inverse heat conduction from surface-only thermal sequences, combining a continuous coordinate-based prior with a differentiable implicit heat solver and adjoint gradients. To our knowledge, NeFTY is the first label-free ML-based method to recover a full 3D diffusivity field from surface-only thermal measurements.
    \item We give a theoretical and empirical account of soft-constrained PINNs on stiff parabolic inverse problems: in the standard two-network formulation with data, residual, and initial-condition terms, surface data reaches the diffusivity only through the residual, and across four evaluated PINN variants low surface error does not imply correct volumetric recovery.
    \item We carry out an evaluation that pairs synthetic 3D label-free reconstruction with real PVC thermal data, soft-constrained PINN and classical L-BFGS baselines, robustness studies on geometry, noise, and contrast, frequency- and edge-domain diagnostics, component ablations, and a wall-clock, memory, and scaling analysis of the adjoint solver.
\end{itemize}

\section{Related Work}

\textbf{Thermographic Non-Destructive Testing} traditionally extracts subsurface defect information through signal-processing transforms applied to surface temperature decays. Thermographic Signal Reconstruction (TSR)~\cite{shepard2002reconstruction,shepard2015advances} fits log-time polynomials to enhance defect contrast, Pulsed Phase Thermography (PPT)~\cite{maldague2002advances,chung2021latest} maps the decays into a frequency phase domain that suppresses noise and emissivity variations, and the Virtual Wave Concept (VWC)~\cite{burgholzer2017three,schager2020extension} deconvolves the diffusive forward map into a pseudo-wave for depth estimation. Quantitative-thermography refinements add calibrated depth correlations and effective-diffusivity estimates~\cite{ma2025quantitative,perez2025integrating,ali2025effective}, while TSR/PPT-style depth inversions remain pixel-wise and amplify high-frequency measurement noise. In parallel, supervised convolutional pipelines based on U-Net~\cite{ronneberger2015u,oliveira2021employing,shi2021infrared,fang2023automatic} learn defect masks directly from thermal sequences, and several recent surveys~\cite{peng2025machine,liu2025review,kovacs2020deep,rosa2025advanced} catalogue this growing detection-focused literature. The closest prior systems reconstruct 3D defect geometry~\cite{sirikham2020three}, learn defect depth~\cite{wu2023automatic}, or recover internal temperature fields~\cite{leontiou2024three}. None recovers a material-property volume. Pixel-wise transforms such as TSR and PPT ignore lateral heat flux, and supervised volumetric models require 3D diffusivity labels that are unavailable in real NDT. NeFTY recovers a full 3D diffusivity field directly from the surface sequence, without volumetric labels and without 1D pixel-wise approximations.

\textbf{Physics-Informed Neural Networks} embed governing equations as soft penalties alongside data terms in scientific machine learning. Standard PINNs~\cite{raissi2019physics,cai2021physics,leontiou2024three} jointly minimize boundary fit and a PDE residual, and a series of variants attack the resulting optimization pathologies: separable architectures in SPINN~\cite{cho2023spinn} scale up collocation throughput, temporal-causality weighting in Causal-PINN~\cite{wang2024causal} stabilizes long-horizon transients, and dual-cone gradient descent in DCGD~\cite{hwang2024dcgd} resolves residual-versus-data gradient conflicts. Diagnostic studies document conditioning, spectral-bias, and loss-balancing difficulties of PDE-residual penalties in stiff regimes~\cite{krishnapriyan2021characterizing,wang2022and,hao2024training}, even with adaptive loss balancing such as GradNorm~\cite{chen2018gradnorm}. Section~\ref{sec:pinn} analyzes the gradient decoupling of the two-network inverse formulation evaluated here. A complementary line replaces the penalty with a differentiable solver: discretize-then-optimize~\cite{onken2020discretize}, differentiable PDE engines~\cite{holl2020learning,holl2024bf,bouziani2024differentiable}, and end-to-end differentiable physics for control and robotics~\cite{de2018end,degrave2019differentiable,turpin2023fast,zhong2025gagrasp} all enforce the discretized governing equations at every optimization step. Classical IHCP inversion follows the same principle: adjoint-based optimization~\cite{jarny1991general}, iterative regularization~\cite{alifanov1994inverse}, and textbook treatments~\cite{ozisik2021inverse} enforce the heat equation through a numerical solver, within the broader theory of PDE-constrained optimization~\cite{hinze2009optimization}. NeFTY belongs to this discretize-then-optimize adjoint family and contributes a neural-field prior over the diffusivity, which enables label-free 3D recovery from surface-only measurements.

\textbf{Coordinate-based Neural Fields} parameterize a continuous signal as a multilayer perceptron (MLP) over spatial coordinates, typically combined with positional encoding or sinusoidal activations. SIREN~\cite{sitzmann2020implicit} and NeRF~\cite{mildenhall2021nerf} established neural fields as a high-frequency representation prior for visual signals, and deformable extensions such as Nerfies~\cite{park2021nerfies} broadened the class of recoverable scenes. Scientific imaging has begun to adopt the same representation for tomographic and field-reconstruction problems, including X-ray computed tomography~\cite{xu2025tomograf,zhou2025rho}, dynamic medical imaging via ProxNF~\cite{lozenski2024proxnf}, and fluid scalar fields with FluidNeRF~\cite{kelly2023fluidnerf}. These works contribute a continuous, mesh-free function space, and the parameterization itself does not enforce a governing PDE. NeFTY uses the neural field as the diffusivity prior and couples it with a differentiable heat solver. The two are complementary, and with the solver and adjoint held fixed, the neural-field prior carries the largest share of NeFTY's gain over voxel-field inversion among the evaluated configurations (Section~\ref{ssec:synth}, Appendix~\ref{appdx:classical}).

\section{Forward Model, Ill-Posedness, and Soft-Constraint Pathology}
\label{sec:prelim}

\subsection{Forward model and effective-diffusivity parameterization}
\label{sec:forward}

The conservation form of transient heat conduction in an isotropic medium is posed on a bounded Lipschitz domain $\Omega \subset \mathbb{R}^3$, with boundary $\partial \Omega$ and observation window $\Gamma_{\mathrm{obs}} \subseteq \partial \Omega$, as follows:
\begin{equation}
\label{eq:full_pde}
    \rho(\mathbf{x}) C_p(\mathbf{x}) \, \partial_t T \;=\; \nabla \cdot \big( k(\mathbf{x}) \, \nabla T \big) + Q(\mathbf{x}, t), \qquad \mathbf{x} \in \Omega, \; t \in (0, t_{\mathrm{end}}],
\end{equation}
with mass density $\rho$, specific heat capacity $C_p$, thermal conductivity $k$, and source $Q$ ($Q \equiv 0$ post flash). Dividing~\eqref{eq:full_pde} by $\rho C_p$ and applying the product rule gives the equivalent diffusivity form
\begin{equation}
\label{eq:diffusivity_form}
    \partial_t T \;=\; \nabla \cdot \big( \alpha(\mathbf{x}) \nabla T \big) \;-\; k(\mathbf{x}) \, \nabla T \cdot \nabla \!\big( \tfrac{1}{\rho C_p} \big), \qquad \alpha(\mathbf{x}) := \frac{k(\mathbf{x})}{\rho(\mathbf{x}) C_p(\mathbf{x})},
\end{equation}
for smooth, strictly positive $\rho C_p$ (Appendix~\ref{appdx:effective_derivation}). When $\rho C_p$ is uniform within each phase, as in a host material with embedded defects, the second term vanishes inside each phase and acts only on the interfaces $\Sigma$ between them, where a finite-volume grid cannot represent it directly~\cite{patankar1980numerical} (Appendix~\ref{appdx:effective_correction}). NeFTY therefore parameterizes a single effective field $\alpha(\mathbf{x})$ obeying $\partial_t T = \nabla \cdot (\alpha \nabla T)$ within each phase and restores flux continuity across interfaces through the harmonic mean at cell faces.

\begin{proposition}[Harmonic mean as discrete interface continuity for the effective-$\alpha$ flux]
\label{prop:harmonic}
For a one-dimensional cell pair with constant cell-centered diffusivities $\alpha_i, \alpha_{i+1}$, cell spacing $\Delta x$, cell-centered temperatures $T_i, T_{i+1}$, and effective $\alpha$-flux $j_\alpha := -\alpha \partial_x T$, the unique flux satisfying continuity across the cell face $x_{i+1/2}$ is $j_\alpha = -\bar{\alpha}_{i+1/2}(T_{i+1} - T_i)/\Delta x$ with $\bar{\alpha}_{i+1/2} = 2 \alpha_i \alpha_{i+1}/(\alpha_i + \alpha_{i+1})$.
\end{proposition}

The proof, multi-dimensional consequence, and high-contrast behavior are in Appendix~\ref{appdx:effective}. Synthetic and real-PVC initial and boundary condition settings are specified in Appendix~\ref{appdx:bc}.

\subsection{The inverse problem and its ill-posedness}
The IHCP recovers $\alpha$ from noisy surface measurements $\hat{T}(\mathbf{x}, t_i)$, $\mathbf{x} \in \Gamma_{\mathrm{obs}}$, $t_i \in \{t_1, \dots, t_M\} \subset (0, t_{\mathrm{end}}]$. Let $\mathcal{K}(\alpha) := T_\alpha|_{\Gamma_{\mathrm{obs}} \times (0, t_{\mathrm{end}})}$ denote the parameter-to-observation operator (with $T_\alpha$ the
\label{sec:inverse}
\begin{wrapfigure}{r}{0.40\linewidth}
\centering
\includegraphics[width=\linewidth]{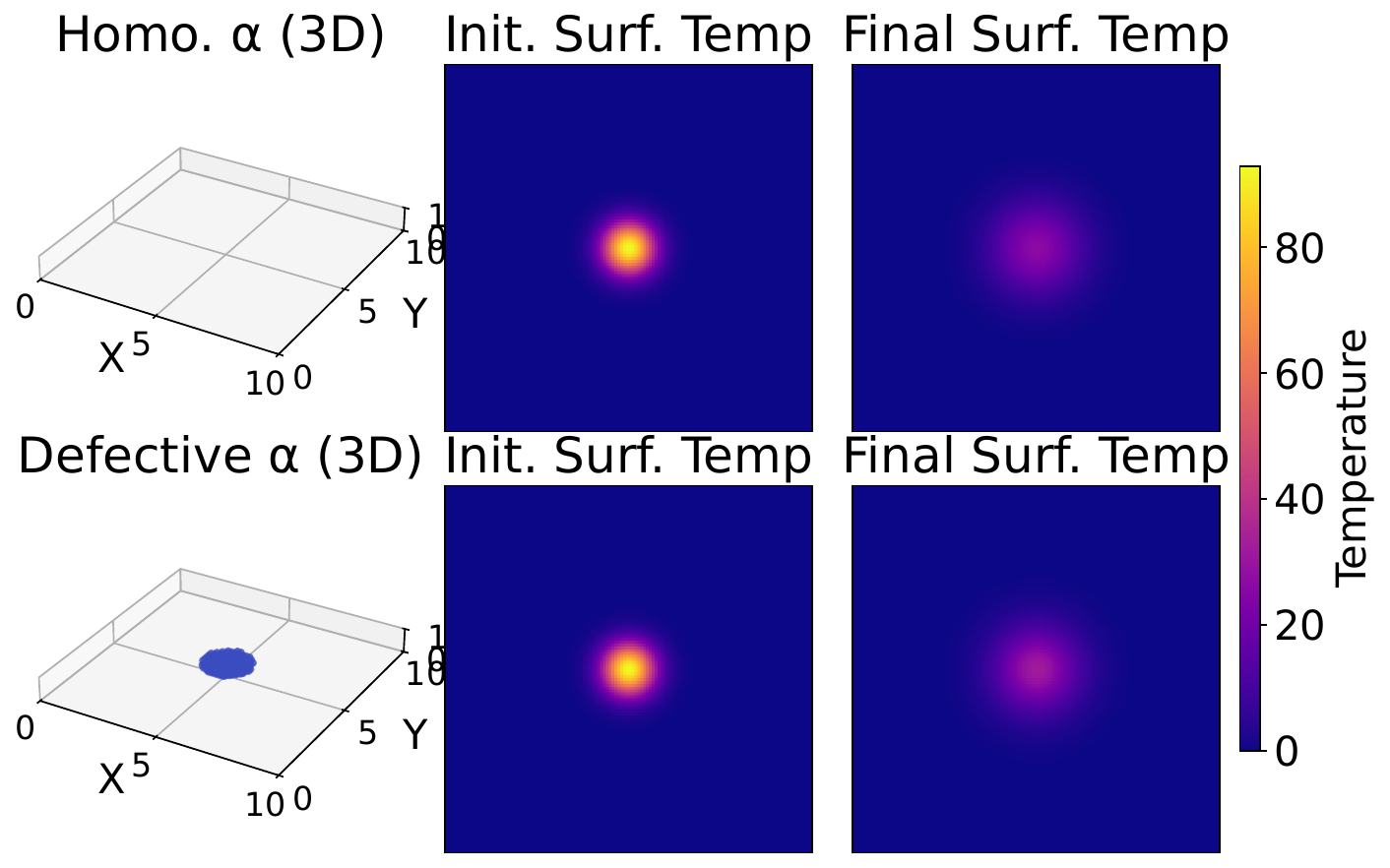}
\captionsetup{belowskip=-4pt}
\caption{\textbf{Empirical ill-posedness.} Distinct interiors yield nearly indistinguishable surface profiles, illustrating Prop.~\ref{prop:svd}.}
\label{fig:illpose}
\vspace{-40pt}
\end{wrapfigure}
solution of~\eqref{eq:diffusivity_form} under fixed initial and lateral boundary conditions) acting on the admissible set $\mathcal{A} \subset H^1(\Omega) \cap L^\infty(\Omega)$. The reconstruction objective is
\begin{equation}
\label{eq:ihcp_obj}
\begin{aligned}
    \alpha^{*} \;&=\; \arg\min_{\alpha \in \mathcal{A}} \, \mathcal{J}(\alpha), \\
    \mathcal{J}(\alpha) \;&=\; \sum_{i=1}^{M} \big\| \mathcal{K}(\alpha)(\cdot, t_i) - \hat{T}(\cdot, t_i) \big\|_{\Gamma_{\mathrm{obs}}}^2 + \lambda \mathcal{R}(\alpha),
\end{aligned}
\end{equation}
with $\|\cdot\|_{\Gamma_{\mathrm{obs}}}$ the $L^2$ norm over $\Gamma_{\mathrm{obs}}$, $\mathcal{R}$ a regularization functional, and $\lambda > 0$. The IHCP is widely recognized as one of the most severely ill-posed problems in mathematical physics~\cite{hadamard1888rayon,martinez2024artificial,engl1996regularization}. We quantify it by bounding the singular values of the linearized observation map.

\begin{proposition}[Compactness and algebraic singular-value decay]
\label{prop:svd}
Let $\alpha_0 > 0$ be a constant background diffusivity with $\nabla T_{\alpha_0} \in L^\infty(\Omega \times (0, t_{\mathrm{end}}))$. The Fr\'echet derivative $d\mathcal{K}_{\alpha_0}: H^1(\Omega) \to L^2(\Gamma_{\mathrm{obs}} \times (0, t_{\mathrm{end}}))$ is bounded and compact, with singular values $\sigma_n \le C \, (1 + \lambda_n)^{-1/2}$, where $\{\lambda_j\}_{j \ge 0}$ are the Laplacian eigenvalues on $\Omega$ under the imposed boundary conditions. On the slab, Weyl's law gives $\sigma_n \lesssim n^{-1/3}$.
\end{proposition}

\begin{corollary}[Hadamard ill-posedness]
\label{cor:hadamard}
The pseudo-inverse $d\mathcal{K}_{\alpha_0}^{\dagger}$ is unbounded, with $\sigma_n^{-1} \ge C^{-1} (1 + \lambda_n)^{1/2}$ amplifying noise along the $n$-th left singular vector.
\end{corollary}

In practical terms, finer interior structure leaves progressively weaker surface signatures, and inverting the map amplifies measurement noise without bound. Proof in Appendix~\ref{appdx:illposed}. Figure~\ref{fig:illpose} visualizes the smoothing. NeFTY counters this with a neural-field prior plus frequency annealing (Section~\ref{sec:method}) and a total-variation penalty $\mathcal{R}(\alpha)$.

\subsection{Soft-constraint pathology in stiff parabolic inversion}
\label{sec:pinn}

PINNs are the dominant and broadly successful paradigm for physics-informed learning. Our analysis concerns their standard two-network inverse formulation, which jointly optimizes a temperature surrogate $T_\phi(\mathbf{x}, t)$ with parameters $\phi$ and a diffusivity surrogate $\alpha_\theta(\mathbf{x})$ with separate parameters $\theta$, against
\begin{equation}
\label{eq:pinn_loss}
    \mathcal{L}_{\mathrm{PINN}} \;=\; \underbrace{\| T_\phi - \hat{T} \|^2_{\Gamma_{\mathrm{obs}}}}_{\mathcal{L}_{\mathrm{data}}} + \lambda_{\mathrm{PDE}} \underbrace{\| \partial_t T_\phi - \nabla \cdot (\alpha_\theta \nabla T_\phi) \|^2_{\Omega}}_{\mathcal{L}_{\mathrm{PDE}}} + \lambda_{\mathrm{IC}} \underbrace{\| T_\phi(\cdot, 0) - T_0 \|^2_\Omega}_{\mathcal{L}_{\mathrm{IC}}},
\end{equation}
with weights $\lambda_{\mathrm{PDE}}, \lambda_{\mathrm{IC}} > 0$ and $T_0$ the prescribed initial temperature field. Because $T_\phi$ is parameterized by $\phi$ alone, $\nabla_\theta \mathcal{L}_{\mathrm{data}} = \nabla_\theta \mathcal{L}_{\mathrm{IC}} = 0$, so the diffusivity update reduces to $\nabla_\theta \mathcal{L}_{\mathrm{PINN}} = \lambda_{\mathrm{PDE}} \nabla_\theta \mathcal{L}_{\mathrm{PDE}}$, with magnitude controlled pointwise by the Laplacian $\Delta T_\phi$ of the under-fitted temperature surrogate. Appendix~\ref{appdx:pinn} gives the full derivation and shows that adaptive loss balancing (GradNorm), separable temperature parameterizations (SPINN), causal weighting (Causal-PINN), and gradient-conflict resolution (DCGD) all preserve this decoupling.

In this formulation, surface observations reach $\theta$ only indirectly through the residual penalty, independently of the optimizer. The derivation establishes this decoupling. Combined with the unbounded amplification in Corollary~\ref{cor:hadamard}, it gives a plausible mechanism for the PINN training pathology in which $\mathcal{L}_{\mathrm{PDE}}$ shrinks while $\alpha_\theta$ remains near a trivial constant~\cite{wang2022and,hao2024training,krishnapriyan2021characterizing}, the collapse we observe in the four evaluated variants (Section~\ref{ssec:synth}). Architectures whose temperature network depends on $\theta$ lie outside this derivation. Appendix~\ref{appdx:pinn_coupled} studies such a directly coupled variant. In NeFTY, surface data propagates to $\alpha_\theta$ through the adjoint of the discretized heat operator (Section~\ref{sec:method}).

\section{Method}
\label{sec:method}

\begin{figure}[t]
  \centering
  \includegraphics[width=0.98\linewidth]{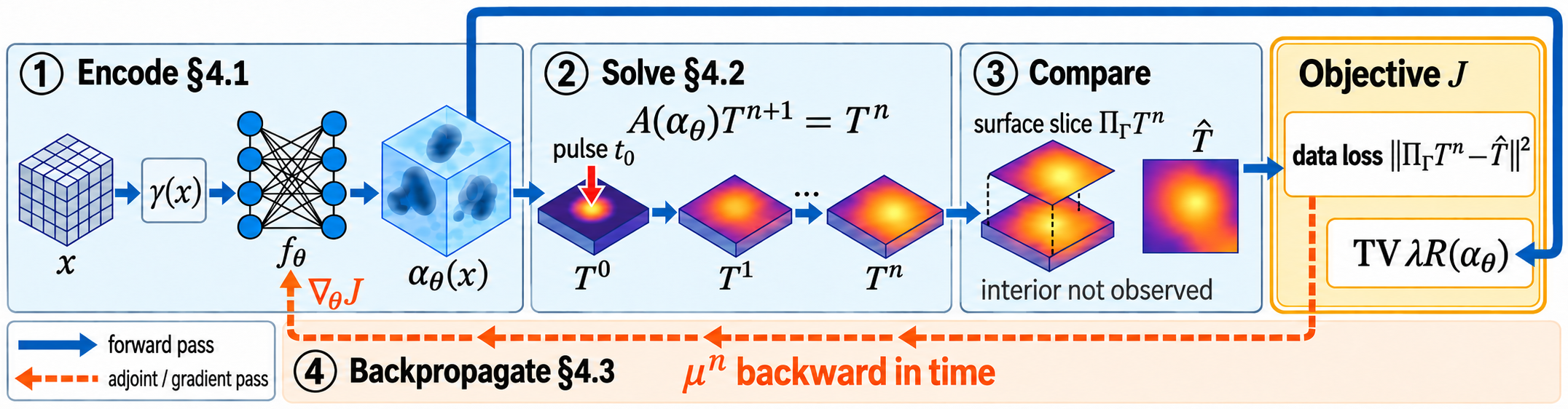}
  \captionsetup{belowskip=-4pt}
  \caption{\textbf{NeFTY pipeline}, read left to right. (1)~Encode: a coordinate MLP $f_\theta$ maps each grid node $\mathbf{x}$ through a frequency-annealed positional encoding $\gamma(\cdot)$ into a bounded diffusivity $\alpha_\theta(\mathbf{x})$ (\S\ref{sec:nf}). (2)~Solve: a differentiable implicit-Euler solver with harmonic-mean coefficients evolves the temperature $\mathbf{T}^n$ on a Cartesian grid (\S\ref{sec:hardphys}). (3)~Compare: only the surface slice of $\mathbf{T}^n$ is compared to the measurements. (4)~Backpropagate: the adjoint state $\boldsymbol{\mu}^n$ is propagated backward through the same sparse system to assemble the gradient on $\theta$ without retaining Jacobi iterates (\S\ref{sec:adjoint}).}
  \label{fig:pipeline}
\end{figure}

NeFTY recovers the diffusivity field by directly minimizing the reconstruction objective $\mathcal{J}$ of equation~\eqref{eq:ihcp_obj}, simulating each candidate field with a differentiable heat solver whose temperature satisfies the discretized heat equation up to the chosen spatial and temporal discretization and the iterative linear-solver tolerance. Each step (Figure~\ref{fig:pipeline}) encodes the diffusivity as a coordinate-based neural field $\alpha_\theta$ (\S\ref{sec:nf}), solves the implicit-Euler heat equation with harmonic-mean interface flux (\S\ref{sec:hardphys}), and propagates the surface-fidelity gradient back to $\theta$ with a discrete adjoint (\S\ref{sec:adjoint}). Because the discrete temperature is a function of $\alpha_\theta$, the gradient decoupling of Section~\ref{sec:pinn} does not arise.

\subsection{Neural diffusivity field}
\label{sec:nf}

NeFTY parameterizes the unknown effective diffusivity $\alpha_\theta : \Omega \to \mathbb{R}_{>0}$ as a coordinate multilayer perceptron (MLP) $f_\theta : \mathbb{R}^{d_\gamma} \to \mathbb{R}$ whose input is lifted through a Fourier feature map $\gamma : \mathbb{R}^3 \to \mathbb{R}^{d_\gamma}$:
\begin{equation}
\label{eq:posenc}
    \gamma(\mathbf{x}) \;=\; \big( \sin(2^0 \pi \mathbf{x}), \cos(2^0 \pi \mathbf{x}), \dots, \sin(2^{N-1} \pi \mathbf{x}), \cos(2^{N-1} \pi \mathbf{x}) \big),
\end{equation}
of bandwidth $N$. The encoding combats the spectral bias of plain MLPs and lets the network represent the sharp interfaces characteristic of subsurface defects~\cite{mildenhall2021nerf,sitzmann2020implicit}. To stabilize the high-frequency components against the algebraic noise amplification of Corollary~\ref{cor:hadamard}, we adopt the cosine frequency-annealing schedule of~\citet{park2021nerfies}, gradually unmasking encoding bands during training. The network output is hard-bracketed to the physically admissible range
\begin{equation}
\label{eq:bracket}
    \alpha_\theta(\mathbf{x}) \;=\; \alpha_{\min} + (\alpha_{\max} - \alpha_{\min}) \, \sigma\!\big( f_\theta(\gamma(\mathbf{x})) \big),
\end{equation}
with $\sigma(\cdot)$ the logistic sigmoid. The bounds $\alpha_{\min}, \alpha_{\max} > 0$ are chosen to span the bulk-to-defect contrast of the application without inducing ill-conditioning in the discrete system below. We instantiate the regularizer in~\eqref{eq:ihcp_obj} as the isotropic total variation $\mathcal{R}(\alpha_\theta) = \int_\Omega \| \nabla \alpha_\theta \|$~\cite{rudin1992nonlinear}, consistent with the piecewise-constant defect structure of the effective-diffusivity model in Appendix~\ref{appdx:effective}. The full architecture, the cosine annealing schedule, and the discrete TV form are given in Appendix~\ref{appdx:method_arch}.

\subsection{Hard-constrained differentiable physics}
\label{sec:hardphys}

We discretize the bulk equation $\partial_t T = \nabla \cdot (\alpha \nabla T)$ on a uniform Cartesian grid covering $\Omega$ with spacings $\Delta x, \Delta y, \Delta z$ along the three axes and $N_t$ time steps of size $\Delta t$ aligned with the camera frame rate. Let $\mathbf{T}^n \in \mathbb{R}^{N_g}$ denote the discrete temperature at step $n$ on the $N_g$ grid nodes, and let $\bar{\alpha}_{i+1/2}$ denote the harmonic-mean face-centered diffusivity from Proposition~\ref{prop:harmonic}. The second-order finite-volume diffusion operator $\mathbf{L}(\alpha_\theta) \in \mathbb{R}^{N_g \times N_g}$ acts at an interior node by
\begin{equation}
\label{eq:fv_stencil}
    {[\mathbf{L}(\alpha_\theta) \mathbf{T}]}_{i,j,k} \;=\; \frac{1}{\Delta x^2} \!\left[ \bar{\alpha}_{i+1/2}(T_{i+1,j,k} - T_{i,j,k}) - \bar{\alpha}_{i-1/2}(T_{i,j,k} - T_{i-1,j,k}) \right] + (\text{$y$-, $z$-terms}),
\end{equation}
with lateral periodic and through-thickness adiabatic conditions (Appendix~\ref{appdx:bc}) closing the stencil at the boundary. The harmonic-mean coefficient is the unique discrete realization of effective-flux continuity at interior faces for the single-$\alpha$ model (Proposition~\ref{prop:harmonic}), whereas an arithmetic mean would spuriously allow heat to leak across the interface~\cite{patankar1980numerical}. All discrete temperatures here are interpreted as deviations from the ambient temperature (Appendix~\ref{appdx:bc}), so the implicit-Euler step is affine-free.

To decouple the time step from the diffusivity contrast, we integrate in time with the unconditionally stable implicit Euler scheme~\cite{courant1928partiellen}, advancing $\mathbf{T}^n$ to $\mathbf{T}^{n+1}$ through the sparse linear system
\begin{equation}
\label{eq:state_eq}
    \mathbf{A}(\alpha_\theta) \, \mathbf{T}^{n+1} \;=\; \mathbf{T}^n, \qquad \mathbf{A}(\alpha_\theta) \;:=\; \mathbf{I} - \Delta t \, \mathbf{L}(\alpha_\theta).
\end{equation}
The matrix $\mathbf{A}(\alpha_\theta)$ is sparse, symmetric positive-definite for our boundary conditions, and depends on $\theta$ only through $\alpha_\theta$. We solve~\eqref{eq:state_eq} on the GPU through a fixed number $K$ of unrolled Jacobi iterations, expressed as stencil convolutions within standard autodiff frameworks~\cite{paszke2019pytorch}. The full discretization, choice of $K$, and Jacobi update rule are given in Appendix~\ref{appdx:method_solver}. Solver validation and the distinction between numerical and inversion error are reported in Appendix~\ref{appdx:validation}.

\subsection{Adjoint optimization}
\label{sec:adjoint}

Composing the implicit Euler recurrence~\eqref{eq:state_eq} from $\mathbf{T}^0$ defines $\mathbf{T}^n = \mathbf{T}^n(\alpha_\theta)$ implicitly, hence the discrete analogue $\mathcal{K}_{\mathrm{disc}}$ of the parameter-to-observation map. Let $n_i \in \{1, \dots, N_t\}$ be chosen so that $t_{n_i} = t_i$ for each measurement time $t_i$, and let $\boldsymbol{\Pi}_\Gamma : \mathbb{R}^{N_g} \to \mathbb{R}^{N_\Gamma}$ denote the row-selector matrix that extracts the $N_\Gamma$ surface samples on $\Gamma_{\mathrm{obs}}$. NeFTY minimizes the discrete reconstruction objective
\begin{equation}
\label{eq:nefty_obj}
    \min_{\theta} \;\; \mathcal{J}_{\mathrm{disc}}(\theta) \;=\; \sum_{i=1}^{M} \big\| \boldsymbol{\Pi}_\Gamma \mathbf{T}^{n_i}(\alpha_\theta) - \hat{\mathbf{T}}_i \big\|^2 \;+\; \lambda \, \mathcal{R}(\alpha_\theta),
\end{equation}
the discretization of~\eqref{eq:ihcp_obj} with $\| \cdot \|$ the Euclidean norm on $\mathbb{R}^{N_\Gamma}$ and $\mathcal{R}$ the discrete total-variation regularizer of Appendix~\ref{appdx:method_arch}. Because $\mathbf{T}^n$ is defined implicitly by the state equation~\eqref{eq:state_eq}, gradients $d \mathcal{J}_{\mathrm{disc}} / d \theta$ obtained by backpropagation through the $K$ unrolled Jacobi iterates of every step cost $\mathcal{O}(K N_g N_t)$ memory, which is prohibitive for the $N_g \approx 10^5$, $N_t \approx 10^2$ regime of pulsed thermography. We use the discrete adjoint method instead~\cite{cea1986conception,onken2020discretize}.

Treat~\eqref{eq:state_eq} as the constraint $\mathbf{F}^n(\mathbf{T}^n, \mathbf{T}^{n-1}, \alpha_\theta) := \mathbf{A}(\alpha_\theta) \mathbf{T}^n - \mathbf{T}^{n-1} = \mathbf{0}$ for $n = 1, \dots, N_t$, and introduce adjoint variables $\boldsymbol{\mu}^n$ as Lagrange multipliers for each step. Setting the total derivative of the augmented Lagrangian with respect to $\mathbf{T}^n$ to zero yields the backward-in-time recurrence
\begin{equation}
\label{eq:adjoint_eq}
    \mathbf{A}(\alpha_\theta)^\top \boldsymbol{\mu}^n \;=\; \boldsymbol{\mu}^{n+1} + \big(\partial \ell^n / \partial \mathbf{T}^n\big)^\top, \qquad \boldsymbol{\mu}^{N_t + 1} \;=\; \mathbf{0},
\end{equation}
where $\ell^n$ collects the data-fidelity terms in~\eqref{eq:nefty_obj} that depend on $\mathbf{T}^n$ (nonzero only at measurement frames $n \in \{n_1, \dots, n_M\}$). The adjoint at step $n$ uses the transpose of the same sparse $\mathbf{A}$ that drives the forward solve, so a single backward sweep with the same Jacobi inner loop yields all $\boldsymbol{\mu}^n$ at the cost of one forward pass. The parameter gradient is then assembled by
\begin{equation}
\label{eq:grad_assembly}
    \frac{d \mathcal{J}_{\mathrm{disc}}}{d \theta} \;=\; \frac{\partial (\lambda \mathcal{R})}{\partial \theta} \;+\; \Delta t \, \sum_{n=1}^{N_t} (\boldsymbol{\mu}^n)^\top \, \frac{\partial \!\left( \mathbf{L}(\alpha_\theta) \mathbf{T}^n \right)}{\partial \alpha_\theta} \, \frac{\partial \alpha_\theta}{\partial \theta},
\end{equation}
where $\partial \alpha_\theta / \partial \theta$ is the MLP Jacobian. Because~\eqref{eq:adjoint_eq} requires only $\boldsymbol{\mu}^{n+1}$ and $\mathbf{T}^n$, the adjoint's solver working set is independent of $N_t$. Only the forward states $\mathbf{T}^n$ are stored. The full Lagrangian construction, the derivation of~\eqref{eq:adjoint_eq}--\eqref{eq:grad_assembly}, and the consistency with the continuous adjoint of the bulk equation are given in Appendix~\ref{appdx:method_adjoint}. The implementation hyperparameters (depth, width, encoding bandwidth $N$, grid resolution, Jacobi count $K$, and optimizer schedule) are listed in Appendix~\ref{appdx:method_hparams}. The data term depends directly on $\theta$ through the chain $\theta \to \alpha_\theta \to \mathbf{T}^n \to \mathcal{J}_{\mathrm{disc}}$, the dependency absent from the two-network formulation of Section~\ref{sec:pinn}.

\section{Experiments}
\label{sec:experiments}

\subsection{Setup}
\label{ssec:setup}

\textbf{Datasets.} Synthetic data are generated with PhiFlow~\cite{holl2024bf}, a finite-volume engine independent of the implicit-Euler Jacobi solver of Section~\ref{sec:hardphys}, so reconstruction is not subject to inverse crime. The real PVC datasets~\cite{wei2023pulsed,wei2023depth} provide only 2D defect masks and discrete defect depths from CAD, so real-world evaluation uses projected 2D masks and 2.5D depth annotations rather than 3D volumetric labels. The synthetic benchmark contains $1{,}000$ samples on the unitless slab $\Omega = [0,10]^2 \times [0,1]$ at $64 \times 64 \times 16$ resolution, $100$ frames at $\Delta t = 0.05$, and one to four subsurface defects per sample, split into a homogeneous and a layered configuration. Full data-generation procedure, dimensional analysis, and compute resources are in Appendix~\ref{appdx:experimental}.

\textbf{Baselines.} On synthetic data we compare to direct grid optimization (Grid Opt.; a voxel field through the same solver, adjoint, and TV regularizer, without the neural prior), classical L-BFGS voxel inversion with TV or $H^1$ regularization (Appendix~\ref{appdx:classical}), PINN~\cite{raissi2019physics} with GradNorm~\cite{chen2018gradnorm}, SPINN~\cite{cho2023spinn}, Causal-PINN~\cite{wang2024causal}, DCGD~\cite{hwang2024dcgd}, and a Sound-Only U-Net~\cite{ronneberger2015u} that trains on only defect-free samples and probes data-driven generalization to out-of-distribution defects. On real PVC data we additionally compare to the classical thermography heuristics PPT~\cite{maldague2002advances,ibarra2004ppt} and TSR~\cite{shepard2002reconstruction,shepard2015advances} and to a Swin-UNETR~\cite{hatamizadeh2022swin} trained on the synthetic benchmark and transferred zero-shot. Implementation details for every baseline are in Appendix~\ref{appdx:baselines}.

\textbf{Metrics.} Synthetic 3D reconstruction is evaluated by Mean Squared Error (MSE), Peak Signal-to-Noise Ratio (PSNR), Structural Similarity (SSIM)~\cite{wang2004image}, and volumetric Intersection over Union (IoU; defect mask $\alpha_\theta < 0.03$). Real PVC results are evaluated by 2D IoU and Dice for defect segmentation and by Absolute Relative error (Abs Rel), Root Mean Squared Error (RMSE), and the standard depth thresholds $\delta < 1.25^k$ ($k = 1, 2, 3$)~\cite{eigen2014depth}. Defect Edge F1 and the radial power spectrum of the recovered field, both targeting frequency-domain fidelity, are reported as diagnostics in Appendix~\ref{appdx:add_results}. Precise definitions for all metrics are catalogued in Appendix~\ref{appdx:experimental}.

\subsection{Synthetic 3D reconstruction}
\label{ssec:synth}

\begin{table*}[t]
\centering
\caption{\textbf{Label-free 3D thermal diffusivity reconstruction on synthetic $1{,}000$-sample benchmark} (mean $\pm$ 95\% CI). $\uparrow$ higher is better, $\downarrow$ lower is better. Best label-free entry per metric in \textbf{bold}. PINN~\cite{raissi2019physics}, Causal-PINN~\cite{wang2024causal}, and DCGD~\cite{hwang2024dcgd} $\alpha_{\theta}$ rows coincide because all three collapse to the same near-trivial constant. See Section~\ref{ssec:synth} and Appendix~\ref{appdx:pinn}.}
\label{tab:synth_main}
\begin{sc}
\resizebox{\textwidth}{!}{
\begin{tabular}{l cccc cccc}
\toprule
 & \multicolumn{4}{c}{\textbf{Homogeneous}} & \multicolumn{4}{c}{\textbf{Layered}} \\
\cmidrule(lr){2-5} \cmidrule(lr){6-9}
\textbf{Method} & MSE ($10^{-4}$) $\downarrow$ & PSNR $\uparrow$ & SSIM $\uparrow$ & IoU $\uparrow$ & MSE ($10^{-4}$) $\downarrow$ & PSNR $\uparrow$ & SSIM $\uparrow$ & IoU $\uparrow$ \\
\midrule
\multicolumn{9}{l}{\textit{Data-driven (zero-shot)}} \\
U-Net (Sound-Only)~\cite{ronneberger2015u} & $14.73 \pm 3.83$ & $14.83 \pm 0.71$ & $0.83 \pm 0.02$ & $0.00 \pm 0.00$ & $10.17 \pm 2.21$ & $15.42 \pm 0.98$ & $0.88 \pm 0.02$ & $0.00 \pm 0.00$ \\
\midrule
\multicolumn{9}{l}{\textit{Label-free}} \\
PINN~\cite{raissi2019physics,chen2018gradnorm} & $208.4 \pm 26.2$ & $-0.24 \pm 0.06$ & $0.04 \pm 0.00$ & $0.015 \pm 0.003$ & $200.4 \pm 17.8$ & $1.42 \pm 0.27$ & $0.04 \pm 0.00$ & $0.018 \pm 0.004$ \\
SPINN~\cite{cho2023spinn} & $32.71 \pm 10.02$ & $9.12 \pm 0.93$ & $0.45 \pm 0.06$ & $0.001 \pm 0.001$ & $33.84 \pm 8.57$ & $9.71 \pm 0.77$ & $0.53 \pm 0.07$ & $0.000 \pm 0.000$ \\
Causal-PINN~\cite{wang2024causal} & $208.4 \pm 26.2$ & $-0.24 \pm 0.06$ & $0.04 \pm 0.00$ & $0.015 \pm 0.003$ & $200.4 \pm 17.8$ & $1.42 \pm 0.27$ & $0.04 \pm 0.00$ & $0.018 \pm 0.004$ \\
DCGD~\cite{hwang2024dcgd} & $208.4 \pm 26.2$ & $-0.24 \pm 0.06$ & $0.04 \pm 0.00$ & $0.015 \pm 0.003$ & $200.4 \pm 17.8$ & $1.42 \pm 0.27$ & $0.04 \pm 0.00$ & $0.018 \pm 0.004$ \\
Grid Opt. & $12.61 \pm 4.20$ & $13.99 \pm 1.18$ & $0.56 \pm 0.04$ & $0.04 \pm 0.02$ & $15.01 \pm 2.97$ & $13.27 \pm 0.88$ & $0.57 \pm 0.04$ & $0.03 \pm 0.01$ \\
\textbf{NeFTY (Ours)} & $\mathbf{3.66 \pm 1.31}$ & $\mathbf{18.48 \pm 0.53}$ & $\mathbf{0.77 \pm 0.02}$ & $\mathbf{0.45 \pm 0.04}$ & $\mathbf{9.26 \pm 2.81}$ & $\mathbf{15.88 \pm 0.79}$ & $\mathbf{0.74 \pm 0.03}$ & $\mathbf{0.37 \pm 0.06}$ \\
\bottomrule
\end{tabular}
}
\end{sc}
 \vspace{-6pt}
\end{table*}

\begin{figure*}[t]
  \centering
  \includegraphics[width=0.96\linewidth]{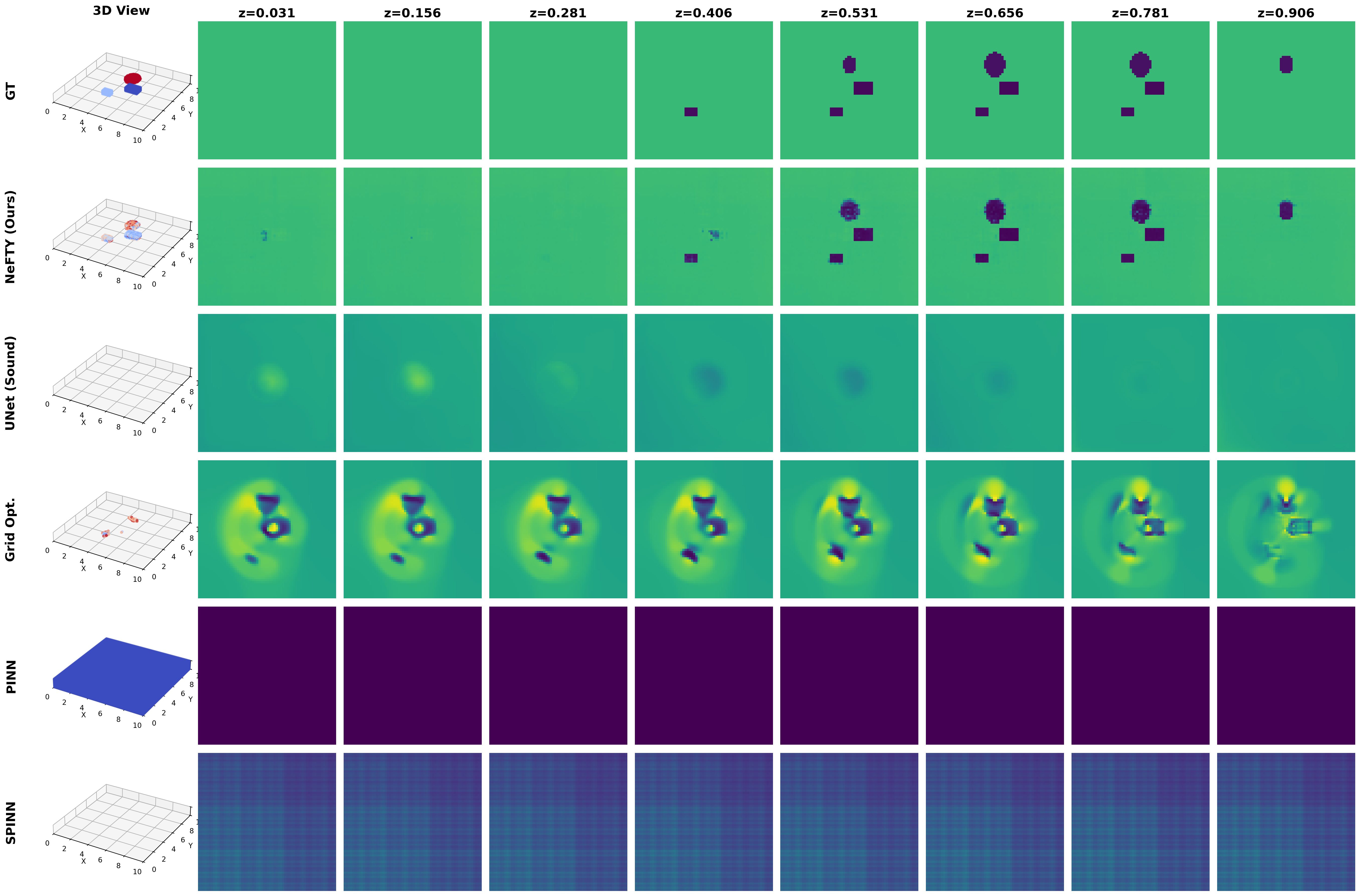}
  \captionsetup{belowskip=-4pt}
  \caption{\textbf{Homogeneous-setting reconstructions:} depth-wise $z$-slices show NeFTY isolating defect outlines at the correct depth while Grid Opt.\ yields a noisy field and PINN variants converge to near-uniform diffusivity.}
  \label{fig:qual_homo}
  \vspace{-6pt}
\end{figure*}

\begin{figure*}[t]
  \centering
  \includegraphics[width=0.96\linewidth]{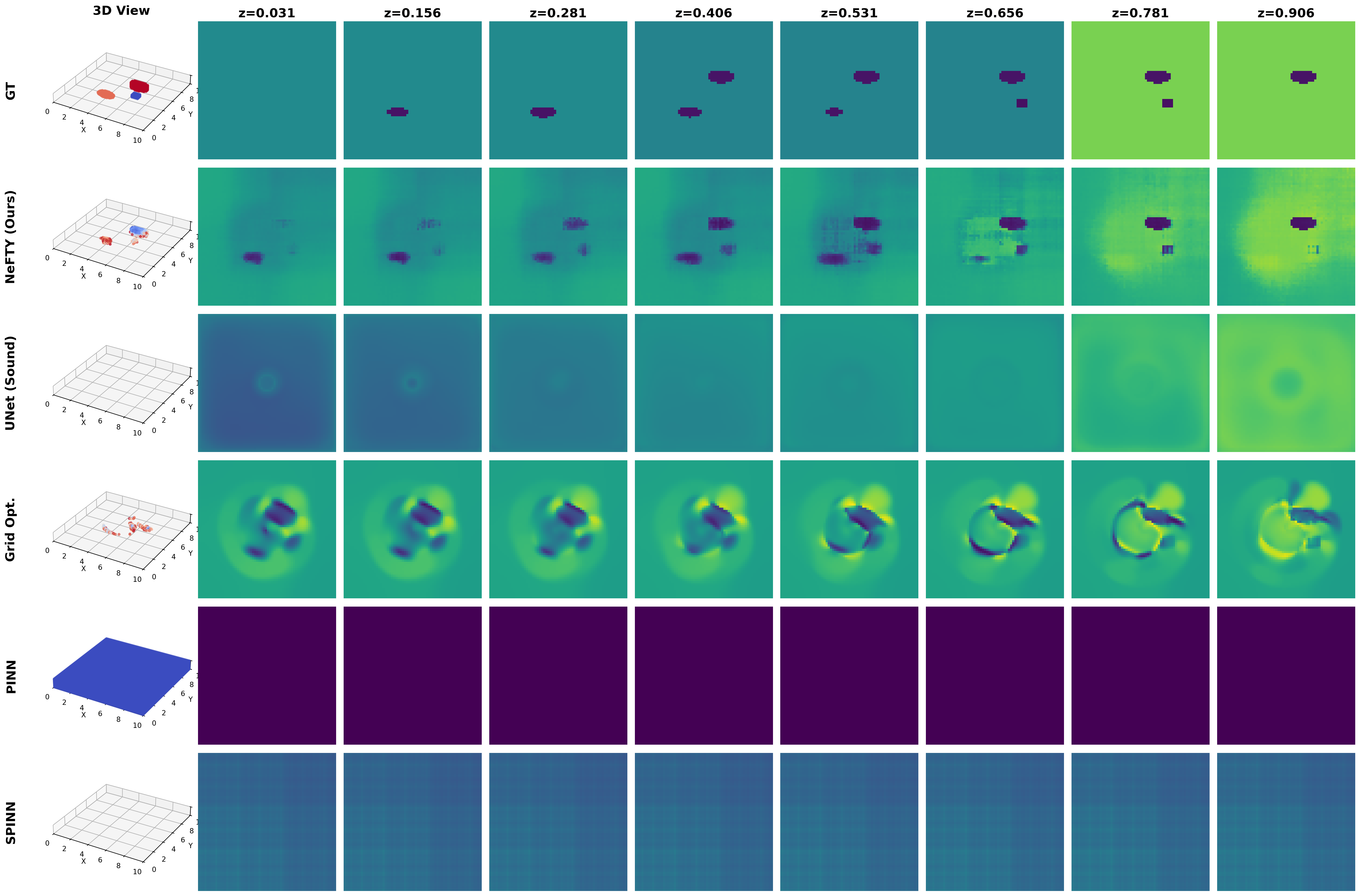}
  \captionsetup{belowskip=-4pt}
  \caption{\textbf{Layered-setting reconstructions:} NeFTY recovers defects against the stratified bulk, whereas Grid Opt.\ exhibits ringing artifacts and PINN variants again collapse to featureless fields.}
  \label{fig:qual_layered}
\end{figure*}

Table~\ref{tab:synth_main} reports the full benchmark. Figures~\ref{fig:qual_homo} and~\ref{fig:qual_layered} show representative qualitative reconstructions in the homogeneous and layered settings (per-defect-count and per-layer breakdowns are in Appendix~\ref{appdx:add_results}). Grid Opt.\ inverts the same hard PDE constraint as NeFTY but lacks a coordinate-based regularizer: it lowers the diffusivity MSE relative to every PINN variant yet caps IoU at $0.04$/$0.03$ with ringing artifacts at defect interfaces. The four soft-constrained PINN variants of Section~\ref{sec:pinn} fit the surface temperature observations through their $T_\phi$ network but fail volumetrically: their $\alpha_\theta$ networks saturate to near-trivial fields with IoU $\le 0.018$ (SPINN's separable surrogate lowers the homogeneous diffusivity MSE from $208.4$ to $32.71$ ($\times 10^{-4}$) without resolving defects), in line with the structural decoupling $\nabla_\theta \mathcal{L}_{\mathrm{data}} = 0$ proven in Appendix~\ref{appdx:pinn} (the bit-identical PINN, Causal-PINN, and DCGD $\alpha_\theta$ rows in Table~\ref{tab:synth_main} reflect this same collapse despite differing $T_\phi$ surrogates). NeFTY balances the physical consistency of the hard solver with the prior of a continuous neural field, reaching IoU $0.45$/$0.37$ and PSNR $18.48$/$15.88$\,dB.

\textbf{Data-fit paradox.} The evaluated PINN variants fit the surface observations while recovering little volumetric structure: vanilla PINN attains a surface PSNR of $\sim 63$\,dB on $T_\phi$ while its volumetric IoU is $0.01$, SPINN reaches $\sim 61$\,dB at IoU $0.000$, and NeFTY $\sim 82$\,dB at IoU $0.45$. Under the soft-constraint formulation of Eq.~\eqref{eq:pinn_loss}, surface observations influence the diffusivity only indirectly, through the residual term (Appendix~\ref{appdx:pinn_coupling}), and in this stiff parabolic regime the low surface error of the evaluated variants does not translate into correct volumetric structure. Frequency-domain diagnostics in Appendix~\ref{appdx:add_results} (Edge F1 $0.470 \pm 0.137$ for NeFTY versus $0.000$ for every PINN variant; full radial power spectrum) confirm that NeFTY is the only label-free method in Table~\ref{tab:synth_main} whose recovered field carries detectable defect-scale gradients.

\textbf{Classical PDE-constrained inversion.} We also optimize a bounded voxel diffusivity field with L-BFGS under TV or $H^1$ regularization through the same solver and adjoint, with weights tuned on a disjoint calibration set (Appendix~\ref{appdx:classical}). On 64 paired benchmark specimens these arms reach IoU $0.129$/$0.112$ (TV) and $0.125$/$0.127$ ($H^1$) on homogeneous/layered data, against $0.45$/$0.37$ for NeFTY and $0.04$/$0.03$ for Grid Opt.\ (Table~\ref{tab:synth_main}, full benchmark). On the same specimens (ratio of means), L-BFGS improves on Grid Opt.\ by about $3.2\times$ and the neural-field prior adds a further $3.3\times$, so among the evaluated configurations the representation carries the largest share of NeFTY's gain.

\textbf{Robustness.} On new irregular-geometry specimens, NeFTY reaches 3D IoU $0.316$ on irregular blobs and $0.222$ on curved strata interfaces, where Grid Opt.\ reaches $0.000$ (Appendix~\ref{appdx:irregular}). Under Gaussian observation noise at $0.5$, $1$, $2$, and $5\%$ of peak signal, NeFTY retains $96$, $82$, $64$, and $37\%$ of its noise-free IoU (Appendix~\ref{appdx:noise}). Across defect-to-bulk contrast from 1:5 to 1:50, IoU shows no detectable change (Appendix~\ref{appdx:contrast}).

\subsection{Real-world PVC validation}
\label{ssec:real}

\begin{figure*}[t]
\centering
\begin{minipage}[b]{0.38\linewidth}
\centering
\captionof{table}{\textbf{Real PVC evaluation:} NeFTY is the best label-free method on every metric for both 2D segmentation~\cite{wei2023pulsed} and 2.5D depth~\cite{wei2023depth}.}
\label{tab:real_pvc}
\vspace{2pt}
\begin{sc}
\footnotesize
\resizebox{\linewidth}{!}{
\begin{tabular}{l cc cc}
\toprule
 & \multicolumn{2}{c}{\textbf{2D segm.}} & \multicolumn{2}{c}{\textbf{2.5D depth}} \\
\cmidrule(lr){2-3} \cmidrule(lr){4-5}
\textbf{Method} & IoU $\uparrow$ & Dice $\uparrow$ & Abs Rel $\downarrow$ & $\delta < 1.25$ $\uparrow$ \\
\midrule
Swin-UNETR~\cite{hatamizadeh2022swin} & 0.085 & 0.151 & 3.131 & 0.052 \\
PPT~\cite{maldague2002advances} & 0.419 & 0.563 & 3.416 & 0.005 \\
TSR~\cite{shepard2002reconstruction} & 0.346 & 0.464 & 13.366 & 0.002 \\
SPINN~\cite{cho2023spinn} & 0.050 & 0.086 & 6.737 & 0.002 \\
Grid Opt. & 0.395 & 0.557 & 0.941 & 0.104 \\
\textbf{NeFTY (Ours)} & \textbf{0.431} & \textbf{0.597} & \textbf{0.465} & \textbf{0.385} \\
\bottomrule
\end{tabular}
}
\end{sc}
\end{minipage}\hfill
\begin{minipage}[b]{0.60\linewidth}
\centering
\includegraphics[width=\linewidth]{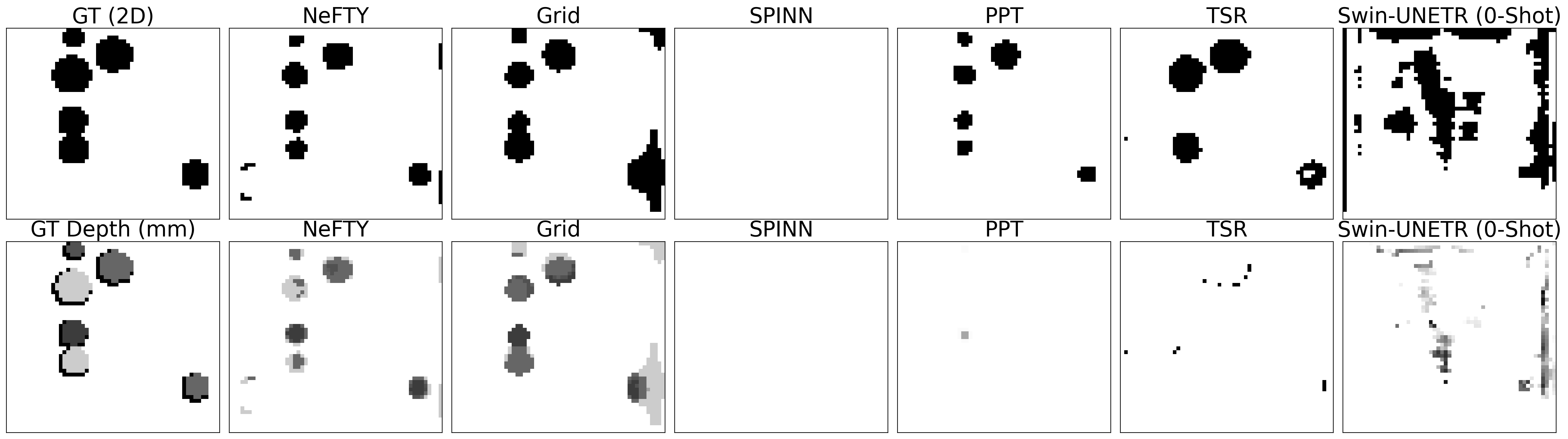}
\captionof{figure}{\textbf{Real PVC reconstruction:} NeFTY recovers both the defect layout (Row~1) and the depth ordering (Row~2) from a single surface IR sequence, while PPT/TSR yield coarse depth and SPINN collapses to a near-uniform field.}
\label{fig:real_world}
\end{minipage}
\end{figure*}

The same NeFTY solver is applied to the PVC datasets after the three physical adaptations of Appendix~\ref{appdx:bc} (near-uniform front-face flash initial condition, centimeter and second-scale Fourier-number rescaling, convective Robin back-face boundary), with the projection from $\hat{\alpha}_\theta$ to 2D masks and 2.5D depth documented in Appendix~\ref{appdx:pvc}. As reported in Table~\ref{tab:real_pvc}, NeFTY is the best label-free method on every metric, with $0.431$ IoU and $0.597$ Dice for 2D segmentation and $0.465$ Abs Rel and $0.385$ $\delta < 1.25$ accuracy for 2.5D depth. The classical heuristics PPT and TSR are competitive on segmentation but degrade sharply on depth, since their inversion formulas assume a semi-infinite one-dimensional geometry that does not hold on the finite PVC slab (Appendix~\ref{appdx:baselines_classical}). The supervised Swin-UNETR collapses under the synthetic-to-real domain shift, in contrast to NeFTY's per-specimen test-time optimization which adapts to the observed data without training labels (Figure~\ref{fig:real_world}; full depth-metric breakdown in Appendix~\ref{appdx:real_pvc_depth_full}). These projected 2D and 2.5D metrics validate segmentation and depth recovery under real acquisition. The full volumetric claim rests on the synthetic benchmark with exact 3D ground truth (Section~\ref{ssec:synth}).

\begin{table*}[t]
\centering
\begin{minipage}[t]{0.36\linewidth}
\centering
\captionof{table}{\textbf{Ablation of NeFTY:} each row adds one component (H/L = homogeneous/Layered).}
\label{tab:ablation_main}
\vspace{2pt}
\begin{sc}
\footnotesize
\resizebox{\linewidth}{!}{
\begin{tabular}{l c c}
\toprule
\textbf{Variant} & \textbf{PSNR} (H/L) $\uparrow$ & \textbf{IoU} (H/L) $\uparrow$ \\
\midrule
Base & $\phantom{0}0.47$ / $\phantom{0}2.89$ & $0.03$ / $0.02$ \\
\,+\,PE & $\phantom{0}4.17$ / $\phantom{0}6.15$ & $0.09$ / $0.09$ \\
\,+\,PE,\,FA & $\phantom{0}8.33$ / $\phantom{0}8.81$ & $0.14$ / $0.14$ \\
\,+\,PE,\,FA,\,$\sigma$ & $\phantom{0}9.19$ / $\phantom{0}9.27$ & $0.18$ / $0.14$ \\
\,+\,PE,\,FA,\,$\sigma$,\,HM & $10.95$ / $11.26$ & $0.24$ / $0.22$ \\
\textbf{Full (\,+\,TV)} & $\mathbf{18.48}$ / $\mathbf{15.88}$ & $\mathbf{0.45}$ / $\mathbf{0.37}$ \\
\bottomrule
\end{tabular}
}
\end{sc}
\end{minipage}\hfill
\begin{minipage}[t]{0.62\linewidth}
\centering
\captionof{table}{\textbf{Solver-level efficiency benchmark:} our discrete adjoint method (AM) matches the simulation accuracy of autograd (AD) while reducing peak GPU memory by orders of magnitude on a single $64\!\times\!64\!\times\!16$ sample over $50$ time steps.}
\label{tab:efficiency_main}
\vspace{2pt}
\begin{sc}
\footnotesize
\resizebox{\linewidth}{!}{
\begin{tabular}{l c c c c}
\toprule
\textbf{Method} & \textbf{Fwd Time (s)} $\downarrow$ & \textbf{Bwd Time (s)} $\downarrow$ & \textbf{Peak Mem} $\downarrow$ & \textbf{Sim.\ Error} $\downarrow$ \\
\midrule
PhiFlow (Ex) & $26.36 \pm 0.10$ & $0.87 \pm 0.01$ & $3.26$\,GB & $7.91 \times 10^{-4}$ \\
PhiFlow (Im) & $\phantom{0}3.26 \pm 0.40$ & $0.76 \pm 0.01$ & $275.5$\,MB & $8.47 \times 10^{-4}$ \\
\midrule
Ours (AD) & $\phantom{0}1.43 \pm 0.04$ & $1.30 \pm 0.02$ & $18.63$\,GB & $\mathbf{3.73 \times 10^{-8}}$ \\
\textbf{Ours (AM)} & $\mathbf{0.46 \pm 0.00}$ & $\mathbf{0.50 \pm 0.00}$ & $\mathbf{21.9}$\,\textbf{MB} & $\mathbf{3.73 \times 10^{-8}}$ \\
\bottomrule
\end{tabular}
}
\end{sc}
\end{minipage}
\end{table*}

\subsection{Ablations}
\label{ssec:ablation}

Table~\ref{tab:ablation_main} traces the contribution of each NeFTY component. Positional encoding (PE) and frequency annealing (FA) push IoU from $0.03$ to $0.14$ by stabilizing high-frequency optimization against the noise amplification of Corollary~\ref{cor:hadamard}. The bounded sigmoid output ($\sigma$) prevents drift into the ill-conditioned regime $\alpha_\theta \to 0$, and the harmonic-mean coefficient (HM, Proposition~\ref{prop:harmonic}) restores effective-flux continuity at high-contrast interfaces. The TV regularizer doubles IoU again to $0.45$/$0.37$ by aligning the prior with the piecewise-constant defect structure of the effective-diffusivity model. The full ablation with $95\%$ CIs and a per-difficulty breakdown is in Appendix~\ref{appdx:add_results}.

\subsection{Computational efficiency}
\label{ssec:efficiency}

Table~\ref{tab:efficiency_main} reports a per-step solver benchmark on a single specimen. The discrete adjoint method (AM) of Section~\ref{sec:adjoint} matches the simulation accuracy of standard Autograd (AD) bit-for-bit while reducing peak GPU memory by three orders of magnitude (from $18.63$\,GB to $21.9$\,MB) and halving the backward time, in agreement with the solver-level $\mathcal{O}(N_g)$ versus $\mathcal{O}(K N_g N_t)$ analysis of Appendix~\ref{appdx:method_adjoint}. The same AM forward pass is also $\sim\!7\times$ faster than PhiFlow's implicit solver. The training-level wall-clock comparison across all baselines (NeFTY converges in $9.6$\,min versus $18.2$-$58.3$\,min for the soft-PINN family on the same hardware) is in Table~\ref{tab:training_benchmark} of Appendix~\ref{appdx:add_results}. Up to $128 \times 128 \times 32$ grids and $200$ frames, peak memory scales with exponents $0.985$ in voxels and $0.694$ in frames, with a near-flat marginal cost per additional frame (Appendix~\ref{appdx:scaling}).

\section{Conclusion}

We presented NeFTY, a hard-constrained neural field framework for three-dimensional inverse heat conduction. NeFTY parameterizes the unknown thermal diffusivity as a continuous coordinate-based field, enforces the governing PDE through a differentiable implicit-Euler solver with harmonic-mean interface flux, and inverts the resulting forward map with adjoint gradients at solver-level memory cost. On synthetic homogeneous and layered settings it reaches IoU $0.45$ and $0.37$, while four soft-constrained PINN variants and a voxel-grid baseline remain at or near zero and classical L-BFGS voxel inversion through the same solver trails by about $3.3\times$ on the same specimens. On real PVC pulsed-thermography data it segments defects at IoU $0.431$ and recovers depth at Abs Rel $0.465$ from surface measurements alone.

\textbf{Limitations and Broader Impact.} NeFTY shares the limitations common to test-time inverse-problem solvers: per-specimen optimization ($9.6$\,min and $4.3$\,GB on one workstation GPU, suited to offline inspection), attenuated quantitative recovery inside very high-contrast voids, and evaluation against projected 2D and 2.5D supervision because public NDT benchmarks do not yet provide 3D volumetric ground truth. Future directions include amortized initialization through meta-learning or hypernetworks, preconditioned solvers in the differentiable loop, and paired CT and material-characterization labels for full volumetric evaluation (Appendix~\ref{appdx:limitations}). We primarily target defects that form discrete regions in a host material, imaged with the same one-sided, non-contact acquisition as PPT and TSR. Smoothly varying fields, non-Cartesian geometries, anisotropy, and temperature-dependent coefficients are important future scope. The intended use of NeFTY is non-destructive evaluation, where it can support safer infrastructure inspection and earlier defect detection without destructive testing or ionizing X-ray exposure. As with any automated inspection pipeline, its outputs are best treated as decision support for human qualification of safety-critical components, and reconstructions in regions of low surface sensitivity should be reported with uncertainty alongside the recovered field.

\begin{ack}
This project is partially supported by the AI Lab Seed Grant from the Princeton Laboratory for Artificial Intelligence. The authors declare no competing interests.
\end{ack}

{\small
\bibliographystyle{plainnat}
\bibliography{main}

\begin{thebibliography}{73}
\providecommand{\natexlab}[1]{#1}
\providecommand{\url}[1]{\texttt{#1}}
\expandafter\ifx\csname urlstyle\endcsname\relax
  \providecommand{\doi}[1]{doi: #1}\else
  \providecommand{\doi}{doi: \begingroup \urlstyle{rm}\Url}\fi

\bibitem[Ali et~al.(2025)Ali, Addepalli, and Zhao]{ali2025effective}
Zain Ali, Sri Addepalli, and Yifan Zhao.
\newblock Effective thermal diffusivity measurement using through-transmission pulsed thermography: Extending the current practice by incorporating multi-parameter optimisation.
\newblock \emph{Sensors}, 25\penalty0 (4):\penalty0 1139, 2025.

\bibitem[Alifanov(1994)]{alifanov1994inverse}
O~M Alifanov.
\newblock \emph{Inverse Heat Transfer Problems}.
\newblock International Series in Heat and Mass Transfer. Springer-Verlag, 1994.

\bibitem[Bouziani et~al.(2024)Bouziani, Ham, and Farsi]{bouziani2024differentiable}
Nacime Bouziani, David~A Ham, and Ado Farsi.
\newblock Differentiable programming across the pde and machine learning barrier.
\newblock \emph{arXiv preprint arXiv:2409.06085}, 2024.

\bibitem[Burgholzer et~al.(2017)Burgholzer, Thor, Gruber, and Mayr]{burgholzer2017three}
Peter Burgholzer, Michael Thor, J{\"u}rgen Gruber, and G{\"u}nther Mayr.
\newblock Three-dimensional thermographic imaging using a virtual wave concept.
\newblock \emph{Journal of Applied Physics}, 121\penalty0 (10), 2017.

\bibitem[Burgholzer et~al.(2018)Burgholzer, Stockner, and Mayr]{burgholzer2018acoustic}
Peter Burgholzer, Gregor Stockner, and Guenther Mayr.
\newblock Acoustic reconstruction for photothermal imaging.
\newblock \emph{Bioengineering}, 5\penalty0 (3):\penalty0 70, 2018.

\bibitem[Cai et~al.(2021)Cai, Wang, Wang, Perdikaris, and Karniadakis]{cai2021physics}
Shengze Cai, Zhicheng Wang, Sifan Wang, Paris Perdikaris, and George~Em Karniadakis.
\newblock Physics-informed neural networks for heat transfer problems.
\newblock \emph{Journal of Heat Transfer}, 143\penalty0 (6):\penalty0 060801, 2021.

\bibitem[C{\'e}a(1986)]{cea1986conception}
Jean C{\'e}a.
\newblock Conception optimale ou identification de formes, calcul rapide de la d{\'e}riv{\'e}e directionnelle de la fonction co{\^u}t.
\newblock \emph{ESAIM: Mod{\'e}lisation math{\'e}matique et analyse num{\'e}rique}, 20\penalty0 (3):\penalty0 371--402, 1986.

\bibitem[Chen et~al.(2018)Chen, Badrinarayanan, Lee, and Rabinovich]{chen2018gradnorm}
Zhao Chen, Vijay Badrinarayanan, Chen-Yu Lee, and Andrew Rabinovich.
\newblock Gradnorm: Gradient normalization for adaptive loss balancing in deep multitask networks.
\newblock In \emph{International conference on machine learning}, pages 794--803. PMLR, 2018.

\bibitem[Cho et~al.(2023)Cho, Nam, Yang, Yun, Hong, and Park]{cho2023spinn}
Junwoo Cho, Seungtae Nam, Hyunmo Yang, Seok-Bae Yun, Youngjoon Hong, and Eunbyung Park.
\newblock Separable physics-informed neural networks.
\newblock \emph{Advances in Neural Information Processing Systems}, 36:\penalty0 23761--23788, 2023.

\bibitem[Chung et~al.(2021)Chung, Lee, and Kim]{chung2021latest}
Yoonjae Chung, Seungju Lee, and Wontae Kim.
\newblock Latest advances in common signal processing of pulsed thermography for enhanced detectability: A review.
\newblock \emph{Applied Sciences}, 11\penalty0 (24):\penalty0 12168, 2021.

\bibitem[Cobos et~al.(2022)Cobos, Norley, Pollmann, and Holdsworth]{cobos2022cost}
S~F Cobos, C~J Norley, S~I Pollmann, and D~W Holdsworth.
\newblock Cost-effective micro-{CT} system for non-destructive testing of titanium {3D} printed medical components.
\newblock \emph{PLOS ONE}, 17\penalty0 (10):\penalty0 e0275732, 2022.

\bibitem[Courant et~al.(1928)Courant, Friedrichs, and Lewy]{courant1928partiellen}
Richard Courant, Kurt Friedrichs, and Hans Lewy.
\newblock {\"U}ber die partiellen differenzengleichungen der mathematischen physik.
\newblock \emph{Mathematische annalen}, 100\penalty0 (1):\penalty0 32--74, 1928.

\bibitem[de~Avila Belbute-Peres et~al.(2018)de~Avila Belbute-Peres, Smith, Allen, Tenenbaum, and Kolter]{de2018end}
Filipe de~Avila Belbute-Peres, Kevin Smith, Kelsey Allen, Josh Tenenbaum, and J~Zico Kolter.
\newblock End-to-end differentiable physics for learning and control.
\newblock \emph{Advances in neural information processing systems}, 31, 2018.

\bibitem[Degrave et~al.(2019)Degrave, Hermans, Dambre, and Wyffels]{degrave2019differentiable}
Jonas Degrave, Michiel Hermans, Joni Dambre, and Francis Wyffels.
\newblock A differentiable physics engine for deep learning in robotics.
\newblock \emph{Frontiers in neurorobotics}, 13:\penalty0 6, 2019.

\bibitem[Eigen et~al.(2014)Eigen, Puhrsch, and Fergus]{eigen2014depth}
David Eigen, Christian Puhrsch, and Rob Fergus.
\newblock Depth map prediction from a single image using a multi-scale deep network.
\newblock \emph{Advances in neural information processing systems}, 27, 2014.

\bibitem[Engl et~al.(1996)Engl, Hanke, and Neubauer]{engl1996regularization}
Heinz~Werner Engl, Martin Hanke, and Andreas Neubauer.
\newblock \emph{Regularization of inverse problems}, volume 375.
\newblock Springer Science \& Business Media, 1996.

\bibitem[Fang et~al.(2023)Fang, Ibarra-Castanedo, Garrido, Duan, and Maldague]{fang2023automatic}
Qiang Fang, Clemente Ibarra-Castanedo, Iv{\'a}n Garrido, Yuxia Duan, and Xavier Maldague.
\newblock Automatic detection and identification of defects by deep learning algorithms from pulsed thermography data.
\newblock \emph{Sensors}, 23\penalty0 (9):\penalty0 4444, 2023.

\bibitem[Gahleitner et~al.(2024)Gahleitner, Thummerer, Plank, Wiedemann, Mayr, H{\"u}hne, Burgholzer, and Cakmak]{gahleitner2024photothermal}
Lukas Gahleitner, Gregor Thummerer, Bernhard Plank, Johannes Wiedemann, G{\"u}nther Mayr, Christian H{\"u}hne, Peter Burgholzer, and Umut Cakmak.
\newblock Photothermal defect imaging in hybrid fiber metal laminates using the virtual wave concept.
\newblock \emph{Journal of Applied Physics}, 135\penalty0 (7), 2024.

\bibitem[Garcea et~al.(2018)Garcea, Wang, and Withers]{garcea2018x}
Serafina~C Garcea, Ying Wang, and Philip~J Withers.
\newblock X-ray computed tomography of polymer composites.
\newblock \emph{Composites Science and Technology}, 156:\penalty0 305--319, 2018.

\bibitem[Glorot and Bengio(2010)]{glorot2010understanding}
Xavier Glorot and Yoshua Bengio.
\newblock Understanding the difficulty of training deep feedforward neural networks.
\newblock In \emph{Proceedings of the thirteenth international conference on artificial intelligence and statistics}, pages 249--256. JMLR Workshop and Conference Proceedings, 2010.

\bibitem[Haber et~al.(2000)Haber, Ascher, and Oldenburg]{haber2000optimization}
E~Haber, U~M Ascher, and D~Oldenburg.
\newblock On optimization techniques for solving nonlinear inverse problems.
\newblock \emph{Inverse Problems}, 16\penalty0 (5):\penalty0 1263--1280, 2000.

\bibitem[Hadamard(1902)]{hadamard1888rayon}
Jacques Hadamard.
\newblock Sur les probl{\`e}mes aux d{\'e}riv{\'e}es partielles et leur signification physique.
\newblock \emph{Princeton university bulletin}, pages 49--52, 1902.

\bibitem[Hao et~al.(2024)Hao, Braga-Neto, Liu, Wang, and Zhong]{hao2024training}
Baoli Hao, Ulisses Braga-Neto, Chun Liu, Lifan Wang, and Ming Zhong.
\newblock Training pinns with hard constraints and adaptive weights: An ablation study.
\newblock \emph{arXiv preprint arXiv:2404.16189}, 2024.

\bibitem[Hatamizadeh et~al.(2021)Hatamizadeh, Nath, Tang, Yang, Roth, and Xu]{hatamizadeh2022swin}
Ali Hatamizadeh, Vishwesh Nath, Yucheng Tang, Dong Yang, Holger~R Roth, and Daguang Xu.
\newblock Swin unetr: Swin transformers for semantic segmentation of brain tumors in mri images.
\newblock In \emph{International MICCAI brainlesion workshop}, pages 272--284. Springer, 2021.

\bibitem[Hinze et~al.(2009)Hinze, Pinnau, Ulbrich, and Ulbrich]{hinze2009optimization}
M~Hinze, R~Pinnau, M~Ulbrich, and S~Ulbrich.
\newblock \emph{Optimization with {PDE} Constraints}.
\newblock Mathematical Modelling: Theory and Applications. Springer, 2009.

\bibitem[Holl and Thuerey(2024)]{holl2024bf}
Philipp Holl and Nils Thuerey.
\newblock $\phi$-flow: Differentiable simulations for pytorch, tensorflow and jax.
\newblock In \emph{Proceedings of the Forty-first International Conference on Machine Learning}, 2024.

\bibitem[Holl et~al.(2020)Holl, Koltun, and Thuerey]{holl2020learning}
Philipp Holl, Vladlen Koltun, and Nils Thuerey.
\newblock Learning to control pdes with differentiable physics.
\newblock \emph{arXiv preprint arXiv:2001.07457}, 2020.

\bibitem[Hwang and Lim(2024)]{hwang2024dcgd}
Youngsik Hwang and Dong-Young Lim.
\newblock Dual cone gradient descent for training physics-informed neural networks.
\newblock \emph{Advances in Neural Information Processing Systems}, 37:\penalty0 98563--98595, 2024.

\bibitem[Ibarra-Castanedo and Maldague(2004)]{ibarra2004ppt}
Clemente Ibarra-Castanedo and Xavier Maldague.
\newblock Pulsed phase thermography reviewed.
\newblock \emph{Quantitative Infrared Thermography Journal}, 1\penalty0 (1):\penalty0 47--70, 2004.

\bibitem[Jarny et~al.(1991)Jarny, {\"O}zi{\c{s}}ik, and Bardon]{jarny1991general}
Y~Jarny, M~N {\"O}zi{\c{s}}ik, and J~P Bardon.
\newblock A general optimization method using adjoint equation for solving multidimensional inverse heat conduction.
\newblock \emph{International Journal of Heat and Mass Transfer}, 34\penalty0 (11):\penalty0 2911--2919, 1991.

\bibitem[Kelly and Thurow(2023)]{kelly2023fluidnerf}
Dustin~L Kelly and Brian~S Thurow.
\newblock Fluidnerf: A scalar-field reconstruction technique for flow diagnostics using neural radiance fields.
\newblock In \emph{AIAA SciTech 2023 Forum}, page 0412, 2023.

\bibitem[Kov{\'a}cs et~al.(2020)Kov{\'a}cs, Lehner, Thummerer, Mayr, Burgholzer, and Huemer]{kovacs2020deep}
P{\'e}ter Kov{\'a}cs, Bernhard Lehner, Gregor Thummerer, G{\"u}nther Mayr, Peter Burgholzer, and Mario Huemer.
\newblock Deep learning approaches for thermographic imaging.
\newblock \emph{Journal of Applied Physics}, 128\penalty0 (15), 2020.

\bibitem[Krishnapriyan et~al.(2021)Krishnapriyan, Gholami, Zhe, Kirby, and Mahoney]{krishnapriyan2021characterizing}
A~S Krishnapriyan, A~Gholami, S~Zhe, R~M Kirby, and M~W Mahoney.
\newblock Characterizing possible failure modes in physics-informed neural networks.
\newblock \emph{Advances in Neural Information Processing Systems}, pages 26548--26560, 2021.

\bibitem[Leontiou et~al.(2024)Leontiou, Frixou, Charalambides, Stiliaris, Papanicolas, Nikolaidou, and Papadakis]{leontiou2024three}
Theodoros Leontiou, Anna Frixou, Marios Charalambides, Efstathios Stiliaris, Costas~N Papanicolas, Sofia Nikolaidou, and Antonis Papadakis.
\newblock Three-dimensional thermal tomography with physics-informed neural networks.
\newblock \emph{Tomography}, 10\penalty0 (12):\penalty0 1930, 2024.

\bibitem[Liu et~al.(2025)Liu, Yao, Wang, Sfarra, and Liu]{liu2025review}
Y~Liu, Y~Yao, F~Wang, S~Sfarra, and K~Liu.
\newblock Review of unsupervised machine learning methods in active infrared thermography for defect detection and analysis.
\newblock \emph{Quantitative InfraRed Thermography Journal}, 23\penalty0 (3):\penalty0 223--250, 2025.

\bibitem[Lozenski et~al.(2024)Lozenski, Cam, Pagel, Anastasio, and Villa]{lozenski2024proxnf}
Luke Lozenski, Refik~Mert Cam, Mark~D Pagel, Mark~A Anastasio, and Umberto Villa.
\newblock Proxnf: neural field proximal training for high-resolution 4d dynamic image reconstruction.
\newblock \emph{IEEE Transactions on Computational Imaging}, 2024.

\bibitem[Ma et~al.(2025)Ma, Sun, and Zhang]{ma2025quantitative}
Botao Ma, Shupeng Sun, and Lin Zhang.
\newblock Quantitative depth estimation in lock-in thermography: Modeling and correction of lateral heat conduction effects.
\newblock \emph{Materials}, 18\penalty0 (22):\penalty0 5247, 2025.

\bibitem[Maldague et~al.(2002)Maldague, Galmiche, and Ziadi]{maldague2002advances}
Xavier Maldague, Fran{\c{c}}ois Galmiche, and Adel Ziadi.
\newblock Advances in pulsed phase thermography.
\newblock \emph{Infrared physics \& technology}, 43\penalty0 (3-5):\penalty0 175--181, 2002.

\bibitem[Martinez~Mundarain(2024)]{martinez2024artificial}
Andrea~Carolina Martinez~Mundarain.
\newblock Artificial neural networks as the solution of inverse heat conduction problems in multidimensional domains.
\newblock 2024.

\bibitem[Mildenhall et~al.(2021)Mildenhall, Srinivasan, Tancik, Barron, Ramamoorthi, and Ng]{mildenhall2021nerf}
Ben Mildenhall, Pratul~P Srinivasan, Matthew Tancik, Jonathan~T Barron, Ravi Ramamoorthi, and Ren Ng.
\newblock Nerf: Representing scenes as neural radiance fields for view synthesis.
\newblock \emph{Communications of the ACM}, 65\penalty0 (1):\penalty0 99--106, 2021.

\bibitem[Nair and Hinton(2010)]{nair2010rectified}
Vinod Nair and Geoffrey~E Hinton.
\newblock Rectified linear units improve restricted boltzmann machines.
\newblock In \emph{Proceedings of the 27th international conference on machine learning (ICML-10)}, pages 807--814, 2010.

\bibitem[Oliveira et~al.(2021)Oliveira, Seibert, Borges, Albertazzi, and Schmitt]{oliveira2021employing}
BCF Oliveira, AA~Seibert, VK~Borges, A~Albertazzi, and RH~Schmitt.
\newblock Employing a u-net convolutional neural network for segmenting impact damages in optical lock-in thermography images of cfrp plates.
\newblock \emph{Nondestructive Testing and Evaluation}, 36\penalty0 (4):\penalty0 440--458, 2021.

\bibitem[Onken and Ruthotto(2020)]{onken2020discretize}
Derek Onken and Lars Ruthotto.
\newblock Discretize-optimize vs. optimize-discretize for time-series regression and continuous normalizing flows.
\newblock \emph{arXiv preprint arXiv:2005.13420}, 2020.

\bibitem[{\"O}zi{\c{s}}ik and Orlande(2021)]{ozisik2021inverse}
M~N {\"O}zi{\c{s}}ik and H~R~B Orlande.
\newblock \emph{Inverse Heat Transfer: Fundamentals and Applications}.
\newblock CRC Press, 2nd edition, 2021.

\bibitem[Park et~al.(2021)Park, Sinha, Barron, Bouaziz, Goldman, Seitz, and Martin-Brualla]{park2021nerfies}
Keunhong Park, Utkarsh Sinha, Jonathan~T Barron, Sofien Bouaziz, Dan~B Goldman, Steven~M Seitz, and Ricardo Martin-Brualla.
\newblock Nerfies: Deformable neural radiance fields.
\newblock In \emph{Proceedings of the IEEE/CVF international conference on computer vision}, pages 5865--5874, 2021.

\bibitem[Paszke et~al.(2019)Paszke, Gross, Massa, Lerer, Bradbury, Chanan, Killeen, Lin, Gimelshein, Antiga, et~al.]{paszke2019pytorch}
Adam Paszke, Sam Gross, Francisco Massa, Adam Lerer, James Bradbury, Gregory Chanan, Trevor Killeen, Zeming Lin, Natalia Gimelshein, Luca Antiga, et~al.
\newblock Pytorch: An imperative style, high-performance deep learning library.
\newblock \emph{Advances in neural information processing systems}, 32, 2019.

\bibitem[Patankar(1980)]{patankar1980numerical}
S.~Patankar.
\newblock \emph{Numerical Heat Transfer and Fluid Flow}.
\newblock Series in computational methods in mechanics and thermal sciences. Taylor \& Francis, 1980.
\newblock ISBN 9780891165224.

\bibitem[Peng et~al.(2025)Peng, Addepalli, and Farsi]{peng2025machine}
Shaoyang Peng, Sri Addepalli, and Maryam Farsi.
\newblock Machine learning in thermography non-destructive testing: a systematic review.
\newblock \emph{Applied Sciences}, 15\penalty0 (17):\penalty0 9624, 2025.

\bibitem[P{\'e}rez et~al.(2025)P{\'e}rez, Ard{\i}{\c{c}}, {\c{C}}ak{\i}ro{\u{g}}lu, Jacob, Kodera, Pompa, Rachid, Wang, Zhou, Zimmer, et~al.]{perez2025integrating}
Eduardo P{\'e}rez, Cemil~Emre Ard{\i}{\c{c}}, Ozan {\c{C}}ak{\i}ro{\u{g}}lu, Kevin Jacob, Sayako Kodera, Luca Pompa, Mohamad Rachid, Han Wang, Yiming Zhou, Cyril Zimmer, et~al.
\newblock Integrating ai in nde: techniques, trends, and further directions.
\newblock \emph{NDT \& E International}, 156:\penalty0 103442, 2025.

\bibitem[Qian et~al.(2023)Qian, Hui, Wang, Zhang, Lin, and Yang]{qian2023physics}
Weijia Qian, Xin Hui, Bosen Wang, Zongwei Zhang, Yuzhen Lin, and Siheng Yang.
\newblock Physics-informed neural network for inverse heat conduction problem.
\newblock \emph{Heat Transfer Research}, 54\penalty0 (4), 2023.

\bibitem[Raissi et~al.(2019)Raissi, Perdikaris, and Karniadakis]{raissi2019physics}
Maziar Raissi, Paris Perdikaris, and George~E Karniadakis.
\newblock Physics-informed neural networks: A deep learning framework for solving forward and inverse problems involving nonlinear partial differential equations.
\newblock \emph{Journal of Computational physics}, 378:\penalty0 686--707, 2019.

\bibitem[Rashidi et~al.(2020)Rashidi, Olfatbakhsh, Crawford, and Milani]{rashidi2020review}
A~Rashidi, T~Olfatbakhsh, B~Crawford, and A~S Milani.
\newblock A review of current challenges and case study toward optimizing micro-computed {X}-ray tomography of carbon fabric composites.
\newblock \emph{Materials}, 13\penalty0 (16):\penalty0 3606, 2020.

\bibitem[Ronneberger et~al.(2015)Ronneberger, Fischer, and Brox]{ronneberger2015u}
Olaf Ronneberger, Philipp Fischer, and Thomas Brox.
\newblock U-net: Convolutional networks for biomedical image segmentation.
\newblock In \emph{International Conference on Medical image computing and computer-assisted intervention}, pages 234--241. Springer, 2015.

\bibitem[Rosa et~al.(2025)Rosa, Barella, Vargas, Tarpani, Herrmann, and Fernandes]{rosa2025advanced}
Renan~Garcia Rosa, Bruno~Pereira Barella, Iago~Garcia Vargas, Jos{\'e}~Ricardo Tarpani, Hans-Georg Herrmann, and Henrique Fernandes.
\newblock Advanced thermal imaging processing and deep learning integration for enhanced defect detection in carbon fiber-reinforced polymer laminates.
\newblock \emph{Materials}, 18\penalty0 (7):\penalty0 1448, 2025.

\bibitem[Rudin et~al.(1992)Rudin, Osher, and Fatemi]{rudin1992nonlinear}
Leonid~I Rudin, Stanley Osher, and Emad Fatemi.
\newblock Nonlinear total variation based noise removal algorithms.
\newblock \emph{Physica D: nonlinear phenomena}, 60\penalty0 (1-4):\penalty0 259--268, 1992.

\bibitem[Santosa(1996)]{santosa1996level}
F~Santosa.
\newblock A level-set approach for inverse problems involving obstacles.
\newblock \emph{ESAIM: Control, Optimisation and Calculus of Variations}, 1:\penalty0 17--33, 1996.

\bibitem[Schager et~al.(2020)Schager, Zauner, Mayr, and Burgholzer]{schager2020extension}
Alexander Schager, Gerald Zauner, G{\"u}nther Mayr, and Peter Burgholzer.
\newblock Extension of the thermographic signal reconstruction technique for an automated segmentation and depth estimation of subsurface defects.
\newblock \emph{Journal of Imaging}, 6\penalty0 (9):\penalty0 96, 2020.

\bibitem[Shepard and Beemer(2015)]{shepard2015advances}
Steven~M Shepard and Maria~Frendberg Beemer.
\newblock Advances in thermographic signal reconstruction.
\newblock In \emph{Thermosense: thermal infrared applications XXXVII}, volume 9485, pages 204--210. SPIE, 2015.

\bibitem[Shepard et~al.(2002)Shepard, Wang, Lhota, Rubadeux, and Ahmed]{shepard2002reconstruction}
Steven~M Shepard, D~Wang, James~R Lhota, Bruce~A Rubadeux, and Tasdiq Ahmed.
\newblock Reconstruction and enhancement of thermographic sequence data.
\newblock In \emph{Nondestructive evaluation and health monitoring of aerospace materials and civil infrastructures}, volume 4704, pages 74--77. SPIE, 2002.

\bibitem[Shi and Hsieh(2021)]{shi2021infrared}
Xijin Shi and Sheng-Jen Hsieh.
\newblock Infrared imaging and machine learning techniques for plant root location and depth prediction.
\newblock In \emph{Thermosense: Thermal Infrared Applications XLIII}, volume 11743, page 1174303. SPIE, 2021.

\bibitem[Sirikham et~al.(2020)Sirikham, Zhao, Liu, Xu, Williams, and Mehnen]{sirikham2020three}
A~Sirikham, Y~Zhao, H~Liu, Y~Xu, S~Williams, and J~Mehnen.
\newblock Three-dimensional subsurface defect shape reconstruction and visualisation by pulsed thermography.
\newblock \emph{Infrared Physics \& Technology}, 104:\penalty0 103151, 2020.

\bibitem[Sitzmann et~al.(2020)Sitzmann, Martel, Bergman, Lindell, and Wetzstein]{sitzmann2020implicit}
Vincent Sitzmann, Julien Martel, Alexander Bergman, David Lindell, and Gordon Wetzstein.
\newblock Implicit neural representations with periodic activation functions.
\newblock \emph{Advances in neural information processing systems}, 33:\penalty0 7462--7473, 2020.

\bibitem[Turpin et~al.(2023)Turpin, Zhong, Zhang, Zhu, Heiden, Macklin, Tsogkas, Dickinson, and Garg]{turpin2023fast}
Dylan Turpin, Tao Zhong, Shutong Zhang, Guanglei Zhu, Eric Heiden, Miles Macklin, Stavros Tsogkas, Sven Dickinson, and Animesh Garg.
\newblock Fast-grasp'd: Dexterous multi-finger grasp generation through differentiable simulation.
\newblock In \emph{2023 IEEE International Conference on Robotics and Automation (ICRA)}, 2023.

\bibitem[Vavilov et~al.(1992)Vavilov, Maldague, Picard, Thomas, and Favro]{vavilov1992dynamic}
V~Vavilov, Xavier Maldague, J~Picard, RL~Thomas, and LD~Favro.
\newblock Dynamic thermal tomography: new nde technique to reconstruct inner solids structure using multiple ir image processing.
\newblock In \emph{Review of progress in quantitative nondestructive evaluation}, pages 425--432. Springer, 1992.

\bibitem[Wang et~al.(2022)Wang, Yu, and Perdikaris]{wang2022and}
Sifan Wang, Xinling Yu, and Paris Perdikaris.
\newblock When and why pinns fail to train: A neural tangent kernel perspective.
\newblock \emph{Journal of Computational Physics}, 449:\penalty0 110768, 2022.

\bibitem[Wang et~al.(2024)Wang, Sankaran, and Perdikaris]{wang2024causal}
Sifan Wang, Shyam Sankaran, and Paris Perdikaris.
\newblock Respecting causality for training physics-informed neural networks.
\newblock \emph{Computer Methods in Applied Mechanics and Engineering}, 421:\penalty0 116813, 2024.

\bibitem[Wang et~al.(2004)Wang, Bovik, Sheikh, and Simoncelli]{wang2004image}
Zhou Wang, Alan~C Bovik, Hamid~R Sheikh, and Eero~P Simoncelli.
\newblock Image quality assessment: from error visibility to structural similarity.
\newblock \emph{IEEE transactions on image processing}, 13\penalty0 (4):\penalty0 600--612, 2004.

\bibitem[Wei et~al.(2023{\natexlab{a}})Wei, Osman, Valeske, and Maldague]{wei2023depth}
Ziang Wei, Ahmad Osman, Bernd Valeske, and Xavier Maldague.
\newblock A dataset of pulsed thermography for automated defect depth estimation.
\newblock \emph{Applied Sciences}, 13\penalty0 (24):\penalty0 13093, 2023{\natexlab{a}}.

\bibitem[Wei et~al.(2023{\natexlab{b}})Wei, Osman, Valeske, and Maldague]{wei2023pulsed}
Ziang Wei, Ahmad Osman, Bernd Valeske, and Xavier Maldague.
\newblock Pulsed thermography dataset for training deep learning models.
\newblock \emph{Applied Sciences}, 13\penalty0 (5):\penalty0 2901, 2023{\natexlab{b}}.

\bibitem[Wu et~al.(2023)Wu, Chen, Feng, Qi, Feng, Tao, and Zhang]{wu2023automatic}
Z~Wu, S~Chen, F~Feng, J~Qi, L~Feng, N~Tao, and C~Zhang.
\newblock Automatic defect detection and three-dimensional reconstruction from pulsed thermography images based on a bidirectional long-short term memory network.
\newblock \emph{Engineering Applications of Artificial Intelligence}, 124:\penalty0 106574, 2023.

\bibitem[Xu et~al.(2025)Xu, Yang, Liu, Lyu, Descovich, Ruan, and Sheng]{xu2025tomograf}
Di~Xu, Yang Yang, Hengjie Liu, Qihui Lyu, Martina Descovich, Dan Ruan, and Ke~Sheng.
\newblock Tomograf: An x-ray physics-driven generative radiance field framework for extremely sparse view ct reconstruction.
\newblock \emph{Plos one}, 20\penalty0 (8):\penalty0 e0330463, 2025.

\bibitem[Zhong and Allen-Blanchette(2025)]{zhong2025gagrasp}
Tao Zhong and Christine Allen-Blanchette.
\newblock Gagrasp: Geometric algebra diffusion for dexterous grasping.
\newblock In \emph{2025 IEEE International Conference on Robotics and Automation (ICRA)}, pages 6771--6778, 2025.
\newblock \doi{10.1109/ICRA55743.2025.11127957}.

\bibitem[Zhou et~al.(2025)Zhou, Fang, Morovati, Liu, Han, Xu, and Yu]{zhou2025rho}
Li~Zhou, Changsheng Fang, Bahareh Morovati, Yongtong Liu, Shuo Han, Yongshun Xu, and Hengyong Yu.
\newblock $\rho$-nerf: leveraging attenuation priors in neural radiance field for 3d computed tomography reconstruction.
\newblock In \emph{2025 IEEE International Conference on Image Processing (ICIP)}, pages 1636--1641. IEEE, 2025.

\end{thebibliography}
}


\appendix
\section*{Appendix}

\section{Effective Diffusivity Model and Interface Flux}
\label{appdx:effective}

\subsection{Derivation of the diffusivity form}
\label{appdx:effective_derivation}

Let $\rho, C_p \in C^1(\Omega)$ with $\rho C_p > 0$ on $\bar{\Omega}$, and let $T \in C^{2,1}(\Omega \times (0, t_{\mathrm{end}}])$ satisfy the conservation form~\eqref{eq:full_pde} with $Q \equiv 0$:
\begin{equation*}
    \rho C_p \, \partial_t T \;=\; \nabla \cdot (k \nabla T).
\end{equation*}
Dividing both sides by the strictly positive function $\rho C_p$ and applying the product rule $\nabla \cdot (\gamma \mathbf{v}) = \gamma \nabla \cdot \mathbf{v} + \nabla \gamma \cdot \mathbf{v}$ with $\gamma = 1/(\rho C_p)$ and $\mathbf{v} = k \nabla T$ gives
\begin{align*}
    \partial_t T \;&=\; \frac{1}{\rho C_p} \, \nabla \cdot (k \nabla T) \;=\; \nabla \cdot \!\left( \frac{k}{\rho C_p} \nabla T \right) \;-\; \nabla \!\left(\frac{1}{\rho C_p}\right) \cdot (k \nabla T) \\
    \;&=\; \nabla \cdot (\alpha \nabla T) \;-\; k \nabla T \cdot \nabla \!\left( \frac{1}{\rho C_p} \right),
\end{align*}
with $\alpha := k/(\rho C_p)$. The reduction is reversible by multiplying through by $\rho C_p$, so the conservation form~\eqref{eq:full_pde} (under $Q \equiv 0$) and the diffusivity form~\eqref{eq:diffusivity_form} are equivalent on $\Omega \times (0, t_{\mathrm{end}}]$ whenever $\rho C_p \in C^1$.

\subsection{Distributional treatment of the heat-capacity correction}
\label{appdx:effective_correction}

When $\rho C_p$ is piecewise-constant with values $c_b > 0$ on a bulk subdomain $\Omega_b$ and $c_d > 0$ on a defect subdomain $\Omega_d$, separated by a smooth interface $\Sigma = \partial \Omega_b \cap \partial \Omega_d$, the gradient $\nabla(1/(\rho C_p))$ is a vector-valued distribution supported on $\Sigma$:
\begin{equation}
\label{eq:interface_distribution}
    \nabla \!\left( \frac{1}{\rho C_p} \right) \;=\; \left( \frac{1}{c_d} - \frac{1}{c_b} \right) \mathbf{n}_\Sigma \, \delta_\Sigma,
\end{equation}
where $\mathbf{n}_\Sigma$ is the unit normal to $\Sigma$ pointing from $\Omega_b$ to $\Omega_d$ and $\delta_\Sigma$ is the surface measure on $\Sigma$. The correction term in~\eqref{eq:diffusivity_form} reads
\begin{equation*}
    -\, k \nabla T \cdot \nabla \!\left( \frac{1}{\rho C_p} \right) \;=\; -\!\left( \frac{1}{c_d} - \frac{1}{c_b} \right) (k \nabla T \cdot \mathbf{n}_\Sigma) \, \delta_\Sigma,
\end{equation*}
which is identically zero on $\Omega \setminus \Sigma$ and concentrates as a jump in the normal physical Fourier flux $q := -k \, \partial_n T$ across $\Sigma$. A finite-volume discretization on a uniform Cartesian grid of spacing $\Delta x$ cannot resolve a Dirac measure on a measure-zero interface~\cite{patankar1980numerical}. Dropping this distributional term yields the bulk equation $\partial_t T = \nabla \cdot (\alpha \nabla T)$ on $\Omega \setminus \Sigma$, which is the model NeFTY actually solves.

The single-$\alpha$ effective model has its own interface conservation law: across a face where $\alpha$ jumps, the effective $\alpha$-flux $j_\alpha := -\alpha \partial_n T$ must be continuous, since otherwise temperature would lose mass at that face. The remainder of this appendix shows that the harmonic-mean stencil enforces this $j_\alpha$-continuity exactly: it is the unique discrete analogue of the jump condition appropriate to the single-$\alpha$ model that NeFTY parameterizes, while the dropped distributional term belongs to the full $\rho C_p$ model.

\subsection{Proof of Proposition~\ref{prop:harmonic} (Harmonic mean as discrete interface continuity)}
\label{appdx:effective_proof2}

Consider two adjacent one-dimensional cells with center coordinates $x_i$ and $x_{i+1} = x_i + \Delta x$, cell-centered diffusivities $\alpha_i$ and $\alpha_{i+1}$, and cell-centered temperatures $T_i$ and $T_{i+1}$. Assume each cell has constant diffusivity, with the discontinuity located at the cell face $x_{i+1/2} = x_i + \Delta x / 2$. Let $T_{i+1/2}$ denote the (a priori unknown) face temperature.

Within the left half-cell $[x_i, x_{i+1/2}]$, Fourier's law for the effective $\alpha$-flux $j_\alpha = -\alpha \partial_x T$ gives a constant flux and a linear temperature profile, so
\begin{equation}
\label{eq:half_left}
    j_\alpha \;=\; -\alpha_i \cdot \frac{T_{i+1/2} - T_i}{\Delta x / 2}.
\end{equation}
Within the right half-cell $[x_{i+1/2}, x_{i+1}]$, by the same argument,
\begin{equation}
\label{eq:half_right}
    j_\alpha \;=\; -\alpha_{i+1} \cdot \frac{T_{i+1} - T_{i+1/2}}{\Delta x / 2}.
\end{equation}
Continuity of the effective flux at $x_{i+1/2}$ in the single-$\alpha$ model imposes the same value of $j_\alpha$ in~\eqref{eq:half_left} and~\eqref{eq:half_right}. From~\eqref{eq:half_left}, $T_{i+1/2} = T_i - j_\alpha \Delta x / (2 \alpha_i)$. From~\eqref{eq:half_right}, $T_{i+1/2} = T_{i+1} + j_\alpha \Delta x / (2 \alpha_{i+1})$. Equating and rearranging,
\begin{equation*}
    T_i - T_{i+1} \;=\; j_\alpha \, \frac{\Delta x}{2} \!\left( \frac{1}{\alpha_i} + \frac{1}{\alpha_{i+1}} \right) \;=\; j_\alpha \, \frac{\Delta x \, (\alpha_i + \alpha_{i+1})}{2 \alpha_i \alpha_{i+1}}.
\end{equation*}
Solving for $j_\alpha$,
\begin{equation*}
    j_\alpha \;=\; -\frac{2 \alpha_i \alpha_{i+1}}{\alpha_i + \alpha_{i+1}} \cdot \frac{T_{i+1} - T_i}{\Delta x} \;=:\; -\bar{\alpha}_{i+1/2} \cdot \frac{T_{i+1} - T_i}{\Delta x}.
\end{equation*}
The expression $\bar{\alpha}_{i+1/2} = 2 \alpha_i \alpha_{i+1} / (\alpha_i + \alpha_{i+1})$ is the harmonic mean. Uniqueness follows from the linearity of the two-equation system in $(j_\alpha, T_{i+1/2})$ with non-degenerate coefficient matrix. \hfill $\square$

\subsection{Multi-dimensional consequence and physical interpretation}
\label{appdx:effective_multid}

In ambient spatial dimension $D \in \{2, 3\}$ on a uniform Cartesian grid, the same argument applied along each principal direction gives the second-order finite-volume diffusion operator $\mathbf{L}(\alpha)$ used by NeFTY, defined by its action on a discrete temperature field $\mathbf{T}$ at grid node $(i, j, k)$:
\begin{equation*}
    {[\mathbf{L}(\alpha) \mathbf{T}]}_{i,j,k} \;:=\; \frac{1}{\Delta x^2} \!\left[ \bar{\alpha}_{i+1/2}(T_{i+1,j,k} - T_{i,j,k}) - \bar{\alpha}_{i-1/2}(T_{i,j,k} - T_{i-1,j,k}) \right] \;+\; (\text{$y$-, $z$-terms}).
\end{equation*}
The harmonic mean has two physically important properties for thermal NDT. First, $\bar{\alpha}_{i+1/2}$ is dominated by the smaller of $\alpha_i, \alpha_{i+1}$ when the two are well separated, so an insulating defect cell $\alpha_{i+1} \ll \alpha_i$ throttles flux at its boundary as physical thermal resistance does. Second, $\bar{\alpha}$ is exact for piecewise-constant $\alpha$ at cell faces and second-order accurate for smooth $\alpha$. Using the arithmetic mean instead is first-order accurate and overestimates flux across high-contrast interfaces, leading to systematic underestimation of defect depth in inverse reconstruction~\cite{patankar1980numerical}. NeFTY's adoption of the harmonic mean is therefore the unique discrete analogue of the effective-flux continuity condition for the single-$\alpha$ bulk equation in~\eqref{eq:diffusivity_form}.

\subsection{Initial and boundary conditions: synthetic and real PVC datasets}
\label{appdx:bc}

The bulk diffusivity equation $\partial_t T = \nabla \cdot (\alpha \nabla T)$ is closed by an initial condition and by lateral and through-thickness boundary conditions. The synthetic dataset and the real PVC datasets~\cite{wei2023pulsed,wei2023depth} require different choices, all consistent with the same underlying physics but adapted to the specimen scale, the heating modality, and the dominant non-dimensional groups.

\textbf{Synthetic dataset.} The synthetic data are generated on a unitless slab $\Omega = [0, L]^2 \times [0, H]$ with $L = 10$, $H = 1$, grid resolution $64 \times 64 \times 16$ and 100 transient frames, on micron-scale specimens at microsecond timescales for which the Biot number is small and convective dissipation is negligible. We accordingly impose:
\begin{itemize}[leftmargin=*,topsep=2pt,itemsep=1pt]
    \item \textit{Initial condition.} A spatially Gaussian post-flash temperature distribution localized near the front face,
    \begin{equation*}
        T_0(\mathbf{x}) \;=\; T_{\mathrm{amb}} + A \exp\!\big( -((x - x_c)^2 + (y - y_c)^2)/(2 w_{xy}^2) \big) \exp\!\big( - z^2 / (2 w_z^2) \big),
    \end{equation*}
    where $T_{\mathrm{amb}}$ is the ambient temperature, $A$ is the deposited pulse amplitude, $(x_c, y_c)$ is the lateral footprint center, $w_{xy}$ is the lateral pulse width, and $w_z \ll H$ is the absorption depth set by the focused-laser pulse profile.
    \item \textit{Lateral boundaries ($\partial \Omega_{xy}$).} Periodic conditions on $\partial \Omega_{xy} := [0,L]^2 \times \{0,L\}_x \times \{0,L\}_y$ (the $x = 0/L$ and $y = 0/L$ faces), modeling a semi-infinite slab and removing reflections from the truncated simulation grid.
    \item \textit{Through-thickness boundaries ($\partial \Omega_{z}$).} Adiabatic conditions $\mathbf{n} \cdot (\alpha \nabla T) = 0$ on $\partial \Omega_z$ (the front $z = 0$ and back $z = H$ faces), modeling negligible convective loss during the short inspection window.
\end{itemize}

\textbf{Real PVC dataset.} The PVC-Infrared~\cite{wei2023pulsed} and PVC-Depth~\cite{wei2023depth} benchmarks consist of pulsed thermography sequences on $100 \times 100 \times 5$ mm PVC specimens with cylindrical flat-bottom holes of varying size and depth, recorded over second-scale dynamics. Three physical adaptations are required so that the same NeFTY solver applies without retraining the ground-truth simulator:
\begin{itemize}[leftmargin=*,topsep=2pt,itemsep=1pt]
    \item \textit{Initial condition.} The Gaussian pulse is replaced by a near-uniform front-face flash $T_0(\mathbf{x}) = T_{\mathrm{amb}} + A \exp(-z^2 / (2 w_z^2))$, matching the broad flash-lamp excitation used in the benchmark setups (lateral profile uniform over the inspection footprint, absorption depth $w_z$ small relative to specimen thickness).
    \item \textit{Centimeter and second-scale Fourier-number rescaling.} The PVC specimens are centimeter-scale at second-scale dynamics rather than the synthetic micron / microsecond regime, so the unitless solver is rescaled through the Fourier number $\mathrm{Fo} = \alpha_{\mathrm{phys}}\, t_{\mathrm{total}} / L_0^2$ (full scaling below).
    \item \textit{Convective Robin back-face boundary.} The centimeter-scale specimen and second-scale dynamics make convective dissipation non-negligible. On the back face we therefore impose a Robin condition $\mathbf{n} \cdot (\alpha \nabla T) = -h \, (T - T_{\mathrm{amb}})$ with $h > 0$ a calibrated convective coefficient, recovering the asymptotic Biot regime appropriate for the PVC scale. The front face retains the adiabatic Neumann condition, and lateral conditions remain periodic on interior crops or are replaced by adiabatic Neumann depending on the recording.
\end{itemize}

The unitless and dimensional scales are aligned through the Fourier number $\mathrm{Fo} = \alpha_{\mathrm{phys}} \, t_{\mathrm{total}} / L_0^2$, with $\alpha_{\mathrm{phys}}$ the physical diffusivity (in $\mathrm{m}^2/\mathrm{s}$), $t_{\mathrm{total}}$ the physical experiment duration (in $\mathrm{s}$), and $L_0$ a characteristic specimen length scale (in $\mathrm{m}$). The unitless solver evolves in $(t/t_{\mathrm{total}}, \mathbf{x}/L_0)$ coordinates with effective diffusivity $\alpha = \alpha_{\mathrm{phys}} \, t_{\mathrm{total}} / L_0^2$. A single set of unitless solver parameters represents widely different material classes by reinterpreting $L_0$ and $t_{\mathrm{total}}$, so the synthetic and real PVC settings differ in their initial-condition shape and back-face Robin coefficient but share the same forward operator $\mathcal{K}$ and the same harmonic-mean stencil derived above. Full physical scaling, calibrated $h$ values, and projection from a reconstructed 3D $\alpha$ field to the 2D defect masks and 2.5D depth labels supplied by the benchmarks are documented in Appendix~\ref{appdx:pvc}.

\section{Ill-Posedness of the Inverse Heat Conduction Problem}
\label{appdx:illposed}

This appendix establishes the compactness and algebraic singular-value decay claimed in Proposition~\ref{prop:svd}. The argument is consistent with the standard treatment of parameter identification in parabolic inverse problems~\cite{hadamard1888rayon,martinez2024artificial,engl1996regularization}. We present a self-contained derivation on the slab geometry used in our experiments. The proof factorizes $d\mathcal{K}_{\alpha_0} = B \circ J$, where $B: L^2(\Omega) \to L^2(\Gamma_{\mathrm{obs}} \times (0, t_{\mathrm{end}}))$ is bounded by an energy estimate plus a parabolic boundary trace, and $J: H^1(\Omega) \hookrightarrow L^2(\Omega)$ is the compact Rellich--Kondrachov embedding whose approximation numbers along the Laplacian eigenbasis decay as $(1 + \lambda_n)^{-1/2}$. Truncating the parameter side via $J$ rather than the source side avoids the mode-mixing introduced by multiplication with $\nabla T_{\alpha_0}$.

\subsection{Linearization of the parameter-to-observation map}
\label{appdx:illposed_lin}

Let $\alpha_0 > 0$ be a constant background diffusivity and let $T_{\alpha_0}(\mathbf{x}, t)$ solve the bulk heat equation $\partial_t T = \alpha_0 \Delta T$ with the prescribed initial condition $T_0$ and boundary conditions on the slab $\Omega = [0, L]^2 \times [0, H]$. Throughout the proof we assume the background satisfies $\nabla T_{\alpha_0} \in L^\infty(Q)$ with $Q := \Omega \times (0, t_{\mathrm{end}})$. Consider a perturbation $\alpha = \alpha_0 + \varepsilon \, \delta \alpha$ with $\delta \alpha \in H^1(\Omega)$ and $\varepsilon > 0$ small. To first order in $\varepsilon$, the perturbation $\delta T = (T_\alpha - T_{\alpha_0}) / \varepsilon$ solves the linearized PDE in the weak sense
\begin{equation}
\label{eq:linearized}
    \langle \partial_t (\delta T), v \rangle + \alpha_0 \big( \nabla(\delta T), \nabla v \big)_{L^2(\Omega)} \;=\; -\!\int_\Omega \delta \alpha \, \nabla T_{\alpha_0}(\cdot, t) \cdot \nabla v \, d\mathbf{x},
\end{equation}
for all $v \in H^1(\Omega)$ compatible with the imposed boundary conditions, with vanishing initial condition $\delta T(\cdot, 0) = 0$. We define the source distribution $F_{\delta \alpha}(t) \in H^{-1}(\Omega)$ by $\langle F_{\delta \alpha}(t), v \rangle = -\!\int_\Omega \delta\alpha \, \nabla T_{\alpha_0}(\cdot, t) \cdot \nabla v \, d\mathbf{x}$, satisfying
\begin{equation}
\label{eq:source_Hinv}
    \| F_{\delta \alpha}(t) \|_{H^{-1}(\Omega)} \;\le\; \| \nabla T_{\alpha_0}(\cdot, t) \|_{L^\infty(\Omega)} \, \| \delta \alpha \|_{L^2(\Omega)}.
\end{equation}
The Fr\'echet derivative is the linear map $d \mathcal{K}_{\alpha_0}: H^1(\Omega) \to L^2(\Gamma_{\mathrm{obs}} \times (0, t_{\mathrm{end}}))$ defined by $d\mathcal{K}_{\alpha_0}(\delta \alpha) = \delta T |_{\Gamma_{\mathrm{obs}} \times (0, t_{\mathrm{end}})}$.

\subsection{Proof of Proposition~\ref{prop:svd} (Compactness and algebraic singular-value decay)}
\label{appdx:illposed_proof}

The proof has four steps: (i) an energy estimate in $L^2(0, t_{\mathrm{end}}; H^1(\Omega))$ for $\delta T$; (ii) a parabolic trace inequality giving boundedness of the auxiliary map $B$; (iii) a finite-rank approximation through the Rellich--Kondrachov embedding $J$; (iv) Weyl asymptotics on the slab.

\textbf{Step 1: energy estimate.} Existence and uniqueness of a weak solution $\delta T \in L^2(0, t_{\mathrm{end}}; H^1(\Omega)) \cap H^1(0, t_{\mathrm{end}}; H^{-1}(\Omega))$ to~\eqref{eq:linearized} with vanishing initial data and source $F_{\delta \alpha} \in L^2(0, t_{\mathrm{end}}; H^{-1}(\Omega))$ follow from standard Galerkin / Lions--Magenes parabolic theory~\cite{engl1996regularization}. With this regularity, testing~\eqref{eq:linearized} with $v = \delta T(\cdot, t)$ is permissible and gives
\begin{equation*}
    \tfrac{1}{2} \tfrac{d}{dt} \| \delta T(\cdot, t) \|_{L^2(\Omega)}^2 + \alpha_0 \| \nabla \delta T(\cdot, t) \|_{L^2(\Omega)}^2 \;=\; -\!\int_\Omega \delta \alpha \, \nabla T_{\alpha_0}(\cdot, t) \cdot \nabla (\delta T)(\cdot, t) \, d\mathbf{x}.
\end{equation*}
Applying Young's inequality $|ab| \le \tfrac{1}{2 \alpha_0} a^2 + \tfrac{\alpha_0}{2} b^2$ to the right-hand side and integrating over $(0, t_{\mathrm{end}})$ with the vanishing initial condition,
\begin{equation*}
    \tfrac{1}{2} \| \delta T(\cdot, t_{\mathrm{end}}) \|_{L^2(\Omega)}^2 + \tfrac{\alpha_0}{2} \int_0^{t_{\mathrm{end}}} \| \nabla \delta T \|_{L^2(\Omega)}^2 \, dt \;\le\; \tfrac{1}{2 \alpha_0} \int_0^{t_{\mathrm{end}}} \| \delta \alpha \, \nabla T_{\alpha_0}(\cdot, t) \|_{L^2(\Omega)}^2 \, dt.
\end{equation*}
With $\| \delta \alpha \, \nabla T_{\alpha_0}(\cdot, t) \|_{L^2(\Omega)} \le \| \nabla T_{\alpha_0}(\cdot, t) \|_{L^\infty(\Omega)} \, \| \delta \alpha \|_{L^2(\Omega)}$, this yields
\begin{equation}
\label{eq:energy}
    \| \delta T \|_{L^2(0, t_{\mathrm{end}}; H^1(\Omega))} \;\le\; C_E(\Omega, t_{\mathrm{end}}, \alpha_0) \, \| \nabla T_{\alpha_0} \|_{L^\infty(Q)} \, \| \delta \alpha \|_{L^2(\Omega)},
\end{equation}
with $C_E$ an absolute constant depending only on $\Omega$, $t_{\mathrm{end}}$, $\alpha_0$ (a Poincar\'e-type constant absorbed into the bound).

\textbf{Step 2: boundary trace and the auxiliary operator $B$.} The Sobolev trace inequality $\| u |_{\partial \Omega} \|_{L^2(\partial \Omega)} \le C_\Gamma \| u \|_{H^1(\Omega)}$ on a Lipschitz domain combined with~\eqref{eq:energy} gives
\begin{equation*}
    \| \delta T |_{\Gamma_{\mathrm{obs}} \times (0, t_{\mathrm{end}})} \|_{L^2(\Gamma_{\mathrm{obs}} \times (0, t_{\mathrm{end}}))}^2 \;\le\; C_\Gamma^2 \, \| \delta T \|_{L^2(0, t_{\mathrm{end}}; H^1(\Omega))}^2 \;\le\; C_B^2 \, \| \delta \alpha \|_{L^2(\Omega)}^2,
\end{equation*}
with $C_B := C_\Gamma \, C_E \, \| \nabla T_{\alpha_0} \|_{L^\infty(Q)}$. The map
\begin{equation*}
    B \,:\, L^2(\Omega) \,\to\, L^2(\Gamma_{\mathrm{obs}} \times (0, t_{\mathrm{end}})), \qquad B(\delta \alpha) \,=\, \delta T |_{\Gamma_{\mathrm{obs}} \times (0, t_{\mathrm{end}})},
\end{equation*}
is therefore bounded, $\| B \|_{\mathrm{op}} \le C_B$. The Fr\'echet derivative is $d\mathcal{K}_{\alpha_0} = B \circ J$, where $J: H^1(\Omega) \hookrightarrow L^2(\Omega)$ is the inclusion.

\textbf{Step 3: compactness and singular-value decay via Rellich--Kondrachov.} The embedding $J$ is compact by the Rellich--Kondrachov theorem on bounded Lipschitz domains. Equipping $H^1(\Omega)$ with the spectral norm $\| \delta \alpha \|_{H^1}^2 = \sum_{j \ge 0} (1 + \lambda_j) |a_j|^2$ where $\delta \alpha = \sum_j a_j \phi_j$, the approximation numbers of $J$ along the Laplacian eigenbasis are exactly $(1 + \lambda_n)^{-1/2}$: for the $L^2$-orthogonal projector $P_N$ onto $\mathrm{span}(\phi_0, \dots, \phi_{N-1})$,
\begin{equation*}
    \| (I - P_N) \delta \alpha \|_{L^2(\Omega)}^2 \;=\; \sum_{j \ge N} a_j^2 \;\le\; (1 + \lambda_N)^{-1} \sum_{j \ge N} (1 + \lambda_j) a_j^2 \;\le\; (1 + \lambda_N)^{-1} \, \| \delta \alpha \|_{H^1(\Omega)}^2.
\end{equation*}
Define the rank-$N$ operator $A_N := B \circ P_N \circ J$. For $\| \delta \alpha \|_{H^1} \le 1$,
\begin{equation*}
    \| (d\mathcal{K}_{\alpha_0} - A_N)(\delta \alpha) \|_{L^2(\Gamma_{\mathrm{obs}} \times (0, t_{\mathrm{end}}))} \;\le\; \| B \|_{\mathrm{op}} \, \| (I - P_N)(J \delta \alpha) \|_{L^2(\Omega)} \;\le\; C_B \, (1 + \lambda_N)^{-1/2}.
\end{equation*}
Since approximation numbers and singular values coincide for compact operators between Hilbert spaces~\cite{engl1996regularization}, this yields
\begin{equation}
\label{eq:sv_bound}
    \sigma_{N+1}(d\mathcal{K}_{\alpha_0}) \;\le\; C_B \, (1 + \lambda_N)^{-1/2}, \qquad N \ge 0.
\end{equation}
Compactness of $d\mathcal{K}_{\alpha_0}$ follows from $\sigma_n \to 0$, equivalently the convergence $A_N \to d\mathcal{K}_{\alpha_0}$ in operator norm. The mode-mixing introduced by multiplication with $\nabla T_{\alpha_0}$ in the source $F_{\delta \alpha}$ does not affect this argument: the truncation $P_N$ acts on the parameter $\delta \alpha$ before it enters the linearized PDE, and the small-$L^2$-tail of $\delta \alpha$ for $\delta \alpha \in H^1$-unit-ball produces a small output in $L^2(\Gamma_{\mathrm{obs}} \times (0, t_{\mathrm{end}}))$ via the $L^2$-bounded map $B$.

\textbf{Step 4: slab Weyl asymptotics.} For $\Omega = [0, L]^2 \times [0, H]$ with periodic conditions on the lateral faces and Neumann conditions on the top/bottom, the eigenfunctions are tensor products $\phi_{m, \ell, p}(x, y, z) = e^{2\pi i (m x / L + \ell y / L)} \cos(p \pi z / H)$ with eigenvalues $\lambda_{m, \ell, p} = (2\pi m/L)^2 + (2\pi \ell/L)^2 + (p \pi / H)^2$, indexed by $(m, \ell, p) \in \mathbb{Z}^2 \times \mathbb{Z}_{\ge 0}$. By Weyl's law in three dimensions, enumerated in increasing order $\lambda_n \asymp n^{2/3}$, so~\eqref{eq:sv_bound} reads $\sigma_n \lesssim n^{-1/3}$ on the slab. \hfill $\square$

\begin{remark}
\label{rmk:meanzero}
The constant Laplacian eigenmode $\phi_0 \equiv |\Omega|^{-1/2}$ has $\lambda_0 = 0$, and adding a constant to $\alpha$ only rescales time without changing the inverse problem. Restricting to mean-zero perturbations $\delta \alpha \in \{ a \in H^1(\Omega) : \int_\Omega a \, d\mathbf{x} = 0 \}$ with norm $\| \nabla \delta \alpha \|_{L^2}$ replaces the bound by $\sigma_n \le C \, \lambda_n^{-1/2}$ for $n \ge 1$, recovering exactly the algebraic rate $\nu = 1/2$.
\end{remark}

\subsection{Proof of Corollary~\ref{cor:hadamard} (Hadamard ill-posedness)}
\label{appdx:illposed_cor}

By Proposition~\ref{prop:svd}, the singular values of $d \mathcal{K}_{\alpha_0}$ tend to zero, so the Moore--Penrose pseudo-inverse $d \mathcal{K}_{\alpha_0}^{\dagger}$ is unbounded on the closure of the range. Concretely, the singular value decomposition $d\mathcal{K}_{\alpha_0}(\delta\alpha) = \sum_{n: \sigma_n > 0} \sigma_n \langle \delta\alpha, v_n \rangle u_n$ has right singular vectors $v_n \in H^1(\Omega)$ in the parameter space and left singular vectors $u_n \in L^2(\Gamma_{\mathrm{obs}} \times (0, t_{\mathrm{end}}))$ in the data space, with $d\mathcal{K}_{\alpha_0}^\dagger u_n = \sigma_n^{-1} v_n$ for the indices $n$ with $\sigma_n > 0$ and $d\mathcal{K}_{\alpha_0}^\dagger u_n = 0$ on the kernel of $d\mathcal{K}_{\alpha_0}^*$. A measurement perturbation aligned with a non-degenerate $n$-th left singular vector $u_n$ therefore produces an $H^1$-norm perturbation in the recovered diffusivity of magnitude $\sigma_n^{-1} \ge C_B^{-1} (1 + \lambda_n)^{1/2}$. The forward problem is well-posed (existence, uniqueness, continuous dependence in the forward direction follow from standard parabolic theory), but inversion fails the continuous-dependence criterion in the sense of Hadamard. \hfill $\square$

\subsection{Discussion: pointwise damping versus operator decay}
\label{appdx:illposed_discussion}

Two distinct smoothing rates appear in this analysis and should not be conflated. \emph{Pointwise} in time, the free-space heat kernel $G_t(\mathbf{x}) = (4 \pi \alpha_0 t)^{-3/2} \exp(-|\mathbf{x}|^2 / (4 \alpha_0 t))$ governs the impulse response: a localized defect at depth $z$ contributes a surface temperature signal whose amplitude decays Gaussian-fast in $z$, suppressed by $\exp(-z^2 / (4 \alpha_0 t))$ at observation time $t > 0$. Equivalently in the spectral picture, the heat semigroup $S(t) = e^{\alpha_0 \Delta t}$ damps the $j$-th Laplacian eigenmode by $e^{-\alpha_0 \lambda_j t}$, so a defect supported at frequency $\lambda_j$ is exponentially attenuated at any fixed time. These pointwise statements are the depth- and frequency-decay heuristics that motivate the original ill-posedness intuition for IHCP.

The \emph{operator} decay of $d \mathcal{K}_{\alpha_0}$, in contrast, integrates the boundary trace against the entire time interval $(0, t_{\mathrm{end}})$, and the bound~\eqref{eq:sv_bound} is only algebraic in $\lambda_n$. The two rates coincide in the degenerate observation model where the trace is taken at a single final time $t = t_{\mathrm{end}}$, in which case $\sigma_n(S(t_{\mathrm{end}})) = e^{-\alpha_0 \lambda_n t_{\mathrm{end}}}$ recovers the exponential rate. Integrating over $(0, t_{\mathrm{end}})$ averages out this exponential factor and yields the algebraic bound proven above.

The practical consequences for IHCP follow from the algebraic operator-rate. Longer observation windows $t_{\mathrm{end}}$ and higher-conductivity bulk materials improve early-time signal-to-noise ratio but preserve the algebraic singular-value rate. Regularization remains mandatory at any rate, since any consistent reconstruction method must implicitly or explicitly bound the high-frequency content of admissible $\alpha$. NeFTY combines two such regularizers, namely a continuous neural-field prior with frequency annealing (Section~\ref{sec:method}) and an isotropic total-variation penalty $\mathcal{R}(\alpha) = \int_\Omega \| \nabla \alpha \|$ aligned with the piecewise-constant defect structure assumed by the effective-diffusivity model of Section~\ref{sec:forward}.

\section{Gradient Decoupling in the Two-Network Soft-Constrained PINN Formulation}
\label{appdx:pinn}

PINNs remain the dominant and broadly successful paradigm for physics-informed learning~\cite{raissi2019physics}. This appendix analyzes one formulation of the inverse problem, the standard two-network formulation of Section~\ref{sec:pinn}: a temperature surrogate $T_\phi$ whose parameters $\phi$ are separate from the parameters $\theta$ of the diffusivity surrogate $\alpha_\theta$, trained on the data, residual, and initial-condition objective~\eqref{eq:pinn_loss}, applied to stiff parabolic inverse problems with surface-only data. Under these assumptions the decoupling follows from the parameterization alone (Appendix~\ref{appdx:pinn_proof}), so optimizer quality and tuning play no role in the derivation. This identity is distinct from the conditioning and optimization difficulties that PDE-residual penalties can introduce~\cite{krishnapriyan2021characterizing}. The derivation establishes decoupling. The collapse of $\alpha_\theta$ is the observed behavior of the four evaluated variants (Section~\ref{ssec:synth}). Appendix~\ref{appdx:pinn_coupled} examines a soft variant outside these assumptions.

\subsection{Derivation of the gradient decoupling}
\label{appdx:pinn_proof}

Write $\mathcal{L}_{\mathrm{PINN}} = \mathcal{L}_{\mathrm{data}}(T_\phi) + \lambda_{\mathrm{PDE}} \mathcal{L}_{\mathrm{PDE}}(T_\phi, \alpha_\theta) + \lambda_{\mathrm{IC}} \mathcal{L}_{\mathrm{IC}}(T_\phi)$ as in equation~\eqref{eq:pinn_loss}. The data term $\mathcal{L}_{\mathrm{data}}(T_\phi) = \| T_\phi - \hat{T} \|^2_{\Gamma_{\mathrm{obs}}}$ depends on $\theta$ only if $T_\phi$ does. Because $T_\phi$ is parameterized by $\phi$ alone, $\partial T_\phi / \partial \theta \equiv 0$, hence $\nabla_\theta \mathcal{L}_{\mathrm{data}} = 0$. The same argument applied to $\mathcal{L}_{\mathrm{IC}}(T_\phi)$ gives $\nabla_\theta \mathcal{L}_{\mathrm{IC}} = 0$. Therefore
\begin{equation*}
    \nabla_\theta \mathcal{L}_{\mathrm{PINN}} \;=\; \lambda_{\mathrm{PDE}} \, \nabla_\theta \mathcal{L}_{\mathrm{PDE}}(T_\phi, \alpha_\theta),
\end{equation*}
which establishes the gradient decoupling. To bound the residual-gradient magnitude, expand $\mathcal{L}_{\mathrm{PDE}}$ pointwise:
\begin{equation*}
    \mathcal{L}_{\mathrm{PDE}}(T_\phi, \alpha_\theta) \;=\; \frac{1}{N_c} \sum_{i=1}^{N_c} \big( r_i(T_\phi, \alpha_\theta) \big)^2, \qquad r_i \;=\; \partial_t T_\phi(\mathbf{x}_i, \tau_i) - \nabla \cdot (\alpha_\theta \nabla T_\phi)(\mathbf{x}_i, \tau_i),
\end{equation*}
where $\{(\mathbf{x}_i, \tau_i)\}_{i = 1}^{N_c}$ are the PINN collocation points (using $\tau_i$ for collocation times to distinguish them from the measurement times $t_i$ of Section~\ref{sec:inverse}). Differentiating with respect to $\theta$ and applying the product rule to the divergence,
\begin{equation*}
    \nabla \cdot (\alpha_\theta \nabla T_\phi) \;=\; \alpha_\theta \, \Delta T_\phi \;+\; \nabla \alpha_\theta \cdot \nabla T_\phi,
\end{equation*}
so
\begin{equation}
\label{eq:res_grad}
    \nabla_\theta r_i \;=\; -\Delta T_\phi(\mathbf{x}_i, \tau_i) \, \nabla_\theta \alpha_\theta(\mathbf{x}_i) \;-\; \nabla T_\phi(\mathbf{x}_i, \tau_i) \cdot \nabla_\theta \nabla \alpha_\theta(\mathbf{x}_i).
\end{equation}
The Cauchy--Schwarz bound $\| \nabla_\theta \mathcal{L}_{\mathrm{PDE}} \| \le \frac{2}{N_c} \sum_i | r_i | \, \| \nabla_\theta r_i \|$ combined with~\eqref{eq:res_grad} gives a residual-gradient magnitude controlled by the spatial derivatives $\Delta T_\phi$ and $\nabla T_\phi$ of the temperature surrogate at the collocation points, dominated by $\|\Delta T_\phi\|$ when $\nabla \alpha_\theta$ is bounded. Empirically and analytically, $\| \Delta T_\phi \|$ is large whenever $T_\phi$ has not yet converged, because second-order spatial derivatives of an under-fitted neural temperature field are noisy and large in magnitude.

The derivation establishes that surface observations reach $\theta$ only indirectly through the residual term, so the only avenue for $\hat{T}$ to influence $\alpha_\theta$ is via residual gradients that scale with the under-fitted Laplacian of $T_\phi$. The collapse of $\alpha_\theta$ to a near-trivial constant is the behavior observed for the four evaluated variants (Section~\ref{sec:experiments}). It is consistent with this channel and with PINN training analyses~\cite{wang2022and,hao2024training}, and it is the experimental complement to the present derivation.

\subsection{Coupling with the ill-posedness of Section~\ref{sec:inverse}}
\label{appdx:pinn_coupling}

Even if the gradient decoupling derived in Appendix~\ref{appdx:pinn_proof} were absent, recovering $\alpha_\theta$ through any chain that begins at the surface data would require inverting the parameter-to-observation map $\mathcal{K}$. By Corollary~\ref{cor:hadamard}, this inverse algebraically amplifies the $n$-th left singular vector by $\sigma_n^{-1} \ge C_B^{-1} (1 + \lambda_n)^{1/2}$. In the two-network formulation, the surface data influences $\alpha_\theta$ only second-hand, by first updating $T_\phi$ through $\mathcal{L}_{\mathrm{data}}$ and then propagating to $\alpha_\theta$ through the residual term: the high-frequency information needed to localize defects is already attenuated by the smoothing of the forward map before the residual gradient sees it, and must then be re-amplified by an unbounded pseudo-inverse. This mechanism is consistent with the experimentally observed pattern in which $\mathcal{L}_{\mathrm{data}}$ converges (the temperature net fits the surface) but $\alpha_\theta$ remains near a trivial constant, the soft-constraint analogue of the data-fit paradox documented in our experimental study (Section~\ref{sec:experiments}) and in independent analyses of PINN training~\cite{wang2022and,hao2024training}.

\subsection{Standard remedies within the two-network formulation}
\label{appdx:pinn_remedies}

Several SciML methods address documented PINN training pathologies: loss imbalance, collocation cost and residual variance, causality, and gradient conflict. The four evaluated here all keep the two-network parameterization, so we examine one question for each: whether it creates a path from $\hat{T}$ to $\theta$ that bypasses the residual term. In each case it does not, and the gradient decoupling continues to apply.

\textbf{GradNorm~\cite{chen2018gradnorm}.} GradNorm rescales the loss weights $\lambda_{\mathrm{PDE}}, \lambda_{\mathrm{IC}}$ to balance gradient magnitudes during training. Because the rescaling acts on $\lambda_{\mathrm{PDE}}$ and not on the parameterization, the identity $\nabla_\theta \mathcal{L}_{\mathrm{PINN}} = \lambda_{\mathrm{PDE}} \nabla_\theta \mathcal{L}_{\mathrm{PDE}}$ is unchanged. GradNorm changes the relative size of the residual gradient but cannot manufacture a data-side gradient on $\theta$.

\textbf{SPINN~\cite{cho2023spinn}.} SPINN replaces the joint $T_\phi(\mathbf{x}, t)$ network with a separable product $T_\phi(\mathbf{x}, t) = \sum_k T_\phi^{(1)}(x) \, T_\phi^{(2)}(y) \, T_\phi^{(3)}(z) \, T_\phi^{(4)}(t)$, accelerating collocation and reducing the variance of the residual term. Among the four evaluated variants, SPINN attains the highest aggregate SSIM (Table~\ref{tab:synth_main}), a partial success of this parameterization. The diffusivity network $\alpha_\theta$ remains independent of $\phi$, so the gradient decoupling is unchanged.

\textbf{Causal-PINN~\cite{wang2024causal}.} Causal-PINN reweights the residual integrand by a causal weight that emphasizes early times before late times. The weight enters $\mathcal{L}_{\mathrm{PDE}}$ through a non-negative pointwise factor and again does not change the parameterization. The decoupling result holds.

\textbf{DCGD~\cite{hwang2024dcgd}.} Dual-cone gradient descent projects the data and residual gradients onto a common acceptable cone to avoid destructive interference. It operates on already-computed gradients of $\mathcal{L}_{\mathrm{data}}$ and $\mathcal{L}_{\mathrm{PDE}}$ separately. Because $\nabla_\theta \mathcal{L}_{\mathrm{data}} = 0$ identically, the only gradient applied to $\theta$ remains the (projected) residual gradient.

In NeFTY the temperature is the implicit solution of a discretized PDE driven by $\alpha_\theta$, so the chain of dependencies is $\theta \to \alpha_\theta \to T \to \mathcal{L}_{\mathrm{data}}$ and the surface data term depends directly on $\theta$, with its gradient computed through the adjoint of the discrete heat operator (Section~\ref{sec:method}).

\subsection{A directly coupled soft variant}
\label{appdx:pinn_coupled}

The analysis above requires exactly (i) the two-network parameterization with $T_\phi$ independent of $\theta$ and (ii) the data, residual, and initial-condition objective. Under these, $\partial T_\phi / \partial \theta \equiv 0$ (Appendix~\ref{appdx:pinn_proof}). A soft-constrained architecture that lets surface data influence the diffusivity parameters directly falls outside this scope. We built such a variant to test whether a live data-to-$\theta$ path suffices on its own.

\textbf{Construction.} The temperature network is conditioned on the diffusivity network's scalar output through an input channel $c(\mathbf{x}, t) = w(t)\, \alpha_n(\mathbf{x})$, where $\alpha_n$ is the normalized bounded diffusivity and $w(t) = t / t_{\mathrm{end}}$ ramps from $w(0) = 0$, so the diffusivity-independent initial condition injects no gradient. Thus $\partial T / \partial \theta \neq 0$ for $t > 0$. The control is parameter-identical, with the channel zeroed.

\textbf{Verification of the coupling.} We verified the coupling directly: the surface-data-to-$\theta$ path is live, and a predefined virtual data-gradient perturbation induces a nonzero change in the predicted diffusivity field while that field is still interior to its parameterized range. Table~\ref{tab:coupled} summarizes the paired comparison.

\begin{table}[h]
\centering
\caption{\textbf{Directly coupled soft variant versus its parameter-identical control} (channel zeroed) on paired synthetic specimens.}
\label{tab:coupled}
\vspace{2pt}
\begin{sc}
\footnotesize
\begin{tabular}{p{0.27\linewidth} p{0.30\linewidth} p{0.30\linewidth}}
\toprule
 & \textbf{Control (channel zeroed)} & \textbf{Coupled} \\
\midrule
Surface-data-to-$\theta$ gradient & exactly zero, every logged step & nonzero, every logged step \\
Mean-field trajectory & reference & diverges up to $1.97\%$ pre-collapse \\
Final field & degenerate constant at lower bound & identical; per-specimen IoU differences exactly zero \\
\bottomrule
\end{tabular}
\end{sc}
\end{table}

\textbf{Outcome.} Each specimen's final IoU equals its own ground-truth defect volume fraction (arm mean $0.014$): the constant field at the lower bound marks every voxel as defect and recovers no structure. For context, NeFTY reaches $0.158$ on the same synthetic specimens. On this problem class, the directly coupled architecture did not by itself close the gap under the matched training protocol.

\textbf{Interpretation and scope.} Direct coupling does not remove the underlying ill-posedness of Section~\ref{sec:inverse}, but this experiment does not identify ill-posedness as the cause of the coupled variant's result: optimization, local conditioning, the absent boundary-condition term, the absent TV prior, and gradient conflict all remain live explanations.

\section{Method Details: Neural-Field Architecture, Forward Solver, and Adjoint Gradients}
\label{appdx:method}

This appendix records the implementation of the three components introduced in Section~\ref{sec:method}: the neural diffusivity field of Section~\ref{sec:nf}, the discrete forward solver of Section~\ref{sec:hardphys}, and the discrete adjoint of Section~\ref{sec:adjoint}. The harmonic-mean stencil derived in Appendix~\ref{appdx:effective} (Proposition~\ref{prop:harmonic}) and the boundary conditions described in Appendix~\ref{appdx:bc} are reused without further comment. Notation matches Section~\ref{sec:method} throughout: $\theta$ denotes the MLP parameter set, $\boldsymbol{\mu}^n$ the discrete adjoint variable at step $n$ (kept distinct from the Laplacian eigenvalues $\lambda_n$ of Proposition~\ref{prop:svd} and from the regularization weight $\lambda$ in~\eqref{eq:nefty_obj}), $\boldsymbol{\Pi}_\Gamma$ the observation projector onto $\Gamma_{\mathrm{obs}}$, and $\mathbf{A}(\alpha_\theta) := \mathbf{I} - \Delta t \, \mathbf{L}(\alpha_\theta)$ the implicit-Euler system matrix. The boldface discrete diffusion operator $\mathbf{L}(\alpha_\theta)$ is distinct from the italic lateral side length $L$ of Appendix~\ref{appdx:bc}, the boldface Jacobi diagonal $\mathbf{D}$ and off-diagonal $\mathbf{R}$ of Appendix~\ref{appdx:method_solver} are distinct from the italic spatial dimension $D$ and the calligraphic regularizer $\mathcal{R}$ of~\eqref{eq:nefty_obj}, and the boldface state residual $\mathbf{F}^n$ is distinct from the italic source $F_{\delta\alpha}$ of Appendix~\ref{appdx:illposed_lin}.

\subsection{Neural diffusivity-field architecture}
\label{appdx:method_arch}

\textbf{Coordinate MLP.} The diffusivity field is modeled as a fully connected network $f_\theta : \mathbb{R}^{d_\gamma} \to \mathbb{R}$ with depth $L_{\mathrm{MLP}}$, hidden width $W_{\mathrm{MLP}}$, ReLU activations~\cite{nair2010rectified}, and a single skip connection that re-injects the encoded coordinate $\gamma(\mathbf{x})$ at the middle layer to preserve gradient flow~\cite{mildenhall2021nerf}. Weights are initialized with Xavier scaling~\cite{glorot2010understanding}. Concrete values for $L_{\mathrm{MLP}}, W_{\mathrm{MLP}}$, and the skip-layer index are tabulated in Appendix~\ref{appdx:method_hparams}.

\textbf{Positional encoding.} Each input coordinate $\mathbf{x} \in \Omega \subset \mathbb{R}^D$ is normalized to $[-1, 1]^D$ and mapped to
\begin{equation}
\label{eq:posenc_appdx}
    \gamma(\mathbf{x}) \;=\; \big( \sin(2^0 \pi \mathbf{x}), \cos(2^0 \pi \mathbf{x}), \dots, \sin(2^{N-1} \pi \mathbf{x}), \cos(2^{N-1} \pi \mathbf{x}) \big) \;\in\; \mathbb{R}^{2 D N},
\end{equation}
yielding $d_\gamma = 2 D N$. The bandwidth $N$ controls the maximum spatial frequency the network can express. Raising $N$ improves the network's ability to resolve sharp interfaces but exposes it to the algebraic noise amplification of Corollary~\ref{cor:hadamard} along high-eigenmode left singular vectors.

\textbf{Frequency annealing.} To suppress this amplification during the early phase of optimization, we apply the cosine schedule of~\citet{park2021nerfies} to the encoding bands. Let $\beta(t) \in [0, N]$ be a training-step-dependent annealing parameter. The $k$-th band ($k = 0, \dots, N-1$) of $\gamma(\mathbf{x})$ is multiplied by
\begin{equation}
\label{eq:fa_schedule}
    w_k(\beta) \;=\; \tfrac{1}{2} \!\left[ 1 - \cos\!\big( \pi \, \mathrm{clamp}(\beta - k, 0, 1) \big) \right],
\end{equation}
so that $w_k = 0$ before the schedule reaches band $k$, ramps smoothly to $w_k = 1$ over a unit-width interval in $\beta$, and remains at one thereafter. We linearly increase $\beta$ from $0$ to $N$ over the first $T_{\mathrm{FA}}$ optimization steps (frequency-annealing horizon, value in Appendix~\ref{appdx:method_hparams}), after which all bands are fully unmasked. Compared with binary masking, the cosine schedule avoids abrupt gradient bursts when a new band becomes active and lets the network grow high-frequency detail only after the low-frequency reconstruction has stabilized.

\textbf{Bounded output.} The MLP output is hard-bracketed to a physically admissible range,
\begin{equation*}
    \alpha_\theta(\mathbf{x}) \;=\; \alpha_{\min} + (\alpha_{\max} - \alpha_{\min}) \, \sigma\!\big( f_\theta(\gamma(\mathbf{x})) \big),
\end{equation*}
with $\sigma(\cdot)$ the logistic sigmoid and bounds $0 < \alpha_{\min} < \alpha_{\max}$. The bracketing serves three purposes: (i) it enforces $\alpha_\theta(\mathbf{x}) > 0$ identically, satisfying the positivity required by the diffusivity form~\eqref{eq:diffusivity_form}; (ii) it bounds the diffusivity contrast so that the system matrix $\mathbf{A}(\alpha_\theta)$ remains well-conditioned for the Jacobi inner solver of Appendix~\ref{appdx:method_solver}; (iii) it prevents the early-training optimizer from probing extreme values that would yield zero gradients through the saturation of the sigmoid or through stiff numerical regimes. Concrete numerical values for $\alpha_{\min}, \alpha_{\max}$ are calibrated once per dataset using the bulk-to-defect contrast described in Appendix~\ref{appdx:bc}, and are reported in Appendix~\ref{appdx:method_hparams}.

\textbf{Discrete total-variation regularizer.} The continuous regularizer $\mathcal{R}(\alpha_\theta) = \int_\Omega \| \nabla \alpha_\theta(\mathbf{x}) \| \, d\mathbf{x}$ is implemented on the same Cartesian grid used by the forward solver, with per-axis dimensions $N_x, N_y, N_z$ and spacings $\Delta x, \Delta y, \Delta z$ formally defined in Appendix~\ref{appdx:method_solver}. With $\alpha_{i,j,k} := \alpha_\theta(\mathbf{x}_{i,j,k})$ at grid node $(i,j,k)$, we use the isotropic forward-difference discretization
\begin{multline}
\label{eq:tv_discrete}
    \mathcal{R}(\alpha_\theta) \;=\; \Delta x \, \Delta y \, \Delta z \!\!\sum_{i, j, k = 0}^{N_x - 1, \, N_y - 1, \, N_z - 1}\!\! \Bigg[ \big( \delta_x \alpha_{i,j,k} / \Delta x \big)^{\!2} \\
    + \big( \delta_y \alpha_{i,j,k} / \Delta y \big)^{\!2} + \big( \delta_z \alpha_{i,j,k} / \Delta z \big)^{\!2} + \epsilon_{\mathrm{TV}}^{\,2} \Bigg]^{1/2},
\end{multline}
where the per-axis forward differences are $\delta_x \alpha_{i,j,k} := \alpha_{(i+1) \bmod N_x, j, k} - \alpha_{i,j,k}$, $\delta_y \alpha_{i,j,k} := \alpha_{i, (j+1) \bmod N_y, k} - \alpha_{i,j,k}$ (lateral periodic wrap, matching the lateral periodic BCs of Appendix~\ref{appdx:bc}), and $\delta_z \alpha_{i,j,k} := \alpha_{i, j, k+1} - \alpha_{i,j,k}$ for $k < N_z - 1$ with $\delta_z \alpha_{i, j, N_z - 1} := 0$ (the adiabatic and Robin BCs do not constrain $\alpha$ at the front/back faces, so the through-thickness forward difference at the top slice vanishes, while the lateral $\delta_x, \delta_y$ contributions on that slice are retained). The smoothing constant $\epsilon_{\mathrm{TV}} > 0$ has dimension $[\alpha]/[\mathrm{length}]$ and ensures differentiability at vanishing gradients (concrete value in Appendix~\ref{appdx:method_hparams}, distinct from the perturbation parameter $\varepsilon$ used in Appendix~\ref{appdx:illposed_lin}). The penalty promotes piecewise-constant $\alpha_\theta$ aligned with the model assumed in Appendix~\ref{appdx:effective_correction} and supplements the high-eigenmode suppression provided by frequency annealing.

\subsection{Discrete forward solver}
\label{appdx:method_solver}

\textbf{Ambient shift.} Throughout this appendix and Section~\ref{sec:method}, the discrete temperature field $\mathbf{T}^n$ denotes the deviation from the ambient temperature $T_{\mathrm{amb}}$ defined in Appendix~\ref{appdx:bc}. Under this convention the synthetic post-flash and PVC near-uniform initial conditions of Appendix~\ref{appdx:bc} have ambient floor zero and the convective Robin condition $\mathbf{n} \cdot (\alpha \nabla T) = -h(T - T_{\mathrm{amb}})$ becomes the homogeneous form $\mathbf{n} \cdot (\alpha \nabla \mathbf{T}) = -h \, \mathbf{T}$. All measured frames $\hat{\mathbf{T}}_i$ in Section~\ref{sec:adjoint} are likewise interpreted as ambient-shifted observations $\hat{T}_i - T_{\mathrm{amb}}$. The state equation~\eqref{eq:state_eq} therefore retains the homogeneous form $\mathbf{A}(\alpha_\theta) \mathbf{T}^{n+1} = \mathbf{T}^n$ across both datasets, with no affine right-hand side.

\textbf{Spatial discretization.} The bulk equation $\partial_t T = \nabla \cdot (\alpha \nabla T)$ on the slab $\Omega = [0, L]^2 \times [0, H]$ is discretized on a uniform Cartesian grid with $N_x \times N_y \times N_z$ nodes and spacings $\Delta x, \Delta y, \Delta z$ aligned with the lateral and through-thickness axes. Continuing the one-dimensional argument of Proposition~\ref{prop:harmonic}, the second-order finite-volume diffusion operator $\mathbf{L}(\alpha_\theta) \in \mathbb{R}^{N_g \times N_g}$ acts at an interior node $(i, j, k)$ by
\begin{equation*}
\begin{aligned}
    {[\mathbf{L}(\alpha_\theta) \mathbf{T}]}_{i,j,k}
    \;=\; & \tfrac{1}{\Delta x^2} \!\left[ \bar{\alpha}_{i+1/2, j, k}(T_{i+1, j, k} - T_{i, j, k}) - \bar{\alpha}_{i-1/2, j, k}(T_{i, j, k} - T_{i-1, j, k}) \right] \\
    \;+\; & \tfrac{1}{\Delta y^2} \!\left[ \bar{\alpha}_{i, j+1/2, k}(T_{i, j+1, k} - T_{i, j, k}) - \bar{\alpha}_{i, j-1/2, k}(T_{i, j, k} - T_{i, j-1, k}) \right] \\
    \;+\; & \tfrac{1}{\Delta z^2} \!\left[ \bar{\alpha}_{i, j, k+1/2}(T_{i, j, k+1} - T_{i, j, k}) - \bar{\alpha}_{i, j, k-1/2}(T_{i, j, k} - T_{i, j, k-1}) \right],
\end{aligned}
\end{equation*}
with each face-centered coefficient $\bar{\alpha}$ computed as the harmonic mean of its two neighbouring cell-centered values. By Proposition~\ref{prop:harmonic} this is the unique discrete realization of effective-flux continuity for the single-$\alpha$ model.

\textbf{Boundary conditions.} Lateral periodicity is implemented by circular padding so that the stencil at $i = 0$ wraps to $i = N_x - 1$ and similarly for $j$. Through-thickness adiabatic conditions (zero-flux Neumann) are implemented by replicate padding on the front $z = 0$ and back $z = H$ faces, which enforces $\bar{\alpha}_{i, j, -1/2} (T_{i,j,0} - T_{i,j,-1}) = 0$ at the boundary face by zeroing the cross-face temperature increment. The PVC convective Robin condition described in Appendix~\ref{appdx:bc} replaces the back-face replicate padding by a ghost-cell update consistent with $\mathbf{n} \cdot (\alpha \nabla T) = -h(T - T_{\mathrm{amb}})$.

\textbf{Implicit-Euler step.} The state at frame $n+1$ is the solution of the sparse linear system
\begin{equation}
\label{eq:state_eq_appdx}
    \mathbf{A}(\alpha_\theta) \, \mathbf{T}^{n+1} \;=\; \mathbf{T}^n, \qquad \mathbf{A}(\alpha_\theta) \;:=\; \mathbf{I} - \Delta t \, \mathbf{L}(\alpha_\theta),
\end{equation}
which is unconditionally stable~\cite{courant1928partiellen} so that $\Delta t$ is set by the camera frame rate rather than the CFL bound $\Delta t \le \Delta x^2 / (2 D \, \alpha_{\max})$.

\textbf{Symmetry, sign, and diagonal dominance.} Under the homogeneous boundary conditions enforced by the ambient shift above (lateral periodic, adiabatic through-thickness, and the homogeneous Robin form $\mathbf{n} \cdot (\alpha \nabla \mathbf{T}) = -h \, \mathbf{T}$ on the PVC back face), the harmonic-mean construction makes $\mathbf{L}(\alpha_\theta)$ symmetric: each interior face conductance $a_{pq} := \bar{\alpha}_{pq} / \Delta x_d^2$ ($d \in \{x, y, z\}$ the axis joining grid nodes $p, q$) appears symmetrically in row $p$ and row $q$, lateral periodic wrapping preserves this symmetry, and a finite-volume back-face flux balance $\mathbf{n} \cdot (\alpha \nabla \mathbf{T}) \cdot \Delta x \Delta y = -h \, T_p \cdot \Delta x \Delta y$ divided by the nodal volume $\Delta x \Delta y \Delta z$ yields a strictly positive diagonal sink $\beta_p := h / \Delta z$ on each back-face Robin row. Writing $q \sim p$ for the set of interior-face neighbours of node $p$ and $\beta_p \ge 0$ for the per-row Robin sink (zero for non-Robin rows), the entries of $\mathbf{L}(\alpha_\theta)$ are
\begin{equation*}
\begin{aligned}
    {[\mathbf{L}(\alpha_\theta)]}_{pp} \;&=\; -\beta_p - \!\!\sum_{q \sim p} a_{pq}, & {[\mathbf{L}(\alpha_\theta)]}_{pq} \;&=\; a_{pq} \ge 0 \;\;\text{for}\;\; q \sim p, \\
    & & \sum_{q} {[\mathbf{L}(\alpha_\theta)]}_{pq} \;&=\; -\beta_p \;\le\; 0,
\end{aligned}
\end{equation*}
so the constant-temperature mode is annihilated on rows without a Robin sink and damped at rate $\beta_p$ on Robin boundary rows. Therefore $\mathbf{L}(\alpha_\theta)$ is symmetric and negative semi-definite under purely periodic/adiabatic BCs, and negative definite once at least one $\beta_p > 0$. Strict diagonal dominance of $\mathbf{A}(\alpha_\theta)$ follows from the identity term and the (non-negative) Robin sink:
\begin{equation*}
    {[\mathbf{A}(\alpha_\theta)]}_{pp} - \!\!\sum_{q \neq p} \big| {[\mathbf{A}(\alpha_\theta)]}_{pq} \big| \;=\; \big( 1 + \Delta t \beta_p + \Delta t \!\!\sum_{q \sim p} a_{pq} \big) - \Delta t \!\!\sum_{q \sim p} a_{pq} \;=\; 1 + \Delta t \beta_p \;\ge\; 1 \;>\; 0,
\end{equation*}
holding strictly on every row independently of the conductance pattern, so $\mathbf{A}(\alpha_\theta)$ is symmetric positive-definite and the Jacobi iteration below converges geometrically.

\textbf{Jacobi inner solver.} For autodiff compatibility on the GPU we avoid sparse-matrix factorizations and solve~\eqref{eq:state_eq_appdx} by a fixed unrolled number $K$ of Jacobi iterations. With the splitting $\mathbf{A}(\alpha_\theta) = \mathbf{D} + \mathbf{R}$ where $\mathbf{D}$ is the diagonal of $\mathbf{A}$ and $\mathbf{R}$ the off-diagonal remainder, the iteration with right-hand side $\mathbf{b} := \mathbf{T}^n$ is
\begin{equation}
\label{eq:jacobi_iter}
    \mathbf{T}^{(\kappa+1)} \;=\; \mathbf{D}^{-1} \!\left( \mathbf{b} - \mathbf{R} \, \mathbf{T}^{(\kappa)} \right), \qquad \kappa = 0, 1, \dots, K-1,
\end{equation}
initialized from $\mathbf{T}^{(0)} := \mathbf{T}^n$ (warm-started by the previous frame's solution). Equation~\eqref{eq:jacobi_iter} reduces to a 7-point stencil convolution, expressible as a single fused kernel in standard autodiff frameworks~\cite{paszke2019pytorch} and accelerated with \texttt{torch.compile} just-in-time autotuning. The iteration count $K$ is chosen large enough that the residual $\| \mathbf{A} \mathbf{T}^{(K)} - \mathbf{b} \|$ falls below the noise floor of the camera measurements (Appendix~\ref{appdx:method_hparams}). Empirically, $K = 50$ suffices for our grid resolution and $\Delta t$. Because the Jacobi iteration converges to the unique exact solution of~\eqref{eq:state_eq_appdx} under the strict diagonal dominance established above, we treat $\mathbf{T}^{(K)}$ as the (numerical) exact state $\mathbf{T}^{n+1}$, and the discrete adjoint of Appendix~\ref{appdx:method_adjoint} below is derived for this exact linear solve. The resulting parameter gradients are exact for the converged objective up to the same iterative tolerance.

\subsection{Adjoint recurrence and gradient assembly}
\label{appdx:method_adjoint}

\textbf{Constraint and augmented Lagrangian.} Set $\mathbf{T}^0$ from the prescribed initial condition (Appendix~\ref{appdx:bc}) and define the state residual at step $n$ by
\begin{equation}
\label{eq:residual_F}
    \mathbf{F}^n(\mathbf{T}^n, \mathbf{T}^{n-1}, \alpha_\theta) \;:=\; \mathbf{A}(\alpha_\theta) \, \mathbf{T}^n - \mathbf{T}^{n-1} \;=\; \mathbf{0}, \qquad n = 1, \dots, N_t.
\end{equation}
Define the per-step data term
\begin{equation*}
    \ell^n(\mathbf{T}^n) \;:=\; \begin{cases} \big\| \boldsymbol{\Pi}_\Gamma \mathbf{T}^n - \hat{\mathbf{T}}_i \big\|^2 & \text{if } n = n_i \in \{n_1, \dots, n_M\}, \\[2pt] 0 & \text{otherwise,} \end{cases}
\end{equation*}
so the data part of the objective is $\sum_{n=1}^{N_t} \ell^n(\mathbf{T}^n)$ and matches the data-fidelity sum in~\eqref{eq:nefty_obj}. The squared Euclidean data term in~\eqref{eq:nefty_obj} differs from the discrete squared $L^2(\Gamma_{\mathrm{obs}})$ term of~\eqref{eq:ihcp_obj} by the lateral quadrature factor $\Delta x \Delta y$, and rescaling the objective absorbs this factor into the TV weight $\lambda$. We introduce adjoint variables $\boldsymbol{\mu}^n \in \mathbb{R}^{N_g}$ for $n = 1, \dots, N_t$ as Lagrange multipliers of the state residuals, giving the augmented functional
\begin{equation*}
    \widetilde{\mathcal{L}}(\theta; \{\mathbf{T}^n\}, \{\boldsymbol{\mu}^n\}) \;=\; \sum_{n=1}^{N_t} \ell^n(\mathbf{T}^n) + \lambda \mathcal{R}(\alpha_\theta) - \sum_{n=1}^{N_t} (\boldsymbol{\mu}^n)^\top \mathbf{F}^n(\mathbf{T}^n, \mathbf{T}^{n-1}, \alpha_\theta).
\end{equation*}
For any $\{\mathbf{T}^n\}$ that satisfy the state equation~\eqref{eq:residual_F}, $\widetilde{\mathcal{L}} = \mathcal{J}_{\mathrm{disc}}(\theta)$ regardless of the multipliers $\boldsymbol{\mu}^n$.

\textbf{Stationarity and adjoint recurrence.} Computing $d \mathcal{J}_{\mathrm{disc}} / d \theta$ via the chain rule and discarding terms that are eliminated by the state equation, it suffices to compute $d \widetilde{\mathcal{L}} / d \theta$ in the way that does not require the implicit derivatives $d \mathbf{T}^n / d \theta$. To eliminate them, we choose the multipliers so that the partial derivative of $\widetilde{\mathcal{L}}$ with respect to each $\mathbf{T}^n$ vanishes. Differentiating,
\begin{equation*}
    \frac{\partial \widetilde{\mathcal{L}}}{\partial \mathbf{T}^n} \;=\; \frac{\partial \ell^n}{\partial \mathbf{T}^n} - (\boldsymbol{\mu}^n)^\top \, \frac{\partial \mathbf{F}^n}{\partial \mathbf{T}^n} - (\boldsymbol{\mu}^{n+1})^\top \, \frac{\partial \mathbf{F}^{n+1}}{\partial \mathbf{T}^n} \;=\; 0,
\end{equation*}
where we set $\boldsymbol{\mu}^{N_t+1} = \mathbf{0}$ because no constraint $\mathbf{F}^{N_t+1}$ exists. Reading off the partials from~\eqref{eq:residual_F},
\begin{equation*}
    \frac{\partial \mathbf{F}^n}{\partial \mathbf{T}^n} \;=\; \mathbf{A}(\alpha_\theta), \qquad \frac{\partial \mathbf{F}^{n+1}}{\partial \mathbf{T}^n} \;=\; -\mathbf{I},
\end{equation*}
the stationarity condition becomes
\begin{equation}
\label{eq:adjoint_recurrence_appdx}
    \mathbf{A}(\alpha_\theta)^\top \boldsymbol{\mu}^n \;=\; \big(\partial \ell^n / \partial \mathbf{T}^n\big)^\top + \boldsymbol{\mu}^{n+1}, \qquad \boldsymbol{\mu}^{N_t+1} \;=\; \mathbf{0},
\end{equation}
which is~\eqref{eq:adjoint_eq}. The recurrence is integrated backwards in time from $n = N_t$ down to $n = 1$. The same Jacobi inner solver of Appendix~\ref{appdx:method_solver} applies because the finite-volume discretization with diagonal Robin sink $\beta_p := h / \Delta z$ makes $\mathbf{A}(\alpha_\theta)^\top = \mathbf{A}(\alpha_\theta)$ on every dataset (lateral periodic, adiabatic through-thickness, and homogeneous Robin), so the forward and backward sweeps share the same SPD system matrix.

\textbf{Gradient assembly.} With $\boldsymbol{\mu}^n$ chosen so that $\partial \widetilde{\mathcal{L}} / \partial \mathbf{T}^n = 0$ for $n = 1, \dots, N_t$, the total derivative of $\mathcal{J}_{\mathrm{disc}}$ with respect to $\theta$ is the partial derivative of $\widetilde{\mathcal{L}}$ with respect to $\theta$ at fixed $\{\mathbf{T}^n\}$:
\begin{equation*}
    \frac{d \mathcal{J}_{\mathrm{disc}}}{d \theta} \;=\; \frac{\partial \widetilde{\mathcal{L}}}{\partial \theta} \;=\; \frac{\partial (\lambda \mathcal{R})}{\partial \theta} - \sum_{n=1}^{N_t} (\boldsymbol{\mu}^n)^\top \, \frac{\partial \mathbf{F}^n}{\partial \alpha_\theta} \, \frac{\partial \alpha_\theta}{\partial \theta}.
\end{equation*}
Because $\mathbf{F}^n = \mathbf{A}(\alpha_\theta) \mathbf{T}^n - \mathbf{T}^{n-1}$ and $\mathbf{A}(\alpha_\theta) = \mathbf{I} - \Delta t \, \mathbf{L}(\alpha_\theta)$ depends on $\alpha_\theta$ only through the diffusion operator,
\begin{equation*}
    \frac{\partial \mathbf{F}^n}{\partial \alpha_\theta} \;=\; -\Delta t \, \frac{\partial \!\left( \mathbf{L}(\alpha_\theta) \mathbf{T}^n \right)}{\partial \alpha_\theta},
\end{equation*}
so the assembled gradient becomes
\begin{equation}
\label{eq:gradient_assembly_appdx}
    \frac{d \mathcal{J}_{\mathrm{disc}}}{d \theta} \;=\; \frac{\partial (\lambda \mathcal{R})}{\partial \theta} \;+\; \Delta t \, \sum_{n=1}^{N_t} (\boldsymbol{\mu}^n)^\top \, \frac{\partial \!\left( \mathbf{L}(\alpha_\theta) \mathbf{T}^n \right)}{\partial \alpha_\theta} \, \frac{\partial \alpha_\theta}{\partial \theta},
\end{equation}
which is~\eqref{eq:grad_assembly}. The Jacobian $\partial \alpha_\theta / \partial \theta$ is obtained by ordinary backpropagation through the MLP at each grid node, and the term $\partial (\mathbf{L}(\alpha_\theta) \mathbf{T}^n) / \partial \alpha_\theta$ is a sparse local stencil whose nonzeros are determined by the harmonic-mean derivatives $\partial \bar{\alpha}_{i+1/2} / \partial \alpha_i = 2 \alpha_{i+1}^2 / (\alpha_i + \alpha_{i+1})^2$ and similarly for $\partial / \partial \alpha_{i+1}$.

\textbf{Memory cost.} The forward sweep stores $\mathbf{T}^1, \dots, \mathbf{T}^{N_t}$, about one field per frame, which the gradient assembly~\eqref{eq:gradient_assembly_appdx} consumes one frame at a time during the backward sweep that produces $\boldsymbol{\mu}^n$. Because~\eqref{eq:adjoint_recurrence_appdx} requires only $\boldsymbol{\mu}^{n+1}$ and $\mathbf{T}^n$ at any given step, the adjoint's solver working set (one current and one trailing slice plus the running gradient accumulator) is $\mathcal{O}(N_g)$, independent of $N_t$. The naive backpropagation-through-time alternative would store the entire computational graph of the $K$ unrolled Jacobi iterations at every one of the $N_t$ time steps, with peak memory scaling as $\mathcal{O}(K N_g N_t)$, whereas the adjoint replaces these retained iterates with the $\mathcal{O}(N_g)$ solver working set plus the stored states, at the cost of one backward solve, an exact restatement of the implicit-function-theorem trick in a sequential setting~\cite{cea1986conception}.

\textbf{Sensitivity to initial conditions.} If the initial temperature $\mathbf{T}^0$ is itself uncertain or learnable, the same calculation under the augmented Lagrangian above (which carries a $-\sum_n (\boldsymbol{\mu}^n)^\top \mathbf{F}^n$ term) gives
\begin{equation*}
    \frac{d \mathcal{J}_{\mathrm{disc}}}{d \mathbf{T}^0} \;=\; -\,(\boldsymbol{\mu}^1)^\top \, \frac{\partial \mathbf{F}^1}{\partial \mathbf{T}^0} \;=\; -\,(\boldsymbol{\mu}^1)^\top (-\mathbf{I}) \;=\; +\,\boldsymbol{\mu}^1,
\end{equation*}
where only $\mathbf{F}^1 = \mathbf{A} \mathbf{T}^1 - \mathbf{T}^0$ depends on $\mathbf{T}^0$. The sign is consistent with the adjoint recurrence~\eqref{eq:adjoint_recurrence_appdx} and the gradient assembly~\eqref{eq:gradient_assembly_appdx} under the same Lagrangian convention. The experiments of Section~\ref{sec:experiments} keep $\mathbf{T}^0$ fixed at the post-flash initial profile calibrated externally per Appendix~\ref{appdx:bc}.

\textbf{Consistency with the continuous adjoint.} The recurrence~\eqref{eq:adjoint_recurrence_appdx} is the implicit-Euler discretization, applied backwards in time, of the continuous adjoint problem associated with the bulk equation $\partial_t T = \nabla \cdot (\alpha \nabla T)$ and the cumulative observation functional $\sum_i \| \boldsymbol{\Pi}_\Gamma T(\cdot, t_i) - \hat{T}_i \|^2$: namely, $-\partial_t p = \nabla \cdot (\alpha \nabla p) + \sum_i 2 \boldsymbol{\Pi}_\Gamma^*(\boldsymbol{\Pi}_\Gamma T(\cdot, t_i) - \hat{T}_i) \, \delta(t - t_i)$ with terminal condition $p(\cdot, t_{\mathrm{end}}) = 0$. The discretize-then-optimize formulation we use thus inherits the same backward-parabolic structure as the continuous setting~\cite{onken2020discretize}, while delivering exact gradients of the discrete objective $\mathcal{J}_{\mathrm{disc}}$ rather than discretized gradients of the continuous one.

\subsection{Implementation hyperparameters}
\label{appdx:method_hparams}

The hyperparameters used in the synthetic experiments of Section~\ref{sec:experiments} are listed in Table~\ref{tab:hyperparameters}. Adam is used with the listed learning rate and step decay. Gradients are accumulated through a single forward and a single backward solver sweep per iteration. Compute resources, including the GPUs used for benchmarking, are documented in Appendix~\ref{appdx:experimental}.
\begin{table}[h]
    \centering
    \caption{Hyperparameter configuration for NeFTY on the synthetic benchmark.}
    \label{tab:hyperparameters}
    \vspace{0.05in}
    \begin{small}
    \begin{tabular}{l l c}
        \toprule
        \textbf{Category} & \textbf{Parameter} & \textbf{Value} \\
        \midrule
        \textbf{Network architecture} & MLP depth $L_{\mathrm{MLP}}$ & 10 \\
        & MLP hidden width $W_{\mathrm{MLP}}$ & 512 \\
        & Encoding bandwidth $N$ & 12 \\
        & Skip-connection layer & 4 \\
        & Output activation & scaled sigmoid \\
        \midrule
        \textbf{Domain and grid} & Domain size $(L, L, H)$ & $10.0 \times 10.0 \times 1.0$ \\
        & Grid resolution $(N_x, N_y, N_z)$ & $64 \times 64 \times 16$ \\
        & Lateral spacing $\Delta x = \Delta y$ & $0.156$ \\
        & Through-thickness spacing $\Delta z$ & $0.0625$ \\
        \midrule
        \textbf{Forward solver} & Time step $\Delta t$ & $0.05$ \\
        & Number of frames $N_t$ & $100$ \\
        & Time integrator & implicit Euler \\
        & Inner solver & Jacobi unrolling \\
        & Jacobi iterations $K$ & $50$ \\
        & Face coefficient & harmonic mean (Prop.~\ref{prop:harmonic}) \\
        & Diffusivity bounds $(\alpha_{\min}, \alpha_{\max})$ & $(0.003, \, 0.25)$ \\
        \midrule
        \textbf{Optimization} & Optimizer & Adam \\
        & Learning rate & $5 \times 10^{-5}$ \\
        & Step decay $\gamma$ & $0.1$ per $1000$ steps \\
        & Total iterations & $10{,}000$ \\
        & Frequency-annealing horizon $T_{\mathrm{FA}}$ & $2{,}500$ \\
        \midrule
        \textbf{Regularization} & TV weight $\lambda$ & $1 \times 10^{-3}$ \\
        & TV smoothing $\epsilon$ & $1 \times 10^{-6}$ \\
        \bottomrule
    \end{tabular}
    \end{small}
\end{table}

\section{Experimental Details}
\label{appdx:experimental}

\subsection{Synthetic dataset generation}
\label{appdx:experimental_data}

\textbf{Inverse-crime mitigation.} To avoid evaluating the reconstruction against data generated by its own discretization, we use PhiFlow~\cite{holl2024bf}, an independent finite-volume engine, for ground-truth generation. PhiFlow integrates the bulk equation~\eqref{eq:diffusivity_form} with an explicit scheme. To overcome the explicit stability constraint at fine grid resolution, we apply an adaptive substepping routine. The maximum stable step size is
\begin{equation*}
\Delta t_{\mathrm{stable}} \;\approx\; \frac{\Delta x^2}{2 D \, \alpha_{\max}},
\end{equation*}
with $D \in \{2,3\}$ the spatial dimension and $\alpha_{\max}$ the per-batch maximum diffusivity, and the number of solver substeps per recorded frame is set to
\begin{equation*}
N_{\mathrm{sub}} \;=\; \max\!\left( 10, \; \left\lceil \frac{\Delta t}{\Delta t_{\mathrm{stable}}} \times 2.0 \right\rceil \right),
\end{equation*}
so that the forward simulation often executes dozens of substeps per recorded frame.

\textbf{Simulation configuration.} All synthetic samples are generated on the unitless slab $\Omega = [0,10]^2 \times [0,1]$ at $64 \times 64 \times 16$ grid resolution. The lateral $(x, y)$ faces use periodic boundary conditions and the through-thickness $(z)$ faces use adiabatic Neumann conditions, both consistent with Appendix~\ref{appdx:bc}. The transient is recorded over $100$ frames with $\Delta t = 0.05$. Material properties are sampled from uniform distributions: bulk diffusivity $\alpha_{\mathrm{base}} \sim \mathcal{U}(0.1, 0.2)$, defect diffusivity $\alpha_{\mathrm{defect}} \sim \mathcal{U}(0.005, 0.015)$. Each sample contains one to four ellipsoidal, cylindrical, or box-shaped subsurface defects buried at varying depth. The dataset of $1{,}000$ samples is split into a homogeneous configuration (constant $\alpha_{\mathrm{base}}$) and a layered configuration (three to four bulk strata along $z$), and we evaluate every method on the full $1{,}000$-sample benchmark per configuration.

\textbf{Dimensional analysis.} The unitless solver evolves in $(t / t_{\mathrm{total}}, \mathbf{x} / L_0)$ coordinates with effective diffusivity $\alpha_{\mathrm{sim}} = \alpha_{\mathrm{phys}}\, t_{\mathrm{total}} / L_0^2$, the Fourier-number scaling of Appendix~\ref{appdx:bc}. A simulation with $\alpha_{\mathrm{sim}} \approx 0.1$ corresponds to highly conductive silicon ($\alpha_{\mathrm{phys}} \approx 10^{-4}\,\mathrm{m}^2/\mathrm{s}$) on a microsecond timescale at $L_0 = 10\,\mu\mathrm{m}$, or to a resistive polymer ($\alpha_{\mathrm{phys}} \approx 10^{-7}\,\mathrm{m}^2/\mathrm{s}$) on a millisecond timescale at the same length scale. A single set of unitless parameters thus represents widely different material classes by reinterpreting $L_0$ and $t_{\mathrm{total}}$.

\textbf{Defect contrast scaling.} The defect-to-bulk diffusivity ratio is set to roughly $1\!:\!20$. We choose this contrast over the realistic air-to-solid ratio of $> 1\!:\!1000$ for numerical stability: at $1\!:\!1000$ the linear system $\mathbf{A}(\alpha_\theta) = \mathbf{I} - \Delta t\, \mathbf{L}(\alpha_\theta)$ becomes severely ill-conditioned and the iterative solver stalls. At $1\!:\!20$ the surface-temperature signature already saturates against the perfect-insulator limit (further reductions in $\alpha_{\mathrm{defect}}$ leave the boundary thermograms unchanged), while the condition number remains compatible with efficient gradient-based optimization.

\subsection{Metrics}
\label{appdx:experimental_metrics}

\textbf{Synthetic 3D reconstruction.} For each test sample, let $\alpha^{\star}$ denote the ground-truth diffusivity field on the $N_g = N_x N_y N_z$ grid and $\hat{\alpha}$ the recovered field. We report:
\begin{itemize}[leftmargin=*,topsep=2pt,itemsep=1pt]
    \item \textit{MSE} $= \frac{1}{N_g} \| \hat{\alpha} - \alpha^{\star} \|_2^2$.
    \item \textit{PSNR} $= 10 \log_{10}\!\big( (\alpha_{\max} - \alpha_{\min})^2 / \mathrm{MSE} \big)$.
    \item \textit{SSIM}~\cite{wang2004image} computed slice-by-slice along $z$ and averaged.
    \item \textit{IoU} $= |\,(\hat{\alpha} < \tau) \cap (\alpha^{\star} < \tau)\,| / |\,(\hat{\alpha} < \tau) \cup (\alpha^{\star} < \tau)\,|$ with defect threshold $\tau = 0.03$, chosen as twice the upper end of the defect distribution to provide a clean binary defect mask.
\end{itemize}
The metrics are computed per sample and averaged over the full $1{,}000$-sample benchmark per configuration.

\textbf{Real PVC: 2D segmentation.} Following the convention of the PVC-Infrared benchmark~\cite{wei2023pulsed}, the recovered 3D $\hat{\alpha}_\theta$ is projected onto a 2D defect mask through the depth-averaged anomaly contrast described in Appendix~\ref{appdx:pvc}, and IoU and Dice are computed against the 2D ground-truth annotation.

\textbf{Real PVC: 2.5D depth.} For each pixel $p$ assigned to a defect, let $d^{\star}_p$ denote the ground-truth depth (in mm) supplied by the PVC-Depth benchmark~\cite{wei2023depth} and $\hat{d}_p$ the predicted depth from the spatial-median anomaly contrast along $z$ (Appendix~\ref{appdx:pvc}). Let $P$ denote the set of pixels assigned to a defect by the 2D mask above and $\delta_p := \max(\hat{d}_p / d^{\star}_p, d^{\star}_p / \hat{d}_p)$ the per-pixel depth ratio. We report the standard depth metrics of~\citet{eigen2014depth}: Absolute Relative error $\mathrm{Abs\,Rel} = \frac{1}{|P|} \sum_{p \in P} |\hat{d}_p - d^{\star}_p| / d^{\star}_p$, RMSE $= \big( \frac{1}{|P|} \sum_{p \in P} (\hat{d}_p - d^{\star}_p)^2 \big)^{1/2}$, and the threshold accuracies $\delta < 1.25^k := \frac{1}{|P|} | \{ p \in P : \delta_p < 1.25^k \} |$ for $k \in \{1, 2, 3\}$.

\textbf{Frequency-domain diagnostics.} \textit{Defect Edge F1} thresholds the gradient magnitude $\| \nabla \alpha \|$ at $50\%$ of the ground-truth maximum, restricts to a dilated defect-boundary mask, and reports the harmonic mean of precision and recall. \textit{Radial power spectrum} extracts a tight $16 \times 16 \times 8$ crop centered on the defect cluster, computes the 3D Fourier transform, and radially averages the power, with frequencies normalized so that the through-thickness Nyquist limit equals $\sim 8$ cycles per unit length.

\subsection{Hyperparameters and compute resources}
\label{appdx:experimental_hparams}

The full NeFTY hyperparameter table (architecture, solver, optimization, regularization) is given in Table~\ref{tab:hyperparameters} of Appendix~\ref{appdx:method_hparams}. Our experimental framework was executed on a hybrid infrastructure comprising local workstations for controlled benchmarking and a high-performance computing (HPC) cluster for large-scale training. The local development environment consists of servers equipped with a 32-core CPU and two NVIDIA RTX PRO 6000 Blackwell GPUs (96\,GB VRAM each). For consistency across the efficiency analysis, all hardware-sensitive metrics, specifically the wall-clock times and peak GPU memory usage detailed in Table~\ref{tab:efficiency_main} of the main body and Table~\ref{tab:training_benchmark} of this appendix, were benchmarked exclusively on this local server with reconstruction confined to a single GPU per specimen. For the large-scale synthetic dataset generation under PhiFlow, we utilized a compute cluster where each node is provisioned with dual 26-core CPUs and eight NVIDIA L40 GPUs.

\section{Baseline Implementations}
\label{appdx:baselines}

This appendix gives the configuration of every baseline used in Section~\ref{sec:experiments}. Unless otherwise stated, all baselines are tuned with the same compute budget as NeFTY on the same hardware (Appendix~\ref{appdx:experimental_hparams}).

\subsection{Soft-constrained PINN family}
\label{appdx:baselines_pinn}

\textbf{PINN with GradNorm.} The vanilla PINN baseline instantiates two networks, a temperature surrogate $T_\phi(\mathbf{x}, t)$ and a diffusivity surrogate $\alpha_\theta(\mathbf{x})$, and minimizes the composite loss of equation~\eqref{eq:pinn_loss}. The PDE residual is evaluated through automatic differentiation at $N_c$ space-time collocation points sampled uniformly per iteration, and the initial-condition penalty matches $T_\phi(\mathbf{x}, 0)$ to the prescribed Gaussian post-flash profile of Appendix~\ref{appdx:bc}. The two physics weights $\lambda_{\mathrm{PDE}}$ and $\lambda_{\mathrm{IC}}$ are dynamically balanced through GradNorm~\cite{chen2018gradnorm}, which normalizes per-loss gradient magnitudes to a common scale. We use $T_\phi$ depth $5$ and width $128$ (4.91\,M total parameters across both networks), $N_c = 24{,}576$ collocation points per iteration, Adam with learning rate $10^{-3}$, and $22{,}000$ iterations: $2.2\times$ more iterations and $2\times$ more parameters than NeFTY. Despite this advantage, the recovered $\alpha_\theta$ saturates at a near-trivial constant (Section~\ref{ssec:synth}), in line with the structural decoupling proven in Appendix~\ref{appdx:pinn}.

\textbf{SPINN, Causal-PINN, DCGD.} For the three modern PINN variants we follow the official implementations and hyperparameter settings released by the original authors~\cite{cho2023spinn,wang2024causal,hwang2024dcgd}, paired with the same diffusivity-network $\alpha_\theta$ and the same $24{,}576$ collocation points and $22{,}000$ iterations as the vanilla PINN above so that compute is matched across the family.

\subsection{Voxel-grid optimization}
\label{appdx:baselines_grid}

The Grid Opt.\ baseline replaces the neural diffusivity field $\alpha_\theta$ of Section~\ref{sec:nf} by a learnable tensor $\boldsymbol{\alpha} \in \mathbb{R}^{N_x \times N_y \times N_z}$ (initialized at the bulk mean) and optimizes it through the same differentiable solver and adjoint of Section~\ref{sec:method}. This isolates the contribution of the continuous coordinate-based prior. Grid Opt.\ uses the same Adam learning-rate schedule, the same number of iterations ($10{,}000$), and the same TV regularizer as NeFTY. Without the neural prior, it converges to a noisy field with ringing artifacts (Section~\ref{ssec:synth}).

\subsection{Supervised U-Net baselines}
\label{appdx:baselines_unet}

\textbf{U-Net (Full).} A 3D U-Net~\cite{ronneberger2015u} adapted for spatiotemporal regression. The input temperature sequence of shape $(B, 1, 100, 64, 64)$ is interpolated along the temporal axis from $100$ to $16$ depth slices and treated as a $(B, 1, 16, 64, 64)$ volume. The architecture follows a four-level encoder-decoder with channel sizes $\{32, 64, 128, 256\}$, double 3D convolutions and max-pooling on the contracting path, and trilinear upsampling with skip connections on the expansive path. The output passes through a sigmoid scaled to $[\alpha_{\min}, \alpha_{\max}]$. Training uses MSE between predicted and ground-truth $\alpha$ on the full synthetic dataset (defects included), which represents an upper bound consuming the volumetric labels that the label-free baselines never see.

\textbf{U-Net (Sound-Only).} The same architecture trained exclusively on defect-free samples. This baseline probes the susceptibility of purely data-driven inversion to out-of-distribution defects, and is included in the main Table~\ref{tab:synth_main} as a data-driven zero-shot reference.

\subsection{Classical thermography heuristics}
\label{appdx:baselines_classical}

\textbf{PPT.} Pulsed Phase Thermography~\cite{maldague2002advances,ibarra2004ppt} computes the per-pixel discrete Fourier transform of the surface-temperature decay, identifies the dominant phase peak at frequency $f_{\mathrm{peak}}$, and inverts depth through the diffusion-length formula $d = 1.82\, \sqrt{\alpha_{\mathrm{phys}} / (\pi f_{\mathrm{peak}})}$. We use the implementation of the original authors with six target frequencies spanning $0.01$-$0.5$\,Hz, matching the PVC-Depth recording rate.

\textbf{TSR.} Thermographic Signal Reconstruction~\cite{shepard2002reconstruction,shepard2015advances} fits a polynomial in $\log t$ to each pixel's temperature decay, identifies the time $t_{\mathrm{peak}}$ at which the second log-derivative peaks, and inverts depth as $d = 1.8\, \sqrt{\alpha_{\mathrm{phys}}\, t_{\mathrm{peak}}}$. We use the standard fifth-order polynomial fit and the per-pixel $t_{\mathrm{peak}}$ extraction.

\subsection{Swin-UNETR (real PVC, zero-shot transfer)}
\label{appdx:baselines_swin}

The supervised Swin-UNETR baseline of~\citet{hatamizadeh2022swin} is trained on the synthetic dataset of Appendix~\ref{appdx:experimental_data} for $500$ epochs under MSE supervision (loss converges to $4 \times 10^{-6}$) and transferred zero-shot to the real PVC datasets. Inputs are temporally resampled from the PVC frame rate to the synthetic $100$-frame format and amplitude-normalized to the synthetic peak-temperature scale. The predicted 3D $\hat{\alpha}$ is projected to 2D masks and 2.5D depth through the same procedure used for NeFTY (Appendix~\ref{appdx:pvc}).

\section{Additional Results}
\label{appdx:add_results}

\begin{figure*}[h]
  \centering
  \includegraphics[width=0.96\linewidth]{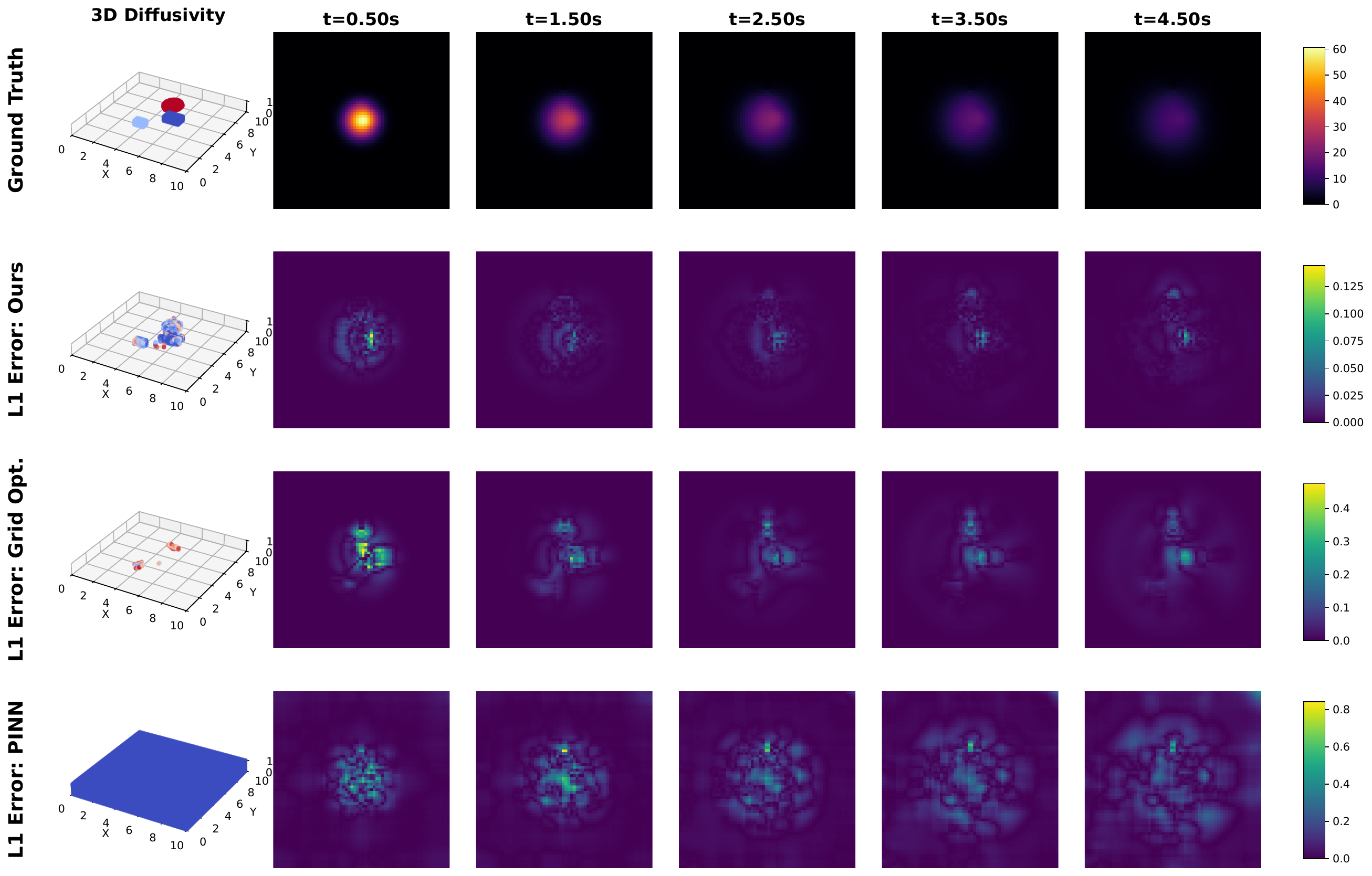}
  \captionsetup{belowskip=-4pt}
  \caption{\textbf{Surface-temperature error maps over time} for a representative three-defect specimen. NeFTY (Row 2) attains the lowest residual error throughout the decay, while Grid Opt.\ (Row 3) and PINN (Row 4) exhibit structured residuals that persist and diffuse outward, signaling that their reconstructed interiors are inconsistent with the observed boundary thermograms.}
  \label{fig:surface_error_maps}
\end{figure*}

This appendix collects the supporting evidence referenced from Section~\ref{sec:experiments}: the per-difficulty robustness breakdown including the supervised U-Net (Full) upper bound (Table~\ref{tab:robustness_breakdown}), the surface-fidelity decomposition that quantifies the data-fit paradox of Section~\ref{ssec:synth} (Table~\ref{tab:surface_fidelity} and Figure~\ref{fig:surface_error_maps}), the cumulative ablation in full (Table~\ref{tab:ablation_full}), the training-level wall-clock comparison referenced from Section~\ref{ssec:efficiency} (Table~\ref{tab:training_benchmark}), the frequency-domain diagnostics (Edge F1 and radial power spectrum), additional per-defect-count and per-layer-count qualitative reconstructions (Figures~\ref{fig:qual_1defect}--\ref{fig:qual_4layer}), and a representative failure mode (Figure~\ref{fig:failure_mode}). It then reports a classical PDE-constrained inversion baseline (Appendix~\ref{appdx:classical}), robustness studies on irregular geometry, observation noise, and material contrast (Appendices~\ref{appdx:irregular}--\ref{appdx:contrast}), and runtime and memory scaling with grid size and frame count (Appendix~\ref{appdx:scaling}).

\subsection{Robustness across defect count and layer count}

\begin{table*}[h]
\centering
\caption{\textbf{Robustness breakdown of Table~\ref{tab:synth_main}.} PSNR$\uparrow$ and IoU$\uparrow$ stratified by defect count (Homogeneous, $1$ to $4$ defects) and layer count (Layered Composite, $3$ and $4$ layers, each with $1$ to $4$ defects). The supervised U-Net (Full) upper bound is included here, as it consumes the volumetric labels the label-free baselines never see. Best label-free entry per column in \textbf{bold}.}
\label{tab:robustness_breakdown}
\begin{sc}
\resizebox{\textwidth}{!}{
\begin{tabular}{l cc cc cc cc | cc cc}
\toprule
 & \multicolumn{8}{c|}{\textbf{Homogeneous}} & \multicolumn{4}{c}{\textbf{Layered Composite}} \\
 & \multicolumn{2}{c}{$1$ Defect} & \multicolumn{2}{c}{$2$ Defects} & \multicolumn{2}{c}{$3$ Defects} & \multicolumn{2}{c|}{$4$ Defects} & \multicolumn{2}{c}{$3$ Layers} & \multicolumn{2}{c}{$4$ Layers} \\
\cmidrule(lr){2-3} \cmidrule(lr){4-5} \cmidrule(lr){6-7} \cmidrule(lr){8-9} \cmidrule(lr){10-11} \cmidrule(lr){12-13}
\textbf{Method} & PSNR & IoU & PSNR & IoU & PSNR & IoU & PSNR & IoU & PSNR & IoU & PSNR & IoU \\
\midrule
\multicolumn{13}{l}{\textit{Supervised}} \\
U-Net (Full) & $27.98$ & $0.72$ & $24.99$ & $0.72$ & $22.97$ & $0.69$ & $20.84$ & $0.66$ & $20.81$ & $0.67$ & $19.26$ & $0.68$ \\
U-Net (Sound-Only) & $18.96$ & $0.00$ & $15.95$ & $0.00$ & $13.19$ & $0.00$ & $11.23$ & $0.00$ & $15.62$ & $0.00$ & $15.22$ & $0.00$ \\
\midrule
\multicolumn{13}{l}{\textit{Label-free}} \\
Grid Opt. & $17.41$ & $0.03$ & $14.01$ & $0.01$ & $13.04$ & $0.06$ & $11.51$ & $0.07$ & $13.15$ & $0.04$ & $13.40$ & $0.02$ \\
PINN & $-0.37$ & $0.01$ & $-0.30$ & $0.01$ & $-0.18$ & $0.02$ & $-0.13$ & $0.02$ & $1.19$ & $0.02$ & $1.64$ & $0.02$ \\
SPINN & $10.48$ & $0.000$ & $\phantom{0}9.42$ & $0.000$ & $\phantom{0}9.53$ & $0.000$ & $\phantom{0}7.04$ & $0.004$ & $\phantom{0}9.59$ & $0.000$ & $\phantom{0}9.84$ & $0.000$ \\
\textbf{NeFTY (Ours)} & $\mathbf{19.99}$ & $\mathbf{0.40}$ & $\mathbf{19.36}$ & $\mathbf{0.51}$ & $\mathbf{17.55}$ & $\mathbf{0.43}$ & $\mathbf{17.04}$ & $\mathbf{0.44}$ & $\mathbf{16.69}$ & $\mathbf{0.41}$ & $\mathbf{15.07}$ & $\mathbf{0.34}$ \\
\bottomrule
\end{tabular}
}
\end{sc}
\end{table*}

NeFTY's IoU stays in the $0.34$-$0.51$ band across defect-count and layer-count slices, while Grid Opt.\ collapses below $0.07$ and PINN stays at $\sim 0.02$. SPINN, the strongest of the four PINN variants on the aggregate SSIM (Table~\ref{tab:synth_main}), still recovers IoU $\le 0.004$ in every slice and its PSNR drifts down with defect density (from $10.48$ at one defect to $7.04$ at four defects), so the soft-PINN collapse persists across every defect-count and layer-count slice. The supervised U-Net (Full) peaks at IoU $0.72$ on the simplest setting and degrades only mildly with complexity, providing the upper bound referenced from Section~\ref{ssec:synth}. NeFTY narrows the gap between the label-free baselines and this upper bound in every slice without using labels.

\subsection{Surface fidelity and the data-fit paradox}

\begin{table*}[h]
\centering
\caption{\textbf{Surface-temperature reconstruction error.} For each method we re-simulate the temperature with the recovered $\alpha_\theta$ (or directly read off $T_\phi$ for the soft-PINN family) and compare to the ground-truth surface frames. Comparing with the volumetric IoU of Table~\ref{tab:robustness_breakdown} surfaces the data-fit paradox of Section~\ref{ssec:synth}: PINN attains a surface PSNR above $62$\,dB while its IoU stays at $\sim 0.01$. MSE is scaled by $10^{-4}$.}
\label{tab:surface_fidelity}
\begin{sc}
\resizebox{\textwidth}{!}{
\begin{tabular}{l cc cc cc cc | cc cc}
\toprule
 & \multicolumn{8}{c|}{\textbf{Homogeneous}} & \multicolumn{4}{c}{\textbf{Layered Composite}} \\
 & \multicolumn{2}{c}{$1$ Defect} & \multicolumn{2}{c}{$2$ Defects} & \multicolumn{2}{c}{$3$ Defects} & \multicolumn{2}{c|}{$4$ Defects} & \multicolumn{2}{c}{$3$ Layers} & \multicolumn{2}{c}{$4$ Layers} \\
\cmidrule(lr){2-3} \cmidrule(lr){4-5} \cmidrule(lr){6-7} \cmidrule(lr){8-9} \cmidrule(lr){10-11} \cmidrule(lr){12-13}
\textbf{Method} & MSE$\downarrow$ & PSNR$\uparrow$ & MSE$\downarrow$ & PSNR$\uparrow$ & MSE$\downarrow$ & PSNR$\uparrow$ & MSE$\downarrow$ & PSNR$\uparrow$ & MSE$\downarrow$ & PSNR$\uparrow$ & MSE$\downarrow$ & PSNR$\uparrow$ \\
\midrule
Grid Opt. & $4.82$ & $75.51$ & $7.02$ & $71.56$ & $9.14$ & $71.03$ & $29.84$ & $67.36$ & $15.94$ & $71.14$ & $6.61$ & $71.95$ \\
PINN & $43.89$ & $63.04$ & $45.10$ & $62.82$ & $52.12$ & $62.20$ & $52.65$ & $62.19$ & $54.28$ & $62.05$ & $55.39$ & $61.95$ \\
\textbf{NeFTY (Ours)} & $\mathbf{0.50}$ & $\mathbf{82.33}$ & $\mathbf{0.52}$ & $\mathbf{82.26}$ & $\mathbf{0.73}$ & $\mathbf{81.34}$ & $\mathbf{0.56}$ & $\mathbf{82.17}$ & $\mathbf{0.54}$ & $\mathbf{82.10}$ & $\mathbf{0.50}$ & $\mathbf{82.42}$ \\
\bottomrule
\end{tabular}
}
\end{sc}
\end{table*}

The PINN baseline attains a surface MSE of $43.89 \times 10^{-4}$ and a PSNR of $63.04$\,dB on the single-defect setting yet a volumetric IoU of $0.01$, the data-fit paradox anticipated by Appendix~\ref{appdx:pinn_coupling}: the soft-constraint formulation can fit the boundary thermograms with a non-physical interior. NeFTY is forced through the hard solver to keep its interior consistent with the boundary, and consequently attains the lowest surface MSE ($0.50 \times 10^{-4}$) and the highest PSNR ($82.33$\,dB) while also recovering the correct internal structure.

\begin{figure}[h]
  \centering
  \includegraphics[width=0.92\linewidth]{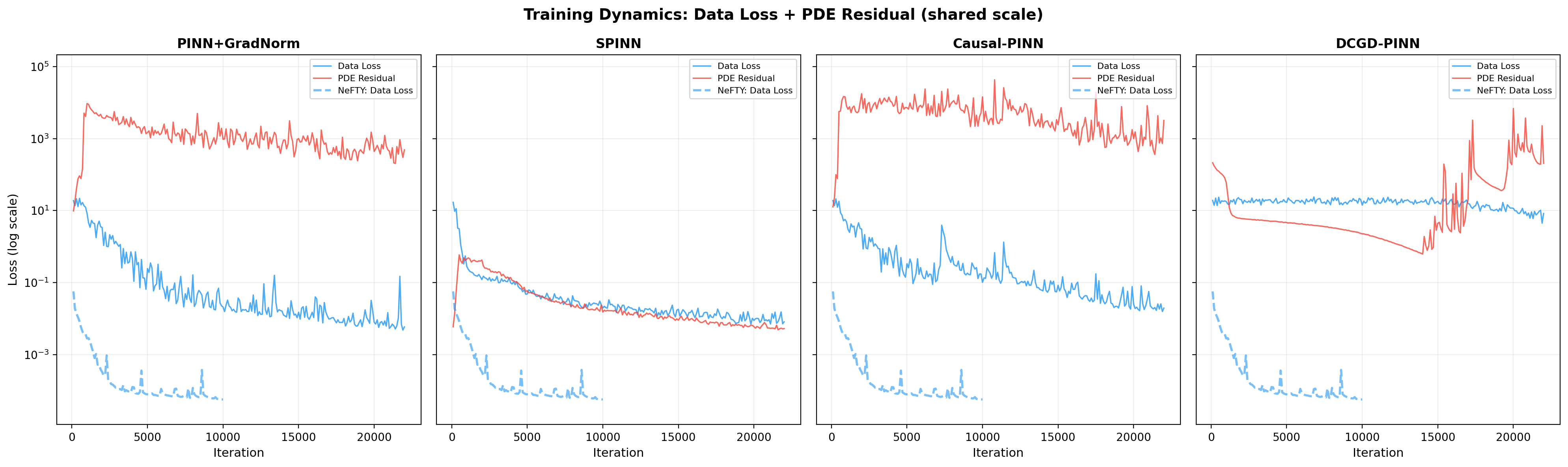}
  \captionsetup{belowskip=-4pt}
  \caption{\textbf{Training dynamics of the four PINN variants compared to NeFTY} (representative homogeneous specimen, identical $y$-axis scale across panels). Solid blue: PINN data loss; red: PDE residual; dashed blue: NeFTY data loss. Across all four variants the PDE residual is orders of magnitude larger than the data term and highly noisy. The optimizer drives the data loss down while the diffusivity network saturates at a near-trivial solution. NeFTY's data loss (dashed) reaches orders of magnitude below all PINN data losses, since the hard solver couples surface fitting and diffusivity recovery directly.}
  \label{fig:pinn_loss_curves}
\end{figure}

\subsection{Cumulative ablation, complete}

\begin{table*}[h]
\centering
\caption{\textbf{Cumulative ablation in full.} Each row adds one component to the previous configuration, mean $\pm$ $95\%$ CI across the $1{,}000$-sample synthetic benchmark. The compact summary version is Table~\ref{tab:ablation_main} in the main body.}
\label{tab:ablation_full}
\begin{sc}
\resizebox{\textwidth}{!}{
\begin{tabular}{l cccc cccc}
\toprule
 & \multicolumn{4}{c}{\textbf{Homogeneous}} & \multicolumn{4}{c}{\textbf{Layered}} \\
\cmidrule(lr){2-5} \cmidrule(lr){6-9}
\textbf{Variant} & MSE ($10^{-4}$)$\downarrow$ & PSNR$\uparrow$ & SSIM$\uparrow$ & IoU$\uparrow$ & MSE ($10^{-4}$)$\downarrow$ & PSNR$\uparrow$ & SSIM$\uparrow$ & IoU$\uparrow$ \\
\midrule
Base & $127.4 \pm 92.2$ & $0.47 \pm 3.17$ & $0.28 \pm 0.05$ & $0.03 \pm 0.02$ & $221.6 \pm 130.5$ & $2.89 \pm 1.88$ & $0.24 \pm 0.05$ & $0.02 \pm 0.02$ \\
\,+\,PE & $126.1 \pm 59.4$ & $4.17 \pm 1.36$ & $0.12 \pm 0.02$ & $0.09 \pm 0.03$ & $87.16 \pm 29.71$ & $6.15 \pm 1.11$ & $0.13 \pm 0.02$ & $0.09 \pm 0.03$ \\
\,+\,PE,\,FA & $29.80 \pm 4.78$ & $8.33 \pm 0.59$ & $0.16 \pm 0.02$ & $0.14 \pm 0.03$ & $44.33 \pm 15.94$ & $8.81 \pm 0.87$ & $0.18 \pm 0.03$ & $0.14 \pm 0.03$ \\
\,+\,PE,\,FA,\,$\sigma$ & $31.43 \pm 8.58$ & $9.19 \pm 0.94$ & $0.26 \pm 0.03$ & $0.18 \pm 0.04$ & $35.60 \pm 6.07$ & $9.27 \pm 0.82$ & $0.24 \pm 0.04$ & $0.14 \pm 0.04$ \\
\,+\,PE,\,FA,\,$\sigma$,\,HM & $21.01 \pm 5.67$ & $10.95 \pm 0.94$ & $0.36 \pm 0.03$ & $0.24 \pm 0.05$ & $23.09 \pm 4.77$ & $11.26 \pm 0.84$ & $0.33 \pm 0.05$ & $0.22 \pm 0.06$ \\
\textbf{Full (\,+\,TV)} & $\mathbf{3.66 \pm 1.31}$ & $\mathbf{18.48 \pm 0.53}$ & $\mathbf{0.77 \pm 0.02}$ & $\mathbf{0.45 \pm 0.04}$ & $\mathbf{9.26 \pm 2.81}$ & $\mathbf{15.88 \pm 0.79}$ & $\mathbf{0.74 \pm 0.03}$ & $\mathbf{0.37 \pm 0.06}$ \\
\bottomrule
\end{tabular}
}
\end{sc}
\end{table*}

Table~\ref{tab:ablation_full} expands the compact ablation of Table~\ref{tab:ablation_main}: each row reports MSE, PSNR, SSIM, and IoU as mean $\pm$ $95\%$ CI across the $1{,}000$-sample synthetic benchmark, separately for the homogeneous and layered configurations. The variance is highest at the Base and ${+}$PE configurations, where the optimizer regularly converges to qualitatively different fields across samples. Subsequent additions (FA, $\sigma$, HM, TV) progressively tighten the CI as they constrain the optimization landscape. The single largest jump is the final TV regularizer, which more than halves the diffusivity MSE and nearly doubles IoU on both configurations by suppressing the residual high-frequency noise that the previous ablation rows still admit.

\subsection{Training-level efficiency benchmark}

\begin{table}[h]
\centering
\caption{\textbf{Training-level wall-clock and memory per specimen on a single NVIDIA RTX PRO 6000 (96\,GB)}, the training-level companion to the solver-level Table~\ref{tab:efficiency_main} in the main body. Wall-clock includes the one-time \texttt{torch.compile} overhead (Grid Opt.\ $\sim 6$\,s, NeFTY $\sim 11$\,s). Iterations are reported at convergence or at the saturation plateau. Collocation columns apply only to the PINN family.}
\label{tab:training_benchmark}
\vspace{2pt}
\begin{sc}
\footnotesize
\begin{tabular}{l r r r r r r}
\toprule
\textbf{Method} & \textbf{Params} & \textbf{Colloc.} & \textbf{Iters} & \textbf{ms/iter} & \textbf{Wall-clock} & \textbf{Peak GPU} \\
\midrule
PINN~\cite{raissi2019physics} & $4.91$\,M & $24{,}576$ & $22{,}000$ & $128$ & $47.0$\,min & $10.4$\,GB \\
Causal-PINN~\cite{wang2024causal} & $4.91$\,M & $24{,}576$ & $22{,}000$ & $144$ & $52.7$\,min & $10.4$\,GB \\
DCGD~\cite{hwang2024dcgd} & $4.91$\,M & $24{,}576$ & $22{,}000$ & $159$ & $58.3$\,min & $11.7$\,GB \\
SPINN~\cite{cho2023spinn} & $2.64$\,M & $24{,}576$ & $22{,}000$ & $49.5$ & $18.2$\,min & $\phantom{0}3.9$\,GB \\
Grid Opt. & $0.07$\,M & --- & $10{,}000$ & $47.8$ & $\phantom{0}8.0$\,min & $\phantom{0}1.2$\,GB \\
\textbf{NeFTY (Ours)} & $\mathbf{2.44}$\,\textbf{M} & --- & $\mathbf{10{,}000}$ & $57.5$ & $\mathbf{9.6}$\,\textbf{min} & $\phantom{0}\mathbf{4.3}$\,\textbf{GB} \\
\bottomrule
\end{tabular}
\end{sc}
\end{table}

NeFTY converges in $9.6$ minutes per specimen, comparable to Grid Opt.\ ($8.0$ minutes) and $1.9\times$ to $6.1\times$ faster than the four PINN variants ($18.2$-$58.3$ minutes), even though the PINN baselines were given $2.2\times$ more iterations and $2\times$ the parameter count. The peak GPU memory of NeFTY ($4.3$\,GB) is dominated by the unrolled Jacobi iterations of the forward solve. The discrete adjoint of Section~\ref{sec:adjoint} keeps the solver working set of the backward sweep independent of $N_t$ and stores about one forward field per frame, in contrast to the $\mathcal{O}(K N_g N_t)$ retained iterates of standard backpropagation through time (Table~\ref{tab:efficiency_main}).

\subsection{Frequency-domain diagnostics: Edge F1 and radial power spectrum}

\begin{figure}[h]
  \centering
  \includegraphics[width=0.92\linewidth]{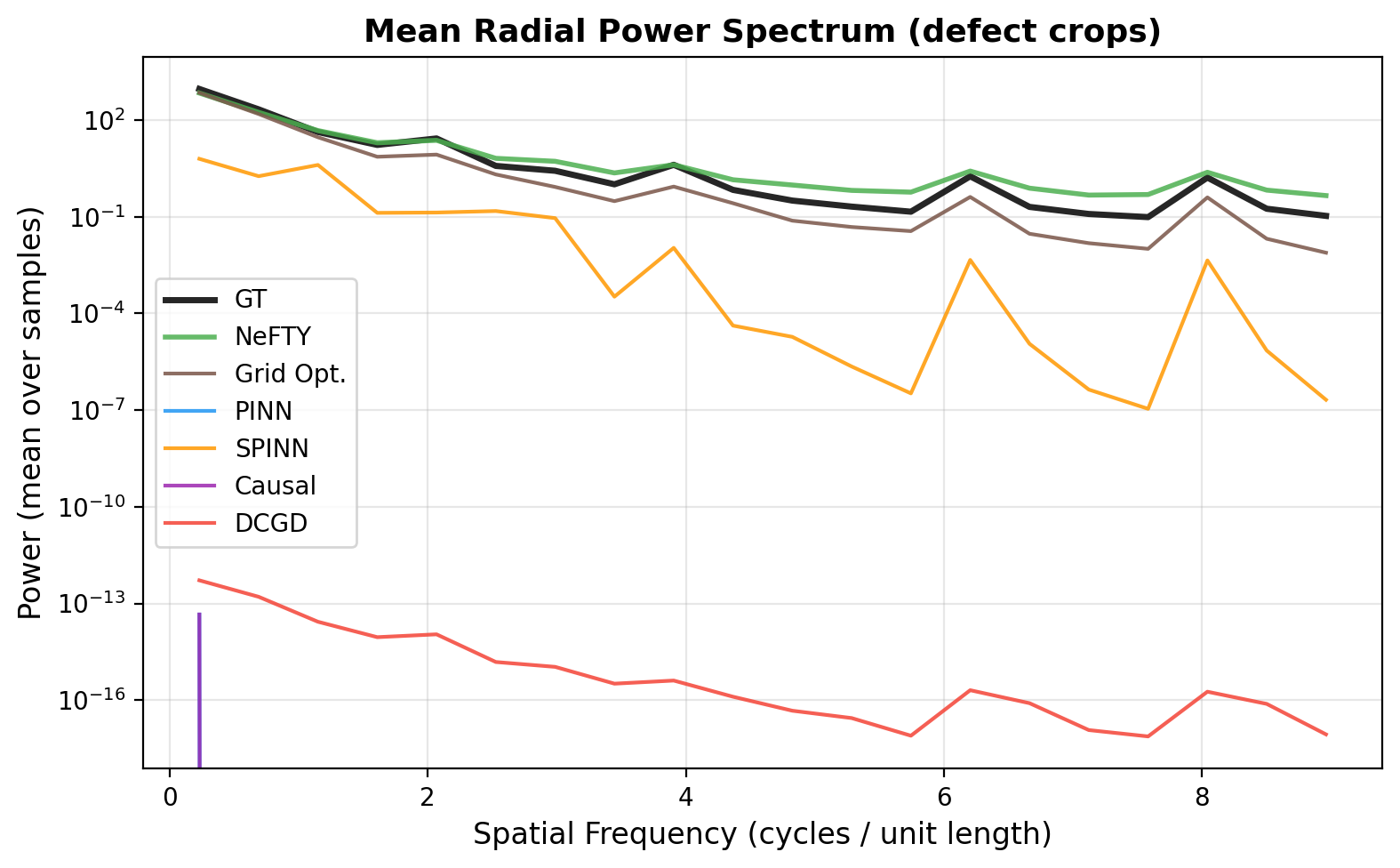}
  \captionsetup{belowskip=-4pt}
  \caption{\textbf{Mean radial power spectrum} of the recovered diffusivity field, averaged across the $1{,}000$-sample homogeneous benchmark. NeFTY (green) tracks the ground-truth spectrum (black) across all spatial frequencies up to the Nyquist limit. Grid Opt.\ tracks at low frequencies but drops off above the mid range, and the four PINN variants collapse to near-zero power.}
  \label{fig:spectral_analysis}
\end{figure}

To quantify whether each method recovers sharp defect boundaries (Section~\ref{ssec:synth}), we threshold the gradient magnitude $\|\nabla \alpha\|$ at $50\%$ of the ground-truth maximum, restrict to a dilated boundary mask, and compute Edge F1. Averaged across the $1{,}000$-sample homogeneous benchmark we obtain $0.470 \pm 0.137$ for NeFTY, $0.029 \pm 0.036$ for Grid Opt., and $0.000$ for vanilla PINN, SPINN, Causal-PINN, and DCGD. NeFTY is the only label-free method whose recovered field carries detectable defect-scale gradients, while the four PINN variants produce featureless interiors. Figure~\ref{fig:spectral_analysis} plots the radial power spectrum, computed over a $16 \times 16 \times 8$ defect crop and radially averaged across the same $1{,}000$-sample benchmark. NeFTY tracks the ground-truth spectrum from the lowest wavenumbers up to the Nyquist limit ($\sim 8$ cycles per unit length in the $z$-direction), while all four PINN variants collapse to near-zero power across the spectrum and Grid Opt.\ falls off above the mid frequencies. The pattern is consistent with Corollary~\ref{cor:hadamard}: the parameter-to-observation map attenuates high spatial frequencies, so a reconstruction recovers defect-scale structure only if it restores that content.

\subsection{Additional qualitative reconstructions}

\begin{figure*}[!htbp]
  \centering
  \includegraphics[width=0.96\linewidth]{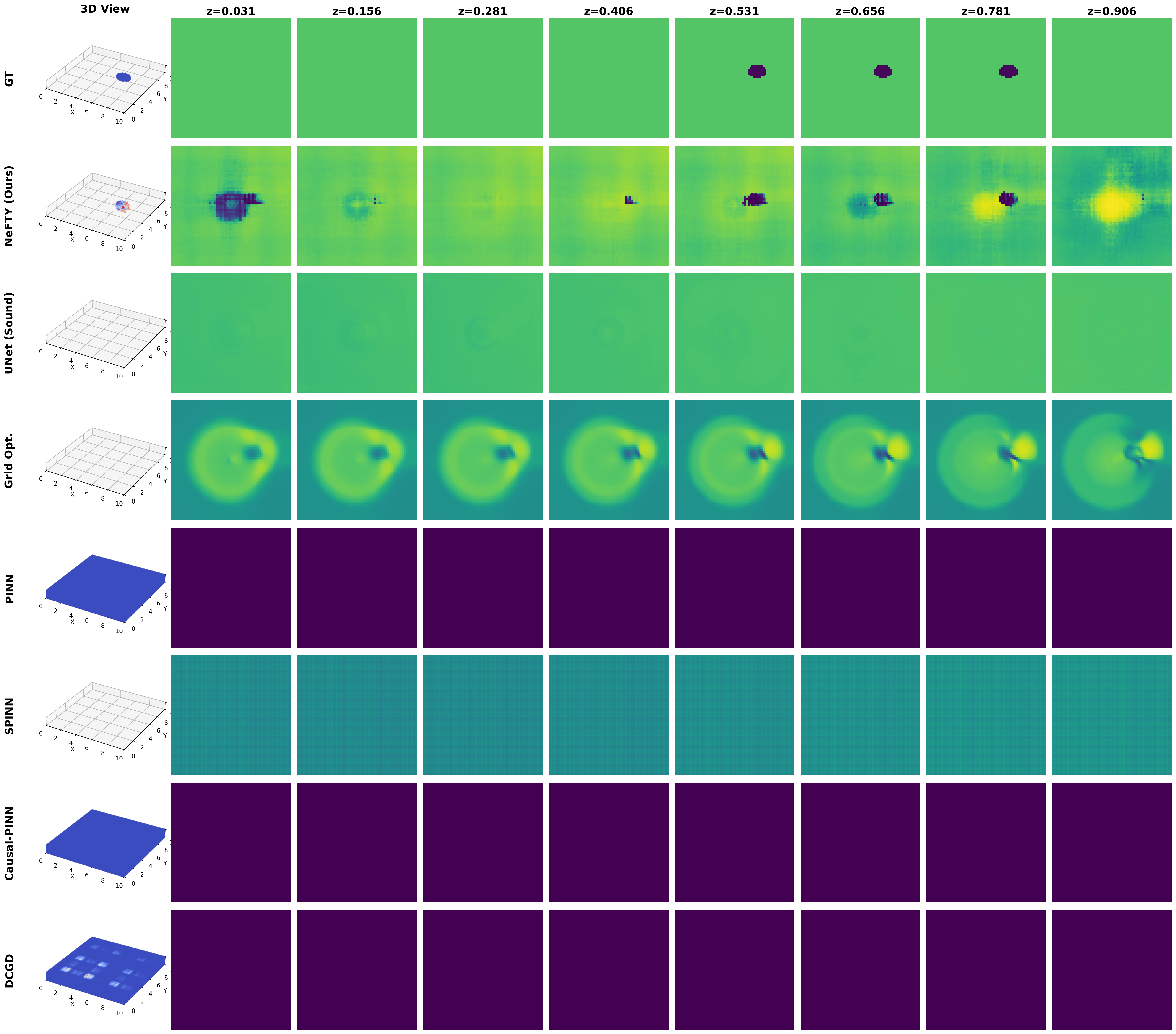}
  \captionsetup{belowskip=-4pt}
  \caption{\textbf{Single-defect reconstruction (homogeneous bulk).} NeFTY recovers a clean defect outline at the correct depth, while Grid Opt.\ exhibits ringing artifacts and the four PINN variants converge to featureless fields.}
  \label{fig:qual_1defect}
\end{figure*}

\begin{figure*}[!htbp]
  \centering
  \includegraphics[width=0.96\linewidth]{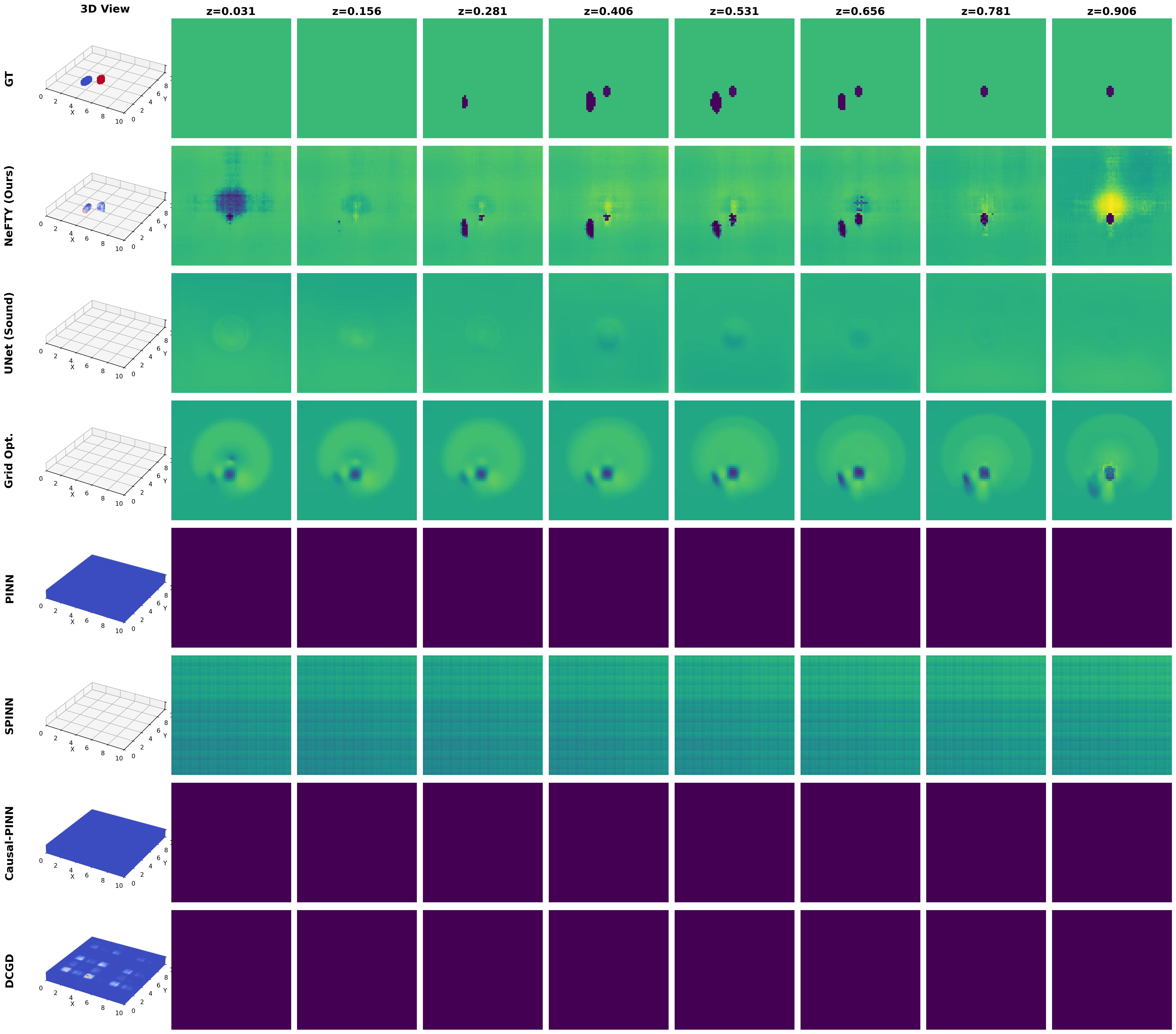}
  \captionsetup{belowskip=-4pt}
  \caption{\textbf{Two-defect reconstruction.} NeFTY separates two laterally adjacent defects without merging their thermal signatures, while Grid Opt.\ blurs them together.}
  \label{fig:qual_2defect}
\end{figure*}

\begin{figure*}[!htbp]
  \centering
  \includegraphics[width=0.96\linewidth]{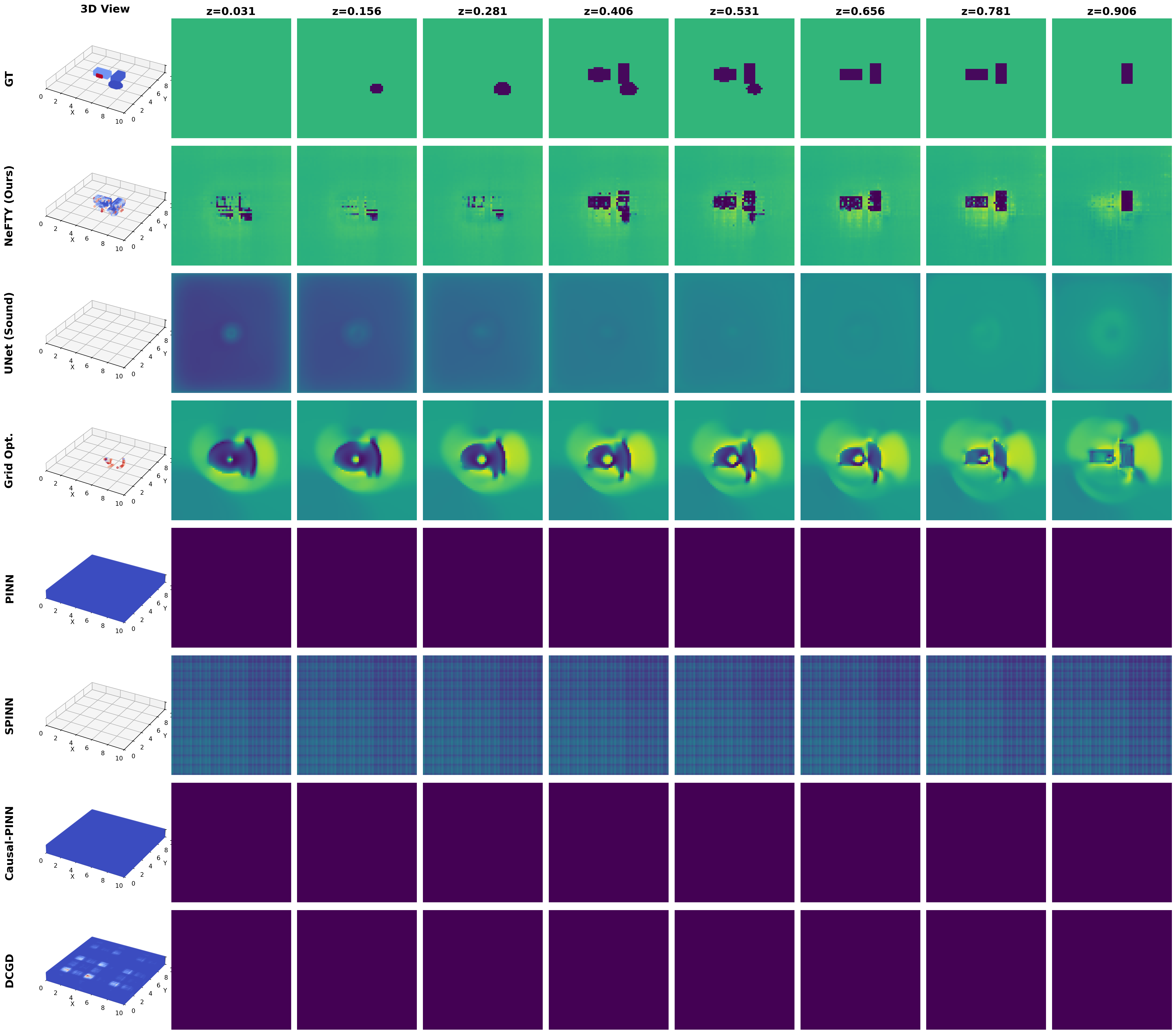}
  \captionsetup{belowskip=-4pt}
  \caption{\textbf{Four-defect reconstruction.} NeFTY isolates each of four defects at the highest density of the synthetic benchmark. The Sound-Only U-Net ghosts all defects, in line with its zero IoU in Table~\ref{tab:robustness_breakdown}.}
  \label{fig:qual_4defect}
\end{figure*}

\begin{figure*}[!htbp]
  \centering
  \includegraphics[width=0.96\linewidth]{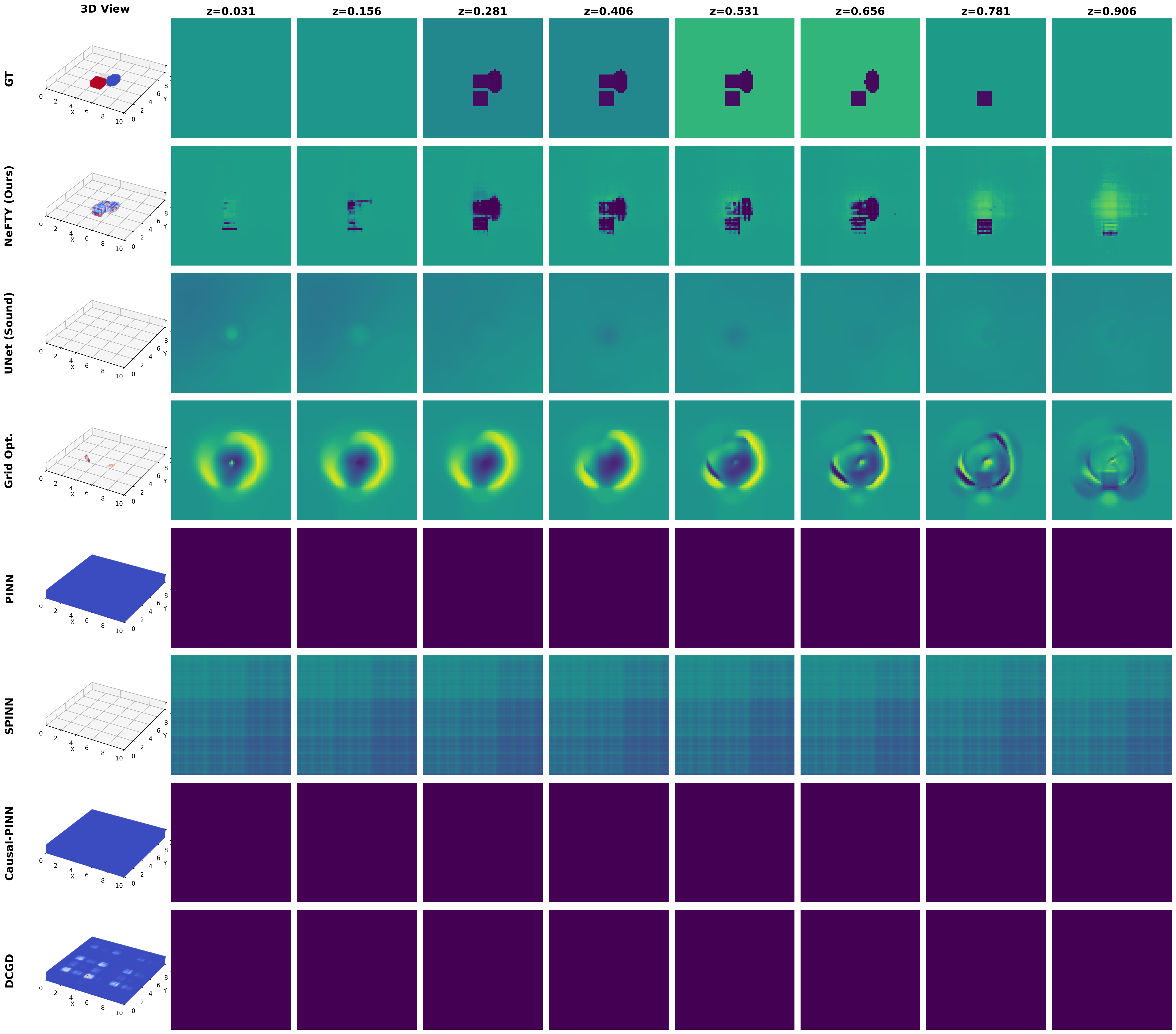}
  \captionsetup{belowskip=-4pt}
  \caption{\textbf{Four-layer composite reconstruction.} The bulk diffusivity steps with depth (visible as changing background intensity in the ground-truth row). NeFTY correctly localizes the embedded defects against this stratified background.}
  \label{fig:qual_4layer}
\end{figure*}

To complement the headline qualitative reconstructions of Figures~\ref{fig:qual_homo} and~\ref{fig:qual_layered} in the main body, we provide a stratified qualitative comparison across defect density and layer count. Figures~\ref{fig:qual_1defect}--\ref{fig:qual_4defect} show single-, two-, and four-defect reconstructions in the homogeneous setting, illustrating that NeFTY's defect-localization quality is preserved as the number of subsurface scatterers grows from one to four. In particular, NeFTY separates two laterally adjacent defects (Figure~\ref{fig:qual_2defect}) without merging their thermal signatures, a regime in which the voxel grid of Grid Opt.\ blurs both defects together. Figure~\ref{fig:qual_4layer} extends the comparison to a four-layer composite where the bulk diffusivity steps with depth (visible as changing background intensity in the ground-truth row). NeFTY isolates the embedded defects against the layered background, confirming the layered-setting trend of Table~\ref{tab:synth_main} at the per-specimen level.

\FloatBarrier

\subsection{Failure mode}

\begin{figure*}[!htbp]
  \centering
  \includegraphics[width=0.96\linewidth]{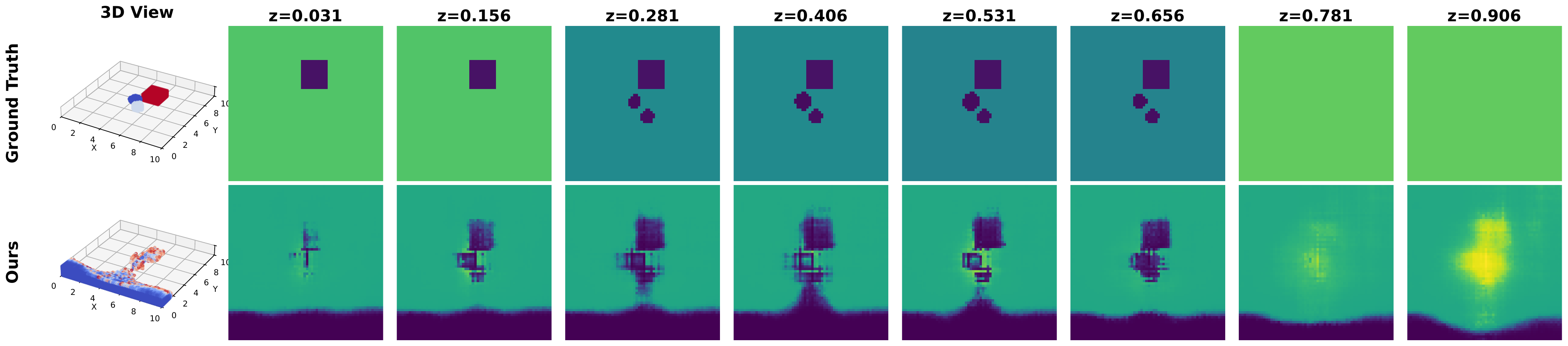}
  \captionsetup{belowskip=-4pt}
  \caption{\textbf{Failure mode: shallow defects close to the heat source.} NeFTY correctly localizes the central defects but introduces a low-diffusivity artifact along the back-face boundary of the lateral plane.}
  \label{fig:failure_mode}
\end{figure*}

We identify one consistent failure mode of NeFTY: when defects are placed shallow and close to the laser-illuminated front surface, the surface-only data become weakly sensitive to the back-face diffusivity, and the optimizer occasionally introduces a spurious low-$\alpha$ artifact along the back boundary that is consistent with the surface decay but not with the ground truth (Figure~\ref{fig:failure_mode}). The defect itself is still recovered at the correct lateral position, so the artifact does not affect 2D segmentation, but it propagates into the depth metric of Table~\ref{tab:real_pvc} for shallow defects. Three remedies target this mechanism directly: stronger back-face priors, a multi-side acquisition that adds back-face thermograms, and a depth-adaptive TV weight that suppresses high-curvature solutions far from observed pixels.

\FloatBarrier

\subsection{Classical PDE-constrained inversion baseline}
\label{appdx:classical}

NeFTY belongs to the discretize-then-optimize, adjoint-state family of PDE-constrained inversion~\cite{hinze2009optimization,jarny1991general,alifanov1994inverse}: the heat equation is enforced by a numerical solver, and gradients reach the unknown field through the discrete adjoint. To separate the contributions of the hard constraint, of the optimizer and regularizer, and of the neural-field representation, we compare NeFTY with a classical member of this family.

\textbf{Protocol.} A bounded voxel diffusivity field is optimized with L-BFGS under total-variation regularization and, separately, $H^1$ (Tikhonov) regularization, through the same solver and discrete adjoint as NeFTY (Section~\ref{sec:method}). Regularization weights are tuned on a calibration set disjoint from the evaluation specimens. The iteration budget is set by convergence: the non-stalled runs reached stationarity by about $300$ L-BFGS iterations (roughly $600$ closure evaluations, since each iteration line-searches through multiple full forward and adjoint solves). The classical arm therefore received more wall-clock than NeFTY: in an eager-mode spot-check ($n = 5$), its median wall-clock was $20.5$\,min per specimen, more than double NeFTY's compiled $9.6$\,min (Table~\ref{tab:training_benchmark}). Both L-BFGS arms are evaluated on $64$ paired benchmark specimens.

\begin{table}[h]
\centering
\caption{\textbf{Classical PDE-constrained inversion versus NeFTY} (volumetric IoU$\uparrow$). The two L-BFGS rows are means over $64$ paired benchmark specimens, while the Grid Opt.\ and NeFTY rows are the full-benchmark values of Table~\ref{tab:synth_main}. The multiples quoted in the text are ratios of means computed on the same $64$ specimens.}
\label{tab:classical}
\vspace{2pt}
\begin{sc}
\footnotesize
\begin{tabular}{l l c c}
\toprule
\textbf{Method} & \textbf{Basis} & \textbf{IoU (Homog.)} & \textbf{IoU (Layered)} \\
\midrule
Grid Opt. & Table~\ref{tab:synth_main}, full benchmark & $0.04$ & $0.03$ \\
Voxel field, L-BFGS\,+\,TV & $64$ paired specimens & $0.129$ & $0.112$ \\
Voxel field, L-BFGS\,+\,$H^1$ & $64$ paired specimens & $0.125$ & $0.127$ \\
\textbf{NeFTY (Ours)} & Table~\ref{tab:synth_main}, full benchmark & $\mathbf{0.45}$ & $\mathbf{0.37}$ \\
\bottomrule
\end{tabular}
\end{sc}
\end{table}

\textbf{Decomposition.} The hard constraint and adjoint alone (Grid Opt., with the same solver, adjoint, and TV regularizer, lacking only the neural prior; Appendix~\ref{appdx:baselines_grid}) set the base. Computed on the same $64$ specimens (ratio of means), a quasi-Newton optimizer with classical regularization multiplies this base by about $3.2\times$, and the neural-field prior adds a further factor of about $3.3\times$ over the L-BFGS arms. Among the evaluated configurations, the representation carries the largest share of the gap, without exhausting optimizer-representation interactions. The two regularizers recover essentially the same fields (a paired comparison detects no difference).

\textbf{Scope of the comparison.} We evaluate the TV/Tikhonov family because it can be tuned to a fair standard under a shared solver and adjoint, which isolates the optimizer and the regularizer. Gauss-Newton inversion~\cite{haber2000optimization} needs repeated linearized PDE solves per iteration, and level-set shape inversion~\cite{santosa1996level} brings its own shape-parameterization machinery. Each merits a dedicated, fully tuned implementation, since an under-tuned one would understate it. Our comparative claims are scoped to the evaluated methods.

\subsection{Irregular defect geometry}
\label{appdx:irregular}

The main benchmark uses ellipsoidal, cylindrical, and box-shaped defects in homogeneous or layered hosts (Appendix~\ref{appdx:experimental_data}), a setting aligned with the piecewise-constant structure that the TV regularizer promotes. To test dependence on regular shapes, we generate two nonregular categories, with $32$ new specimens per category and difficulty parameters fixed before training: irregular blobs, whose boundaries are perturbed by up to $50\%$ of the base defect radius, and curved strata interfaces. All defects are fully embedded. We run the primary NeFTY configuration and the Grid Opt.\ baseline (voxel field through the same solver, adjoint, and TV regularizer, without the neural prior) on each category.

\begin{table}[h]
\centering
\caption{\textbf{Irregular defect geometry.} Volumetric IoU$\uparrow$ on two nonregular categories ($32$ new specimens per category). Brackets give the $95\%$ CI.}
\label{tab:irregular}
\vspace{2pt}
\begin{sc}
\footnotesize
\begin{tabular}{l c c}
\toprule
\textbf{Category} & \textbf{NeFTY 3D IoU [95\% CI]} & \textbf{Grid Opt.\ 3D IoU} \\
\midrule
Irregular blobs & $0.316$ $[0.241, 0.392]$ & $0.000$ \\
Curved strata interfaces & $0.222$ $[0.076, 0.368]$ & $0.000$ \\
\bottomrule
\end{tabular}
\end{sc}
\end{table}

NeFTY recovers volumetric defect structure in both nonregular categories, while Grid Opt.\ reaches IoU $0.000$ in both (Table~\ref{tab:irregular}). We primarily target thermal NDT of specimens whose defects (voids, delaminations, inclusions, flat-bottom holes) form discrete regions in a host material, where piecewise-constant diffusivity predominantly holds. Smoothly varying diffusivity, as in functionally graded materials, is more ill-posed: absent structural assumptions, many smooth interior fields map to nearly identical surface histories (Section~\ref{sec:inverse}). Extending NeFTY to this regime is important future work.

\subsection{Observation noise}
\label{appdx:noise}

We add Gaussian observation noise with standard deviation $\sigma_{\mathrm{n}} \in \{0.5, 1, 2, 5\}\%$ of the peak signal to $16$ synthetic specimens and run the primary NeFTY configuration on each, pairing every noisy run with the same specimen's noise-free run. Table~\ref{tab:noise} reports IoU retention, the mean IoU under noise divided by the mean noise-free IoU on the same specimens (ratio of means).

\begin{table}[h]
\centering
\caption{\textbf{Observation-noise sweep.} NeFTY IoU retention relative to the paired noise-free runs on the same $16$ synthetic specimens (ratio of means).}
\label{tab:noise}
\vspace{2pt}
\begin{sc}
\footnotesize
\begin{tabular}{l c c c c}
\toprule
\textbf{Noise (\% of peak signal)} & $0.5$ & $1$ & $2$ & $5$ \\
\midrule
NeFTY IoU retention (ratio of means) & $96\%$ & $82\%$ & $64\%$ & $37\%$ \\
\bottomrule
\end{tabular}
\end{sc}
\end{table}

Degradation is monotone in the noise level, with no collapse at low noise. The confidence interval of the paired IoU drop excludes zero only at $5\%$. The failure mode under noise is over-segmentation: across all runs of the sweep, including the noise-free anchors ($16$ specimens at five noise levels including $\sigma_{\mathrm{n}} = 0$), only $3.75\%$ return empty predictions. These are within-method curves against NeFTY's own noise-free anchor. Its noise levels are percentages of the peak post-flash signal, whereas the camera that recorded the real PVC sequences, a cooled FLIR SC5000~\cite{wei2023pulsed}, has a manufacturer-specified NETD of about $20$\,mK, which is roughly $0.007\%$ of an absolute room-temperature ($293$\,K) reading.

\subsection{Material contrast and solver diagnostic}
\label{appdx:contrast}

We vary the defect-to-bulk diffusivity contrast over $1\!:\!5$, $1\!:\!10$, $1\!:\!20$, and $1\!:\!50$ with $5$ seeds per ratio and run the primary NeFTY configuration at each ratio. Seed-paired IoU differences relative to $1\!:\!5$ all have confidence intervals straddling zero, so IoU shows no detectable change from $1\!:\!5$ to $1\!:\!50$.

\textbf{Solver diagnostic.} Production inversion uses a fixed budget of $K = 50$ Jacobi sweeps per implicit-Euler step, so the work per step does not depend on contrast. To check whether the linear solve itself becomes harder with contrast, a convergence diagnostic at ground-truth coefficients (first implicit-Euler step, fixed bulk diffusivity $\alpha_{\mathrm{base}} = 0.15$) instead counts Jacobi sweeps until a fixed diagnostic tolerance is reached (Table~\ref{tab:contrast}).

\begin{table}[h]
\centering
\caption{\textbf{Solver convergence diagnostic across material contrast}: Jacobi sweeps until a fixed diagnostic tolerance, at ground-truth coefficients (first implicit-Euler step, fixed $\alpha_{\mathrm{base}} = 0.15$). Production inversion instead runs a fixed budget of $K = 50$ sweeps per step.}
\label{tab:contrast}
\vspace{2pt}
\begin{sc}
\footnotesize
\begin{tabular}{l c c c c}
\toprule
\textbf{Defect:bulk contrast} & $1\!:\!5$ & $1\!:\!10$ & $1\!:\!20$ & $1\!:\!50$ \\
\midrule
Sweeps to tolerance (mean) & $55.4$ & $55.6$ & $55.6$ & $55.6$ \\
\bottomrule
\end{tabular}
\end{sc}
\end{table}

Across all specimens the sweep count spans $53$ to $57$, every solve converged, and the mean is essentially unchanged with contrast ($55.4$ to $55.6$), so solver work is insensitive to contrast in this range. Beyond roughly $1\!:\!1000$ the linear system becomes severely ill-conditioned and the solver stalls (Appendix~\ref{appdx:experimental_data}). Preconditioned or multigrid solvers are the path to that regime (Appendix~\ref{appdx:limitations}).

\subsection{Runtime and memory scaling}
\label{appdx:scaling}

To measure how cost grows with problem size, we record NeFTY's runtime and peak memory on the same hardware used for Tables~\ref{tab:efficiency_main} and~\ref{tab:training_benchmark}, on grids up to $128 \times 128 \times 32$ and sequences up to $200$ frames.

\textbf{Memory.} The adjoint's solver-level working set is $\mathcal{O}(N_g)$, independent of the frame count (Appendix~\ref{appdx:method_adjoint}; $21.9$\,MB versus $18.63$\,GB for autograd in Table~\ref{tab:efficiency_main}). The fitted peak-memory exponents are $0.985$ in voxels and $0.694$ in frames ($R^2 = 0.9998$): measured total memory is a frame-independent solver term plus about one stored field per frame, versus the $K = 50$ iterates per frame that backpropagation through the solver would retain.

\textbf{Runtime.} Runtime follows an affine model in the frame count whose base term scales as $N_g^{0.909}$, with a per-frame marginal cost that stays near-flat across a $64\times$ increase in voxels (Table~\ref{tab:scaling}). Held-out extrapolation of the fitted model is within $11.8\%$.

\begin{table}[h]
\centering
\caption{\textbf{Per-frame marginal cost across grid sizes}: the fitted per-frame term of the affine runtime model, measured on the hardware of Tables~\ref{tab:efficiency_main} and~\ref{tab:training_benchmark}.}
\label{tab:scaling}
\vspace{2pt}
\begin{sc}
\footnotesize
\begin{tabular}{l c c c c}
\toprule
\textbf{Grid} & $32 \times 32 \times 8$ & $64 \times 64 \times 16$ & $96 \times 96 \times 24$ & $128 \times 128 \times 32$ \\
\midrule
Per-frame marginal cost (ms/frame) & $0.49$ & $0.50$ & $0.59$ & $0.68$ \\
\bottomrule
\end{tabular}
\end{sc}
\end{table}

\FloatBarrier

\section{Real PVC Data Details}
\label{appdx:pvc}

\subsection{Datasets}

The PVC-Infrared~\cite{wei2023pulsed} and PVC-Depth~\cite{wei2023depth} benchmarks consist of $19$ and $38$ pulsed-thermography sequences respectively on $100 \times 100 \times 5$\,mm PVC specimens with cylindrical flat-bottom holes of varying diameter and depth, recorded with a cooled mid-wave-infrared camera (FLIR SC5000~\cite{wei2023pulsed}) at second-scale dynamics under flash-lamp excitation. PVC-Infrared annotates each specimen with a 2D binary defect mask derived from CAD drawings and bottom-view inspection, and PVC-Depth additionally annotates discrete defect depths. Neither benchmark provides a 3D volumetric diffusivity ground truth: X-ray CT provides volumetric defect geometry~\cite{garcea2018x}, with reported scan times of $5$ to $96$ hours per specimen~\cite{rashidi2020review} on systems priced from roughly \$100K to over \$1M~\cite{cobos2022cost}, and volumetric diffusivity labels additionally require material characterization or destructive sectioning. The dataset authors themselves describe labeling the 3D spatial location of defects as a substantial challenge~\cite{wei2023depth}. Thermal sequences are temporally resampled and amplitude-normalized as documented in Appendix~\ref{appdx:baselines_swin} for compatibility across baselines.

\subsection{Physical adaptations}

The synthetic benchmark of Appendix~\ref{appdx:experimental_data} is set on micron-scale specimens at microsecond timescales with negligible convective dissipation, while the PVC specimens are centimeter-scale at second-scale dynamics. Three physical adaptations of Appendix~\ref{appdx:bc} are required for NeFTY to apply without re-training the ground-truth simulator.

\textbf{Initial condition.} The Gaussian post-flash profile is replaced by a near-uniform front-face flash $T_0(\mathbf{x}) = T_{\mathrm{amb}} + A \exp(-z^2 / (2 w_z^2))$, with $w_z$ the absorption depth and $A$ calibrated against the first measured frame. This matches the broad flash-lamp excitation used by~\citet{wei2023pulsed,wei2023depth}.

\textbf{Convective Robin boundary.} On the back face we replace the synthetic adiabatic Neumann boundary by a Robin condition $\mathbf{n} \cdot (\alpha \nabla T) = -h\, (T - T_{\mathrm{amb}})$, with $h > 0$ a calibrated convective coefficient. The unitless $h$ is set to match the asymptotic Biot regime of natural convection on PVC over the relevant second-scale window. After the ambient shift of Section~\ref{sec:hardphys}, this becomes $\mathbf{n} \cdot (\alpha \nabla T) = -h\, T$, so the discrete implicit-Euler step remains affine-free and the diagonal-dominance argument of Appendix~\ref{appdx:method_solver} carries over without change.

\textbf{Dimensional rescaling.} The unitless solver evolves in $(t / t_{\mathrm{total}}, \mathbf{x} / L_0)$ coordinates with effective diffusivity $\alpha_{\mathrm{sim}} = \alpha_{\mathrm{phys}}\, t_{\mathrm{total}} / L_0^2$ (Appendix~\ref{appdx:bc}). For PVC we set $L_0 = 100$\,mm (lateral specimen size) and $t_{\mathrm{total}}$ to the recording window of the benchmark, so the unitless diffusivity ranges of the synthetic benchmark map onto the physical PVC diffusivity range. Lateral boundaries remain periodic when the camera footprint is far from the specimen edge. On cropped specimens we replace lateral periodicity by adiabatic Neumann boundaries.

\subsection{Projection from 3D diffusivity to 2D and 2.5D labels}

The reconstructed 3D field $\hat{\alpha}_\theta(x, y, z)$ is reduced to the labels supplied by the benchmarks through two complementary projections.

\textbf{2D defect mask.} For each lateral pixel $(x,y)$, we compute the depth-averaged diffusivity $\bar{\alpha}(x,y)=N_z^{-1}\sum_z \hat{\alpha}_\theta(x,y,z)$ and the normalized one-sided diffusivity deficit $d(x,y)=\left[1-\bar{\alpha}(x,y)/\alpha_{\mathrm{base}}\right]_+$. Because sound pixels should have $\bar{\alpha}(x,y)\approx\alpha_{\mathrm{base}}$, we estimate the reconstruction noise floor by $\hat{\sigma}=1.4826\,\mathrm{MAD}\{d(x,y)\}_{(x,y)}$ and define the mask to be $\mathbf{1}\{d(x,y)>2\hat{\sigma}\}$. The same fixed post-processing rule is applied to NeFTY, Grid Opt., Swin-UNETR, and SPINN, while PPT/TSR follow their publication-default segmentation rules.

\textbf{2.5D defect depth.} For each pixel $p$ inside the predicted 2D mask we extract the through-thickness depth $\hat{d}_p$ as the spatial median of the depths of the $z$-voxels whose $\hat{\alpha}_\theta$ falls below the bulk by more than the noise floor of the defect-free region. PPT and TSR use their published depth formulas with the calibrated PVC diffusivity. Swin-UNETR uses the same spatial-median rule applied to its supervised prediction.

\subsection{Real PVC depth, full metric breakdown}
\label{appdx:real_pvc_depth_full}
\begin{figure*}[!htbp]
  \centering
  \includegraphics[width=0.96\linewidth]{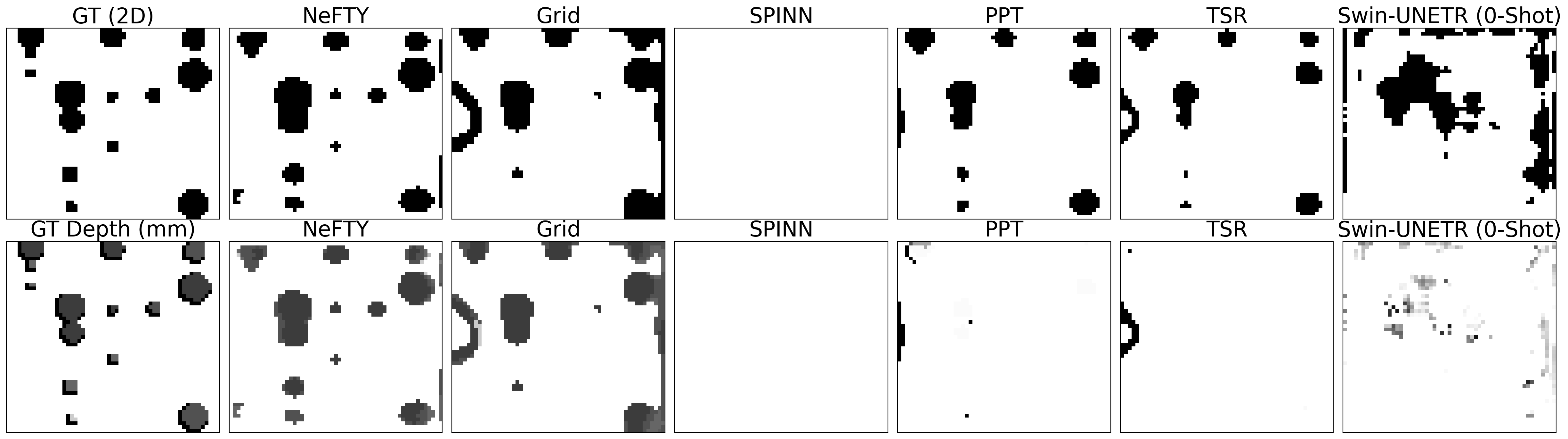}
  \captionsetup{belowskip=-4pt}
  \caption{\textbf{Real PVC reconstruction on a second held-out specimen} (companion to Figure~\ref{fig:real_world}). Row 1: defect masks; Row 2: predicted depth (greyscale, darker is deeper). NeFTY recovers both the defect layout and the depth ordering from the surface IR sequence, in contrast to PPT/TSR which produce coarse depth and SPINN which collapses to a near-uniform field.}
  \label{fig:real_world_extra}
\end{figure*}

We provide the full table and an additional qualitative example in Table~\ref{tab:real_pvc_depth_full} and Figure~\ref{fig:real_world_extra}.

\begin{table}[h]
\centering
\caption{\textbf{Real PVC 2.5D depth recovery, full metric breakdown.} Companion to Table~\ref{tab:real_pvc} in the main body, reporting RMSE in millimetres and the full $\delta < 1.25^k$ threshold accuracies for $k \in \{1, 2, 3\}$ alongside Abs Rel. NeFTY is best label-free on every metric.}
\label{tab:real_pvc_depth_full}
\begin{sc}
\footnotesize
\begin{tabular}{l c c c c c}
\toprule
\textbf{Method} & Abs Rel $\downarrow$ & RMSE (mm) $\downarrow$ & $\delta < 1.25$ $\uparrow$ & $\delta < 1.25^2$ $\uparrow$ & $\delta < 1.25^3$ $\uparrow$ \\
\midrule
Swin-UNETR~\cite{hatamizadeh2022swin} & $3.131$ & $1.983$ & $0.052$ & $0.126$ & $0.190$ \\
PPT~\cite{maldague2002advances} & $3.416$ & $2.075$ & $0.005$ & $0.006$ & $0.073$ \\
TSR~\cite{shepard2002reconstruction} & $13.366$ & $7.655$ & $0.002$ & $0.002$ & $0.002$ \\
SPINN~\cite{cho2023spinn} & $6.737$ & $3.913$ & $0.002$ & $0.002$ & $0.004$ \\
Grid Opt. & $0.941$ & $0.983$ & $0.104$ & $0.399$ & $0.614$ \\
\textbf{NeFTY (Ours)} & $\mathbf{0.465}$ & $\mathbf{0.541}$ & $\mathbf{0.385}$ & $\mathbf{0.621}$ & $\mathbf{0.743}$ \\
\bottomrule
\end{tabular}
\end{sc}
\end{table}

\FloatBarrier

\section{Validation of the Differentiable Heat Diffusion Simulator}
\label{appendix:simulator_validation}
\label{appdx:validation}
We validate the correctness of our differentiable heat diffusion simulator, implemented using an implicit Euler time discretization solved via a Jacobi iterative scheme, through both analytical consistency checks and numerical experiments.

\subsection{Governing Equation and Analytical Behavior}

To validate our simulator, we fix $\alpha$ as a constant, set the source term $Q$ in Eq.~\eqref{eq:full_pde} to zero, and prescribe a Gaussian initial temperature distribution. Under these settings the simulator reduces to
\begin{equation}
\frac{\partial T(\mathbf{x}, t)}{\partial t} = \alpha \nabla^2 T(\mathbf{x}, t),
\label{eq:heat_equation}
\end{equation}
where $T(\mathbf{x}, t)$ denotes temperature and $\alpha$ is a constant, isotropic thermal diffusivity.

For an initial Gaussian temperature distribution
\begin{equation}
T(\mathbf{x}, 0) = A \exp\left(-\frac{\|\mathbf{x}-\mathbf{x}_0\|^2}{2\sigma_0^2}\right),
\label{eq:gaussian_ic}
\end{equation}
where $A > 0$ is the peak amplitude, $\mathbf{x}_0 \in \mathbb{R}^3$ the source center, and $\sigma_0 > 0$ the initial standard deviation along each axis. The analytical solution of Eq.~\eqref{eq:heat_equation} remains Gaussian for all $t > 0$. In particular, the variance along each spatial dimension evolves as
\begin{equation}
\sigma^2(t) = \sigma_0^2 + 2 \alpha t,
\label{eq:variance_growth}
\end{equation}
which implies a linear growth rate
\begin{equation}
\frac{d\sigma^2}{dt} = 2\alpha.
\label{eq:variance_slope}
\end{equation}
This property provides a quantitative criterion for validating the physical fidelity of a numerical diffusion solver.

\subsection{Numerical Setup}

We discretize the spatial domain using a uniform Cartesian grid and advance Eq.~\eqref{eq:heat_equation} in time using an implicit Euler scheme. The resulting linear system at each time step is solved using a fixed number of Jacobi iterations, yielding a fully differentiable simulation pipeline.

Periodic boundary conditions are used in the $x$ and $y$ directions, and zero-flux (Neumann) boundary conditions are applied along the $z$ axis. Temperature observations are taken from the top surface of the domain to match the sensing configuration used in thermal imaging.

\subsection{Gaussian Diffusion Rate Verification}
\begin{figure*}[htbp!]
\centering
\includegraphics[width=\textwidth]{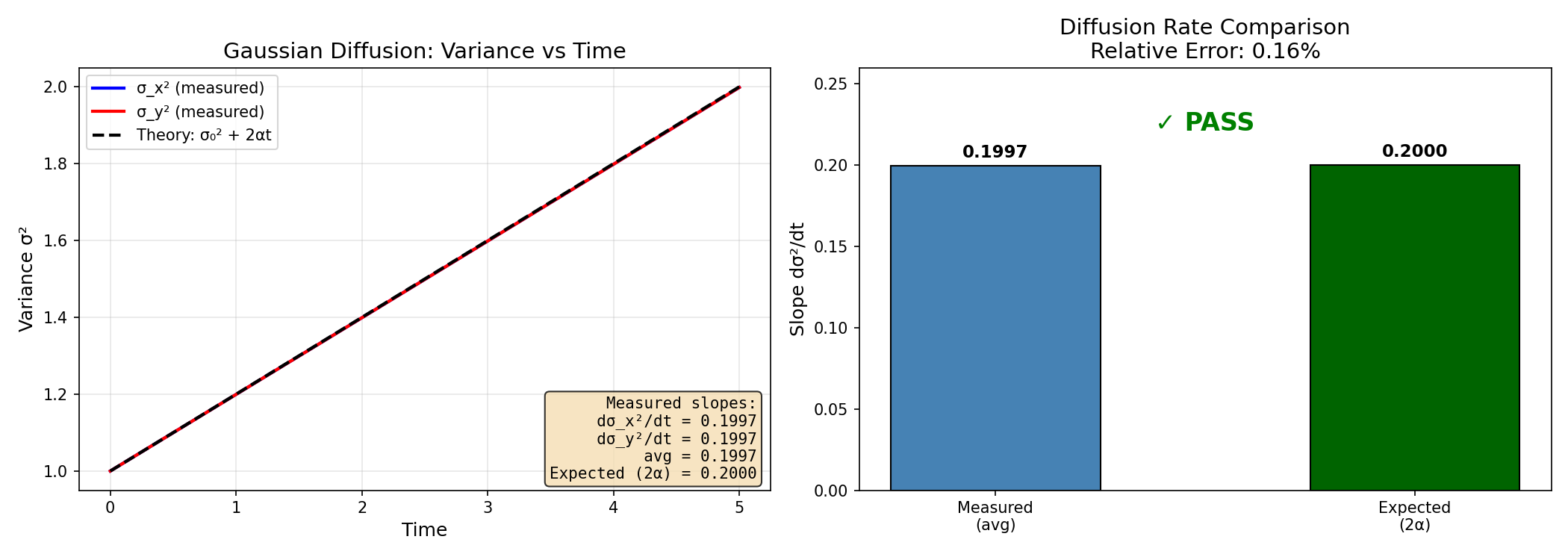}
\caption{\textbf{Gaussian Diffusion Rate Validation.}
(\textit{Left}) Temporal evolution of the temperature variance $\sigma^2$ along the $x$ and $y$ directions, measured on the surface for a diffusing Gaussian heat source with constant diffusivity $\alpha = 0.1$. Both $\sigma_x^2(t)$ and $\sigma_y^2(t)$ grow linearly over time and closely follow the analytical prediction $\sigma^2(t) = \sigma_0^2 + 2\alpha t$.
(\textit{Right}) Comparison between the measured average slope $\mathrm{d}\sigma^2/\mathrm{d}t$ and the theoretical value $2\alpha$, showing a relative error of $0.16\%$.
This result quantitatively confirms that the proposed simulator reproduces the correct diffusion rate of the heat equation.}
\label{fig:gaussion_diffusion_rate}
\end{figure*}
To verify the analytical variance growth in Eq.~\eqref{eq:variance_growth}, we simulate the diffusion of a 3D Gaussian heat source with constant diffusivity $\alpha = 0.1$ in a large domain. At each time step, we compute the temperature-weighted second moments on the surface,
\begin{equation}
\sigma_x^2(t) = \frac{\sum (x - \bar{x})^2 T(x,y,t)}{\sum T(x,y,t)}, \quad
\sigma_y^2(t) = \frac{\sum (y - \bar{y})^2 T(x,y,t)}{\sum T(x,y,t)},
\end{equation}
with $\bar{x} = \sum x \, T(x,y,t) / \sum T(x,y,t)$ and $\bar{y}$ defined analogously, the temperature-weighted lateral centroids on the observed top surface.

Linear regression is performed on $\sigma_x^2(t)$ and $\sigma_y^2(t)$ after an initial transient period. As shown in Fig.~\ref{fig:gaussion_diffusion_rate}, both measured variances exhibit a linear increase over time, with an average slope of $0.1997$, compared to the theoretical value $2\alpha = 0.2000$. The resulting relative error with respect to the analytical solution is $0.16\%$.

\subsection{Qualitative Validation: Constant and Variable Diffusivity}

We further validate the simulator through qualitative visualization of the temperature evolution.

\paragraph{Constant diffusivity.}
\begin{figure*}[htbp!]
\centering
\includegraphics[width=\textwidth]{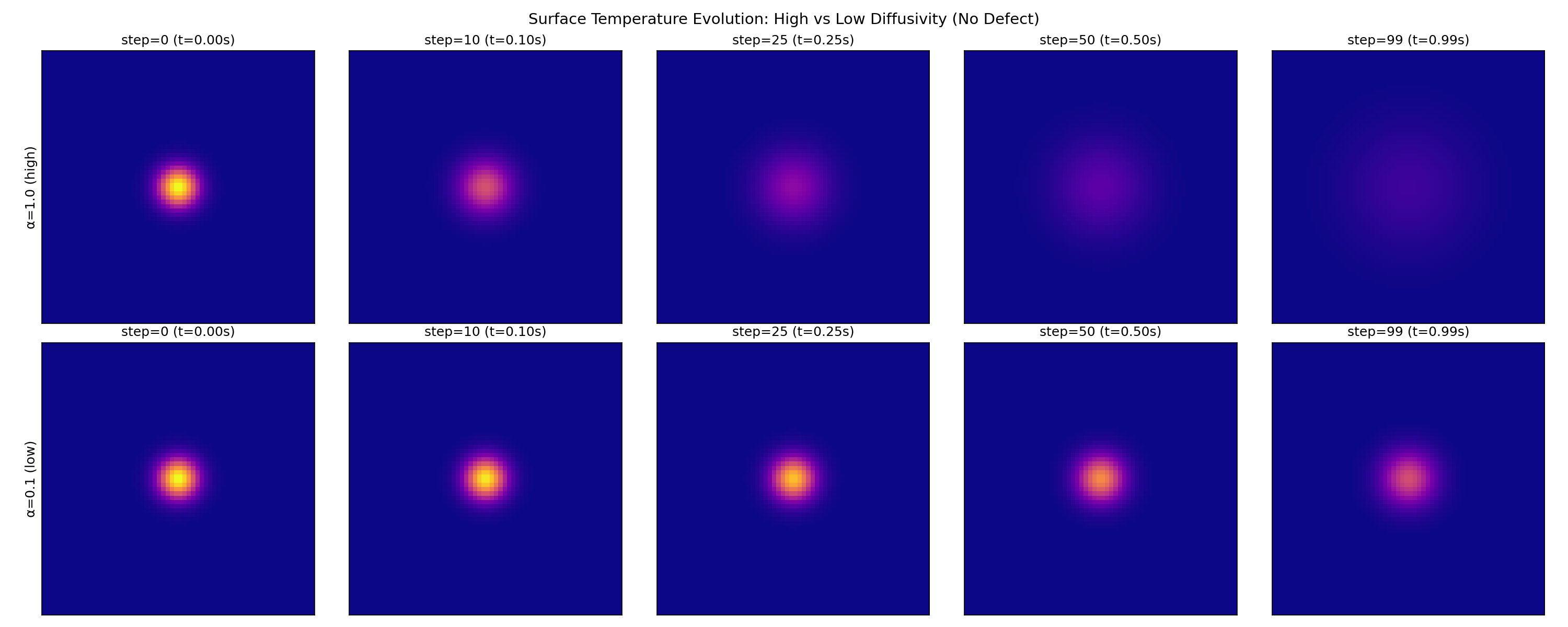}
\caption{\textbf{Effect of Diffusivity Magnitude on Surface Heat Diffusion.}
Surface temperature evolution for a defect-free domain under high diffusivity ($\alpha = 1.0$, top row) and low diffusivity ($\alpha = 0.1$, bottom row), shown at matched time steps.
With identical spatial and temporal discretization, the high-diffusivity case exhibits substantially faster spatial spreading and a more rapid decay of peak temperature, while the low-diffusivity case retains a localized, high-contrast heat profile.
These results qualitatively illustrate the expected dependence of diffusion dynamics on the thermal diffusivity parameter $\alpha$.}
\label{fig:high_diffusivity_evolution}
\end{figure*}
Figure~\ref{fig:high_diffusivity_evolution} shows the surface temperature evolution under uniform diffusivity ($\alpha = 0.1$ and $\alpha = 1.0$). The initially localized heat source spreads isotropically over time, preserving Gaussian symmetry as expected from Eq.~\eqref{eq:heat_equation}.

\paragraph{Variable diffusivity with defect.}
\begin{figure*}[htbp!]
\centering
\includegraphics[width=\textwidth]{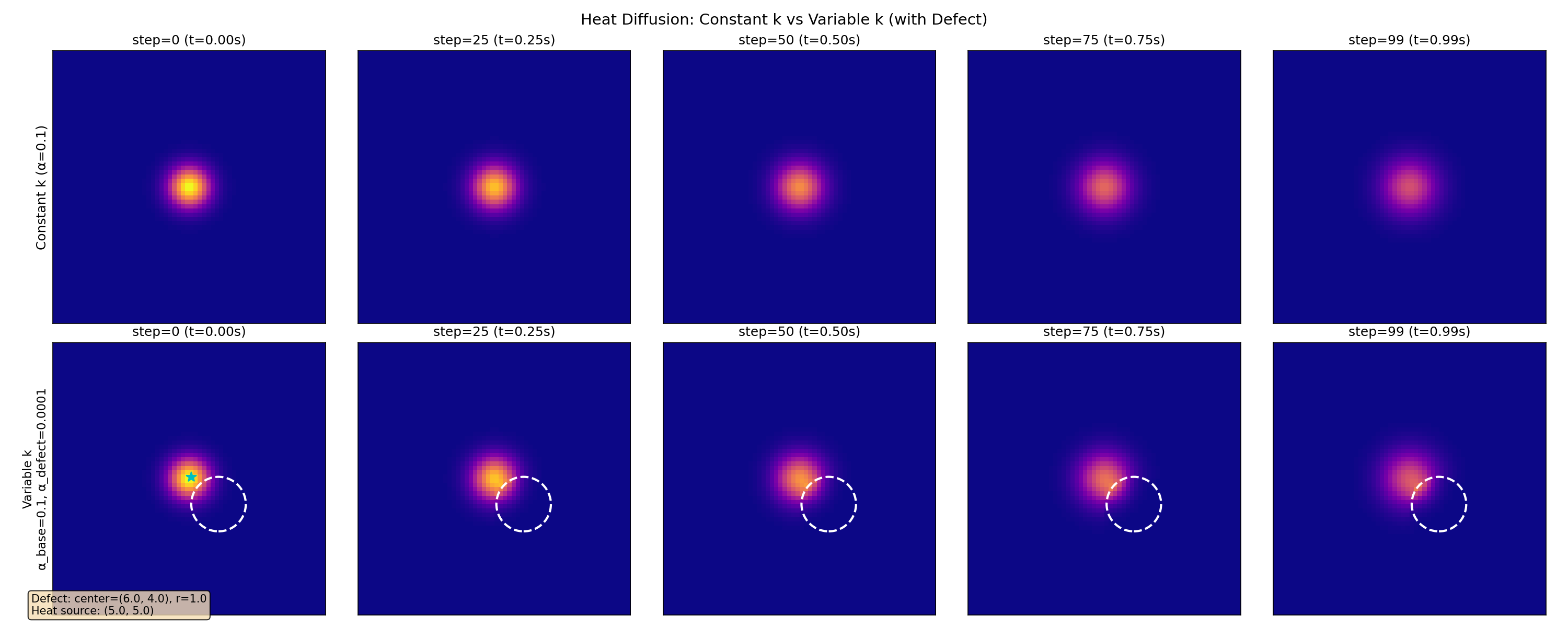}
\caption{\textbf{Surface Temperature Evolution under Constant and Spatially Varying Diffusivity.}
Surface temperature maps at selected time steps for heat diffusion with constant diffusivity ($\alpha = 0.1$, top row) and spatially varying diffusivity with an embedded low-diffusivity defect (bottom row).
In the homogeneous case, the heat source spreads isotropically and preserves Gaussian symmetry over time.
In contrast, the presence of a defect locally impedes heat propagation, leading to asymmetric temperature distributions and delayed diffusion in the defect region (outlined by the dashed circle).
These results demonstrate that the simulator correctly captures both uniform and heterogeneous diffusion behavior.}

\label{fig:surface_evolution}
\end{figure*}
To test spatially varying material properties, we introduce a low-diffusivity spherical defect embedded in a homogeneous background. As illustrated in Fig.~\ref{fig:surface_evolution} (bottom row), the heat propagation is locally impeded near the defect region, resulting in a clearly asymmetric temperature field. This behavior is consistent with the physical interpretation of reduced thermal diffusivity.

\subsection{Effect of Diffusivity Magnitude}
\begin{figure*}[htbp!]
\centering
\includegraphics[width=\textwidth]{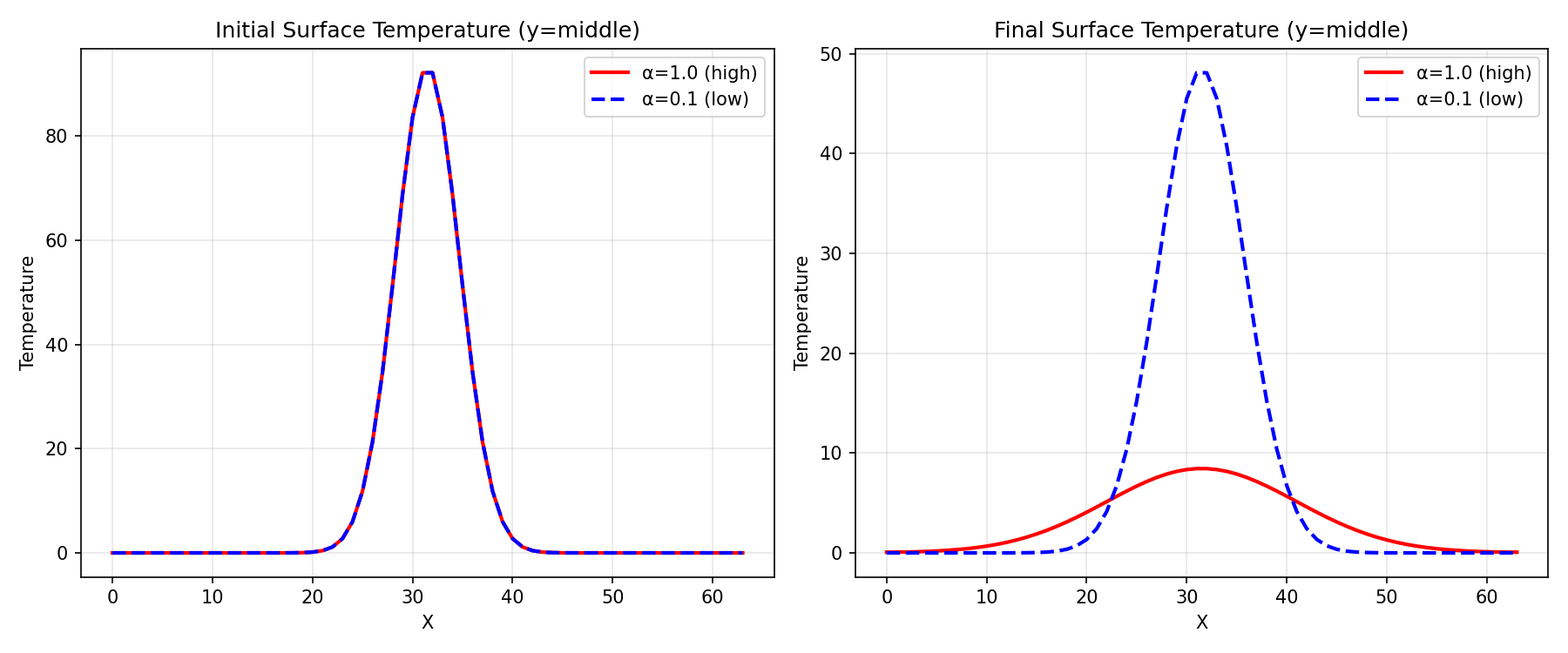}
\caption{\textbf{Surface Temperature Profiles under Different Diffusivities.} 
One-dimensional cross-sections of the surface temperature along the $x$ direction at the midline ($y=\mathrm{middle}$), comparing high diffusivity ($\alpha=1.0$) and low diffusivity ($\alpha=0.1$). 
(\textit{Left}) At the initial time step, both cases exhibit identical Gaussian profiles, confirming consistent initialization.
(\textit{Right}) At the final time step, the high-diffusivity case shows a significantly broader and lower-amplitude profile, reflecting faster spatial spreading of heat, while the low-diffusivity case retains a sharper peak.
These results qualitatively validate the expected dependence of diffusion dynamics on the thermal diffusivity parameter $\alpha$.}
\label{fig:temperature_profile}
\end{figure*}
We also compare diffusion dynamics under different diffusivity values in a defect-free setting. Figure~\ref{fig:temperature_profile} contrasts $\alpha = 1.0$ (high diffusivity) with $\alpha = 0.1$ (low diffusivity) using identical spatial and temporal resolutions. As expected, higher diffusivity leads to faster spreading and lower peak temperatures, while preserving the overall Gaussian structure.

\subsection{Error Sources: Discretization, Solver, and Inversion}

NeFTY's temperature satisfies the discretized heat equation up to the chosen spatial and temporal discretization and the iterative linear-solver tolerance (Section~\ref{sec:hardphys}), and the adjoint gradient is exact up to the same solver tolerance (Appendix~\ref{appdx:method_solver}). Isolating continuous-model error requires an exact solution of the continuous PDE, which exists only for special configurations. We therefore quantify the combined numerical error where an analytic reference exists: the Gaussian diffusion-rate test above gives a relative error of $0.16\%$, and Table~\ref{tab:efficiency_main} reports the simulation error of the solver. The benchmark metrics of Section~\ref{sec:experiments} measure inversion error against the ground-truth diffusivity.

\section{Limitations and Future Work}
\label{appdx:limitations}

\textbf{Per-specimen optimization.} NeFTY targets offline inspection: it recovers one specimen in $\sim 10$ minutes on one workstation GPU ($9.6$\,min and $4.3$\,GB of peak memory in Table~\ref{tab:training_benchmark}), from the same one-sided, non-contact acquisition of a single surface temperature sequence that standard thermographic methods (PPT/TSR) already use, and without volumetric training labels. For reference, X-ray CT provides volumetric defect geometry~\cite{garcea2018x}, with reported scan times of $5$ to $96$ hours per specimen~\cite{rashidi2020review} on systems priced from roughly \$100K to over \$1M~\cite{cobos2022cost}, and volumetric diffusivity labels additionally require material characterization or destructive sectioning. Supervised feedforward inversion (e.g., U-Net) runs in milliseconds once trained but requires volumetric labels. Online or high-throughput screening calls for lower per-specimen latency. Amortized initialization, for example meta-learning or hypernetworks that predict a good initialization for the neural field, is an important future direction. The measured memory and runtime scaling (Appendix~\ref{appdx:scaling}) supports growing the problem size.

\textbf{Quantitative diffusivity under high contrast.} Our benchmark uses a defect-to-bulk diffusivity contrast of $\sim 1\!:\!20$ (Appendix~\ref{appdx:experimental_data}), and IoU shows no detectable change from $1\!:\!5$ to $1\!:\!50$ (Appendix~\ref{appdx:contrast}). At realistic air-to-solid ratios ($> 1\!:\!1000$), the linear system $\mathbf{A}(\alpha_\theta)$ becomes severely ill-conditioned and the iterative solver stalls. NeFTY recovers defect \emph{geometry} in the evaluated regime (Section~\ref{ssec:synth}, Section~\ref{ssec:real}), while the precise quantitative magnitude of $\alpha$ inside high-contrast voids is weakly determined by the surface data, because the surface-temperature signature saturates as the diffusion time $L^2 / \alpha$ inside an insulator diverges for $\alpha \to 0$. Preconditioned or multigrid solvers within the differentiable loop are the path to the regime of stiff air voids.

\textbf{Material and domain model.} The solver assumes a rectilinear slab with isotropic, temperature-independent effective diffusivity. Non-Cartesian geometries, anisotropy, and temperature-dependent coefficients are natural extensions for deployment beyond this setting. We primarily target specimens whose defects form discrete regions in a host material, where piecewise-constant diffusivity predominantly holds, and NeFTY also recovers irregular blobs and curved strata interfaces (Appendix~\ref{appdx:irregular}). Smoothly varying diffusivity, as in functionally graded materials, is more ill-posed and is important future work.

\textbf{Volumetric real labels.} The real PVC datasets used in Section~\ref{ssec:real} provide 2D defect masks and discrete defect depths from CAD. No public NDT dataset provides volumetric diffusivity ground truth, because X-ray CT supplies volumetric defect geometry at the time and cost above, and diffusivity labels additionally need material characterization or destructive sectioning (Appendix~\ref{appdx:pvc}). The full volumetric claim therefore rests on the $1{,}000$-sample synthetic benchmark with exact 3D ground truth, generated by a finite-volume engine independent of NeFTY's solver (Appendix~\ref{appdx:experimental_data}), while the real experiments validate 2D segmentation and 2.5D depth recovery under real acquisition. A benchmark that pairs surface thermograms with volumetric diffusivity ground truth would be a standalone materials-science contribution and would enable direct volumetric validation of label-free methods.



\end{document}